\documentclass[review]{elsarticle}

\usepackage{hyperref}
\usepackage{amsmath}
\usepackage{amsfonts}
\usepackage{amsthm}
\usepackage{amsbsy}
\usepackage{appendix}
\usepackage{epstopdf}
\usepackage{subfloat}
\usepackage{float}
\usepackage{array}
\usepackage{color}
\usepackage{inputenc}
\usepackage{hyperref}
\usepackage{caption}
\usepackage{graphicx}
\usepackage{enumitem}   
\usepackage{tabularx} 
\usepackage{booktabs}
\usepackage{subcaption}
\numberwithin{equation}{section}
\allowdisplaybreaks

\DeclareMathOperator*{\argmin}{arg\,min}
\usepackage[linesnumbered,ruled,vlined]{algorithm2e}
\SetKwInput{KwInput}{Input}                
\SetKwInput{KwOutput}{Output}              
\SetNlSty{textbf}{}{.}
\IncMargin{.2em}
\SetAlgoNlRelativeSize{-0.5}
	
\journal{European Journal of Mechanics / A Solids}

\begin{document}

\begin{frontmatter}

\title{Deep Autoencoder based Energy Method for the Bending, Vibration, and Buckling Analysis of Kirchhoff Plates}

\author[add3,add4,add5]{Xiaoying Zhuang\fnref{email2}}
\fntext[email2]{email: zhuang{@}ikm.uni-hannover.de}
\author[add3]{Hongwei Guo\fnref{email1}}
\fntext[email1]{email: ghway0723{@}gmail.com}
\author[add6]{Naif Alajlan \fnref{email3}}
\fntext[email3]{email: najlan@ksu.edu.sa}
\author[add1,add2]{Timon Rabczuk\corref{mycorrespondingauthor}}
\cortext[mycorrespondingauthor]{Corresponding author}
\ead{timon.rabczuk@tdtu.edu.vn}

\address[add1]{Division of Computational Mechanics,\\ Ton Duc Thang University,\\ Ho Chi Minh City, Vietnam}
\address[add2]{Faculty of Civil Engineering,\\ Ton Duc Thang University,\\ Ho Chi Minh City, Vietnam}
\address[add3]{Institute of Continuum Mechanics,\\ Leibniz Universität Hannover,\\ Hannover, Germany}
\address[add4]{Department of Geotechnical Engineering,\\ Tongji University,\\ Shanghai, China.}
\address[add5]{Key Laboratory of Geotechnical and Underground Engineering of Ministry of Education,\\ Tongji University,\\ Shanghai, China.}
\address[add6]{ALISR Laboratory,\\ College of Computer and Information Sciences,\\ King Saud University,\\ P. O. Box 51178, Riyadh 11543, Saudi Arabia}

\begin{abstract}
In this paper, we present a deep autoencoder based energy method (DAEM)  for the bending, vibration and buckling analysis of Kirchhoff plates. The DAEM exploits the higher order continuity of the DAEM and integrates a deep autoencoder and the minimum total potential principle in one framework yielding an unsupervised feature learning method. The DAEM is a specific type of feedforward deep neural network (DNN) and can also serve as function approximator. With robust feature extraction capacity, the DAEM can more efficiently identify patterns behind the whole energy system, such as the field variables, natural frequency and critical buckling load factor studied in this paper. The objective function is to minimize the total potential energy. The DAEM performs unsupervised learning based on random generated points inside the physical domain so that the total potential energy is minimized at all points. For vibration and buckling analysis, the loss function is constructed based on Rayleigh's principle and the fundamental frequency and the critical buckling load is extracted. A scaled hyperbolic tangent activation function for the underlying mechanical model is presented which meets the continuity requirement and alleviates the gradient vanishing/explosive problems under bending analysis. The DAEM can be easily implemented and we employed the Pytorch library and the LBFGS optimizer. A comprehensive study of the DAEM configuration is performed for several numerical examples with various geometries, load conditions, and boundary conditions.  
\end{abstract}

\begin{keyword}
Deep learning\sep  Autoencoder\sep Activation function\sep Energy method\sep Kirchhoff plate\sep Vibration\sep Buckling.
\end{keyword}

\end{frontmatter}

\section{Introduction}
Thin plate models  are commonly used in engineering and mechanics \cite{ventsel2001thin} due to their computational efficiency. Their mechanical analysis are of major importance in engineering practice. Due to the limitations of analytical methods, a variety of numerical methods have been developed including the finite element method \cite{bathe2006finite,hughes2012finite}, boundary element method \cite{katsikadelis2016boundary,brebbia2016boundary}, meshfree method \cite{NGUYEN2008763,bui2011moving}, isogeometric analysis (IGA) formulations \cite{nguyen2015isogeometric}, numerical manifold method \cite{zheng2013numerical,guo2018linear,guo2019numerical} and recently deep learning based methods \cite{anitescu2019artificial, guo2019deep,nguyen2019deep}. 

Deep learning was proposed in 2006 \cite{hinton2006fast, bengio2007greedy}. It is an unsupervised feature learning method with neural network architectures including multiple hidden layers \cite{lecun2015deep}. Equipped with this hierarchical structure, it can extract information from complicated raw input data with multiple levels of abstraction through a layer-by-layer process \cite{goodfellow2016deep}. Various variants such as multilayer perceptron (MLP), convolutional neural networks (CNN) and recurrent/recursive neural networks (RNN) \cite{patterson2017deep} have been developed and applied to e.g. image processing \cite{yang2018visually,kermany2018identifying}, object detection \cite{ouyang2015deepid,zhao2019object}, speech recognition \cite{amodei2016deep,nassif2019speech}, biology \cite{yue2018deep,ching2018opportunities} and even finance \cite{heaton2017deep,fischer2018deep}. 

Artificial neural networks (ANN) can be traced back to the 1940's \cite{mcculloch1943logical} but they became especially popular in the past few decades due to the vast development in computer science and computational science such as backpropagation technique and advances in deep neural networks. Due to the simplicity and feasibility of ANNs to deal with nonlinear and mult-dimensional problems, they were applied in inference and identification by data scientists \cite{dias2004artificial}. They were also adopted to solve partial differential equations (PDEs) \cite{lagaris1998artificial,lagaris2000neural,mcfall2009artificial} but shallow ANNs are unable to learn the complex nonlinear patterns effectively. With improved theories incorporating unsupervised pre-training, stacks of auto-encoder variants, and deep belief nets, deep learning has become also an interesting alternative to classical methods such as FEM. 

According to the universal approximation theorem \cite{FUNAHASHI1989183,HORNIK1989359}, any continuous function can be approximated by a feedforward neural network with one single hidden layer. However, the number of neurons of the hidden layer tends to increase exponentially with increasing complexity and non-linearity of a model. Recent studies show that DNNs render better approximations for nonlinear functions \cite{mhaskar2016deep}. Some researchers employed deep learning for the solution of PDEs. E et al. developed a deep learning-based numerical method for high-dimensional parabolic PDEs and back-forward stochastic differential equations \cite{weinan2017deep,han2018solving}. Raissi et al. \cite{RAISSI2019686} introduced physics-informed neural networks for supervised learning of nonlinear partial differential equations. Beck et al. \cite{Beck_2019}  employed deep learning to solve nonlinear stochastic differential equations and Kolmogorov equations. Sirignano and Spiliopoulos \cite{sirignano2018dgm} provided a theoretical proof for deep neural networks as PDE approximators, and concluded that it converged as the number of hidden layers tend to infinity.  Anitescu et al. \cite{anitescu2019artificial}, Guo et al. \cite{guo2019deep}, and Nguyen-Thanh et al. \cite{nguyen2019deep} applied deep neural networks for finding the solutions for second and forth order boundary value problems.

The learning ability of deep neural networks has been enhanced with different architectures, such as deep belief network (DBN) or deep autoencoder (DAE) \cite{shao2017novel,yu2019deep}. DAE is widely used in dimensionality reduction and feature learning. It has also been proven to be an effective way to learn and describe latent codes that reflect meaningful variations in data with an encoding and decoding layer \cite{snoek2012nonparametric}. DAE seems therefore ideally suited for learning and describing the underlying physical patterns from the governing partial differential equation or associated potential energy. In this paper, we therefore propose a deep autoencoder based energy learning method for Kirchhoff plate analysis. In this context, we exploit the higher order continuity of the DAE approximation.

The paper is organised as follows: 
First, we describe the Kirhhoff plate model. Then we introduce the basic elements of the deep autoencoder theory, and present a tailored activation function for this mechanical model. Subsequently, the deep autoencoder based energy method is presented. Finally, we demonstrate the efficiency and accuracy of the DAEM for various benchmark problems.

\section{Kirchhoff plate model}
\label{sec2:concept}
\begin{figure}
\captionsetup{width=0.85\columnwidth}
\centering
\begin{tabular}{c}
\includegraphics[width=.75\columnwidth]{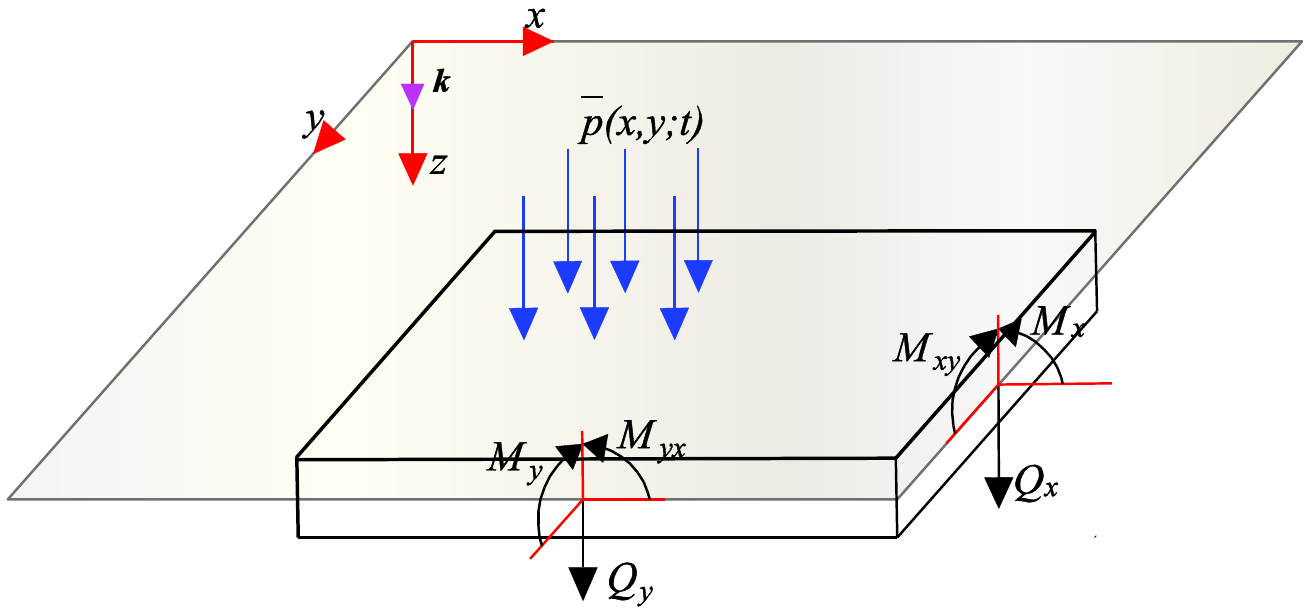}
\end{tabular}
\caption{Kirchhoff thin plate in the Cartesian coordinate system.}
\label{Figure1:plate}
\end{figure}
Consider a Kirchhoff plate as shown in Figure~\ref{Figure1:plate}. The displacement field  can be expressed as:
\begin{equation}
\begin{aligned}
& u\left ( x,y,z \right )=-z\frac{\partial w}{\partial x}, \\
& v\left ( x,y,z \right )=-z\frac{\partial w}{\partial y}, \\
& w\left ( x,y,z \right )=w\left ( x,y \right ).
\end{aligned}
\label{Displacement field}
\end{equation}
where the relation between lateral deflection $w\left (x,y \right )$ of the middle surface $(z=0)$ and rotations about the $x$,$y$-axis is given by 
\begin{equation}
\theta_{x} =\frac{\partial w}{\partial x}, \qquad \theta_{y} =\frac{\partial w}{\partial y}.
\label{Rotation}
\end{equation}
The transversal deflection of the mid-plane is regarded as field variable and the corresponding bending and twisting curvatures are generalized strains:
\begin{equation}
	k_{x}=-\frac{\partial^2 w}{\partial x^2},\: k_{y}=-\frac{\partial^2 w}{\partial y^2},\: k_{xy}=-2\frac{\partial^2 w}{\partial x \partial y}.
\label{generalizedstrain}	
\end{equation}
The geometric equations can be then obtained as:
\begin{equation}
\textbf{\textit{k}}=\begin{Bmatrix}
k_{x}\\ 
k_{y}\\ 
k_{xy}
\end{Bmatrix}=-\begin{Bmatrix}
\frac{\partial^2 w}{\partial x^2} \\[5pt]
\frac{\partial^2 w}{\partial y^2} \\[5pt]
2\frac{\partial^2 w}{\partial x \partial y}
\end{Bmatrix}=\textbf{\textit{L}}w,
\end{equation}           
with $\textbf{\textit{L}}$ being the differential operator defined as $\textbf{\textit{L}}=-\begin{pmatrix}
				\frac{\partial^2 }{\partial x^2} &\frac{\partial^2 }{\partial y^2}  & 2\frac{\partial^2 }{\partial x \partial y}
				\end{pmatrix}^{T}$. 
				
Accordingly, the bending and twisting moments shown in Figure~\ref{Figure1:plate} can be expressed as:
\begin{equation}
\begin{aligned}
& M_{x}=-D_{0}\left ( \frac{\partial^2w }{\partial x^2} +\nu \frac{\partial^2w }{\partial y^2} \right ), \\
& M_{y}=-D_{0}\left ( \frac{\partial^2w }{\partial y^2} +\nu \frac{\partial^2w }{\partial x^2} \right ), \\
& M_{xy}=M_{yx}=-D_{0}\left ( 1-\nu \right )\frac{\partial^2w }{\partial xy}.
\end{aligned}
\label{moment}
\end{equation}				
Here, $D_{0}=\frac{Eh^{3}}{12\left ( 1-\nu^{2} \right )}$	is the bending rigidity, $E$ and $\nu$ denote the Young's modulus and Poisson ratio, and $h$ is the thickness of the thin plate.	It can be rewritten in a Matrix form
\begin{equation}
	\textbf{\textit{M=Dk}}
\end{equation}	 
with $\mathit{ \mathit{D}}=D_{0}\begin{bmatrix}
1 &  \nu & 0 \\ 
  \nu& 1  &0 \\ 
 0& 0 & \left ( 1-\nu \right )/2
\end{bmatrix}$.	
The shear forces of the mid-surface is obtained in terms of the bending and twisting moments as
\begin{equation}
Q_{x}=\frac{\partial M_{x}}{\partial x}+\frac{\partial M_{xy}}{\partial y},\, \, Q_{y}=\frac{\partial M_{xy}}{\partial x}+\frac{\partial M_{y}}{\partial y}.
\label{shearforce}
\end{equation}	

The boundary conditions can be categorized into three parts, namely,
\begin{equation}
\partial\Omega =\Gamma_{1}+\Gamma_{2}+\Gamma_{3}.
\label{boundary}
\end{equation}

At the clamped boundary, $\Gamma_{1}:  w=\tilde{w}, \  \frac{\partial w}{\partial n} = \tilde{ \theta}_{n} $, $w=\tilde{w}, \  \tilde{ \theta}_{n} $ are functions of the arc length along this boundary. At the simply supported edge boundary, $\Gamma_{2}:  w=\tilde{w}, \  M_{n} =\tilde{ M}_{n}$, where $\tilde{ M}_{n} $ is also a function of arc length along this boundary.  At the free boundary conditions, $\Gamma_{3}:  M_{n} =\tilde{ M}_{n} , \  \frac{\partial M_{ns} }{\partial s}+Q_{n}=\tilde{q}$, where $\tilde{q}$ is the load exerted along this boundary where $\textit{\textbf{n}},\textit{\textbf{s}}$ corresponds to the normal and tangent directions along the boundaries.

The total potential energy consists of the strain energy $\textit{U}$ and the potential energy of the external forces $\textit{W}$:
\begin{equation}
\Pi = U+W_{ext}.
\label{totalenergy}
\end{equation}	
where 
\begin{equation}
\begin{aligned}
U&=\frac{1}{2} \mathit{ \mathit{D_{0}}} \int_{\Omega } \left [ ( k_{x}+k_{y} )^2+2\left ( 1-\nu  \right )\left ( k_{xy}^2-k_{x}k_{y}  \right )\right ]d\Omega\\
&=\frac{1}{2}\int_{\Omega } \textbf{\textit{k}}^T\textbf{\textit{Dk}}d\Omega.
\end{aligned}
\label{strainenergy}
\end{equation}	
and 
\begin{equation}
W_{ext}=-\left [ \int_{\Omega } p\left ( x,y \right )wd\Omega +\int _{\Gamma_{3}}\tilde{q}wd\Gamma-\int _{\Gamma_{M}}\tilde{M_n}\frac{\partial w}{\partial n}d\Gamma \right],
\label{externalwork}
\end{equation} with $\Gamma_{M}=\Gamma_{2}+\Gamma_{3}$.

For vibration analysis assuming that the plate is undergoing harmonic vibrations, we can approximate the vibrating mid-surface of the plate by  
\begin{equation}
w\left(x,y,t\right)=W\left(x,y\right)sin\omega t,
\label{externalwork}
\end{equation}
The maximum kinetic energy can be obtained by choosing $\textup{cos} \omega t=1$ as
\begin{equation}
K_{max}=\frac{\omega^2}{2}\int_\Omega\rho hW^2\left(x,y\right)d\Omega
\end{equation}
The maximum strain energy $U_{max}$ occurs when $\textup{sin} \omega t=1$. Accounting for the $Rayleigh's \ principle$, the lowest natural frequency of a vibrating plate can be obtained by setting:
\begin{equation}
K_{max}=U_{max}
\end{equation} where
$U=\frac{1}{2} \mathit{ \mathit{D_{0}}} \int_{\Omega } \left [ ( \nabla^2W )^2+2\left ( 1-\nu  \right )\left ( \left ( \frac{\partial^2W }{\partial xy} \right )^2-\frac{\partial^2 W}{\partial x^2} \frac{\partial^2 W}{\partial y^2}\right )\right ]d\Omega$

The $Rayleigh's \ quotient$ can be defined as
\begin{equation}
\omega^2=\frac{2U_{max}}{\int_\Omega\rho hW^2\left(x,y\right)d\Omega},
\label{Rayleighquotient}
\end{equation} which will be used in the subsequent analysis.

Moreover, we  will introduce the energy criterion to the classical eigenvalue buckling of Kirchhoff plates. Bifurcation of an initial configuration of equilibrium occurs \cite{ventsel2001thin} when the increment in the total potential energy of plate upon buckling equals zero:
\begin{equation}
\Delta \Pi=0
\label{energycriterion}
\end{equation}
The increment in the total potential energy of the plate upon buckling $\Delta \Pi$ can be expressed as strain energy of the bending and twisting of a plate $U$ plus the work done by in-plane forces \cite{ventsel2001thin},
\begin{equation}
\begin{aligned}
\Delta \Pi = \frac{\mathit{ \mathit{D_{0}}} }{2}  \int_{\Omega } \left [ ( \nabla^2w)^2+2\left ( 1-\nu  \right )\left ( \left ( \frac{\partial^2w}{\partial xy} \right )^2-\frac{\partial^2 w}{\partial x^2} \frac{\partial^2 w}{\partial y^2}\right )\right ]d\Omega \\+\frac{1}{2}  \int_{\Omega } \left [ \bar{N_x}\left ( \frac{\partial w}{\partial x} \right ) ^2+\bar{N_y}\left ( \frac{\partial w}{\partial y} \right ) ^2+2\bar{N}_{xy} \frac{\partial w}{\partial x} \frac{\partial w}{\partial y}\right ]d\Omega
\end{aligned}
\label{incertexternalwork}
\end{equation}
Here, $w$ can denotes the perturbed transverse deflection and in-plane forces can be chosen as $\bar{N_x}=\lambda N_x,\bar{N_y}=\lambda N_y,\bar{N}_{xy}=\lambda N_{xy} $ with $\lambda$ a reference value of the in-plane force. 
In practical application, $N_x,N_y,N_{xy}$ is often set to be unity and thus $\lambda$ becomes the desired buckling load factor and can be obtained from Equation \ref{energycriterion}:
\begin{equation}
\begin{aligned}
\lambda = \frac{-\frac{\mathit{ \mathit{D_{0}}} }{2}  \int_{\Omega } \left [ ( \nabla^2w)^2+2\left ( 1-\nu  \right )\left ( \left ( \frac{\partial^2w}{\partial xy} \right )^2-\frac{\partial^2 w}{\partial x^2} \frac{\partial^2 w}{\partial y^2}\right )\right ]d\Omega }{\frac{1}{2}  \int_{\Omega } \left [ N_x\left ( \frac{\partial w}{\partial x} \right ) ^2+N_y\left ( \frac{\partial w}{\partial y} \right ) ^2+2{N}_{xy} \frac{\partial w}{\partial x} \frac{\partial w}{\partial y}\right ]d\Omega}
\end{aligned}
\label{criticalbucklingload}
\end{equation}
which will also be used in the subsequent analysis. The minimum of the load parameter is the critical buckling load. 

\section{Basic theory of a deep autoencoder}
\label{sec3:methodology}
In deep learning, engineers have further enhanced the learning ability of deep neural networks with different architectures, such as deep belief network (DBN) or deep autoencoder (DAE), to mention a few. Autoencoders play a fundamental role in unsupervised learning and are widely chosen deep architectures for dimensionality reduction and feature learning, which has been proven to be an effective way to learn and describe latent codes that reflect meaningful variations from raw input data.

\subsection{Network architecture}
\begin{figure}[H]
\captionsetup{width=0.85\columnwidth}
\centering
\begin{tabular}{c}
\includegraphics[width=.85\columnwidth]{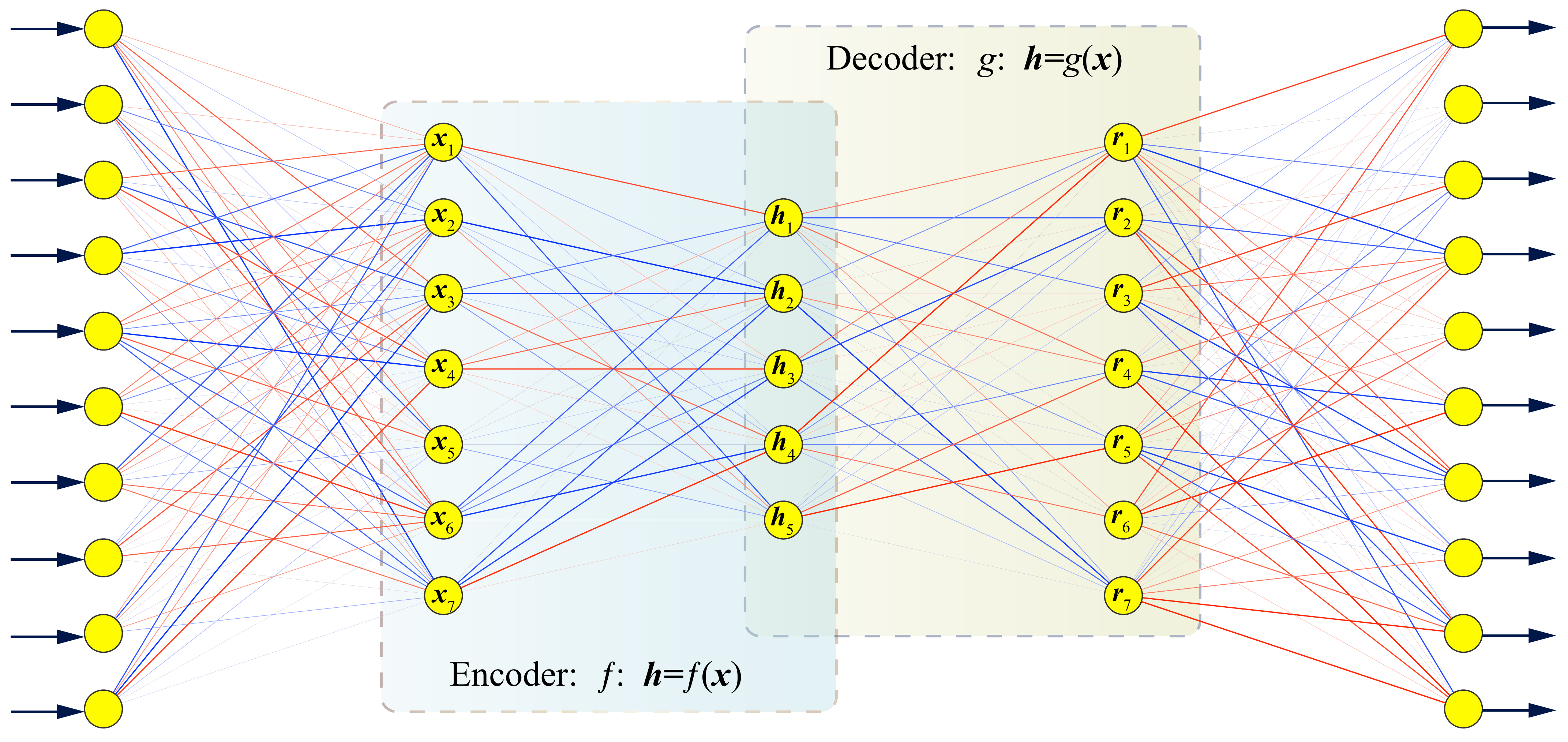}
\end{tabular}
\caption{Basic structure of an autoencoder.}
\label{Figure2:Autoencoder}
\end{figure}

Autoencoders are a specific type of feedforward neural networks including an encoder and a decoder. The basic structure is shown in Figure~\ref{Figure2:Autoencoder}. The network reconstructs the input by mapping them from a high-dimensional space to a low-dimensional space enabling the hidden layer to learn a better representation of the input and the decoder layer then reconstructs the results to another space.

As shown in Figure~\ref{Figure2:Autoencoder}, the fully connected feedforward neural network is composed of multiple layers: an input layer, one encoding layer, one "bottleneck" layer, one decoding layer and one output layer. Each layer consists of one or more nodes called neurons, indicated by small coloured circles in Figure~\ref{Figure2:Autoencoder}. For the interconnected structure, every two neurons in neighbouring layers have a connection, which is represented by a connection weight. The weight between neuron $k$ in the hidden layer $l-1$ and neuron $j$ in hidden layer $l$ is denoted by $w_{jk}^{l}$. No connection exists among neurons in the same layer as well as in the non-neighbouring layers. Data flows through this neural network via connections between neurons, starting from the input layer, through the encoding layer over the hidden layer to the decoding layer and  finally through the output layer. The autoencoder in Figure~\ref{Figure2:Autoencoder} consists of two parts: an encoder mapping: 
\begin{equation}
\textit{f}: \textbf{\textit{h}}=\textit{f}(\textbf{\textit{x}})
\label{encodingmap}
\end{equation}
and a decoding mapping
\begin{equation}
\textit{g}: \textbf{\textit{r}}=\textit{g}(\textbf{\textit{h}})
\label{decodingmap}
\end{equation}
that produces a reconstruction.  The autoencoder defines a mapping 
\begin{equation}
AE: \textit{g}\circ\textit{f}: \textbf{\textit{r}}=\textit{g}(\textit{f} (\textbf{\textit{x}})).
\label{autoencodermap}
\end{equation}
Let $\delta$ be the nonlinear activation function on each layer. There are many choices for the activation function and we will propose an improved version of the $tanh$-activiation function to analyse the mechanical response of Kirchhoff plates. Combined with weight and bias vector defined on each layer, the nonlinear mapping can be written as
\begin{equation}
\textit{f}: \textbf{\textit{h}}=\delta(\boldsymbol{\omega}_1\boldsymbol{x}+\textbf{\textit{b}}_1)
\label{encodmapwithactv}
\end{equation}
Similarly, the decode maps can be written as
\begin{equation}
\textit{g}: \textbf{\textit{r}}=\delta(\boldsymbol{\omega}_2\boldsymbol{h}+\textbf{\textit{b}}_2)
\label{decodmapwithactv}
\end{equation}

Default activation functions available in Pytorch such as the  $tanh$ do not necessarily yield the best results for every model. We will show later in numerical experiments that it results sometimes in unstable results for the deep autoencoder based energy method (DAEM). Therefore, we suggest a modified activation function: 
\begin{equation}
\delta\left(x\right)=tanh\left(\frac{\pi}{2}x\right)
\label{newactivationfun}
\end{equation}

\begin{figure}[H]
\captionsetup{width=0.85\columnwidth}
\centering
\begin{tabular}{c}
\includegraphics[height=8cm,width=9cm]{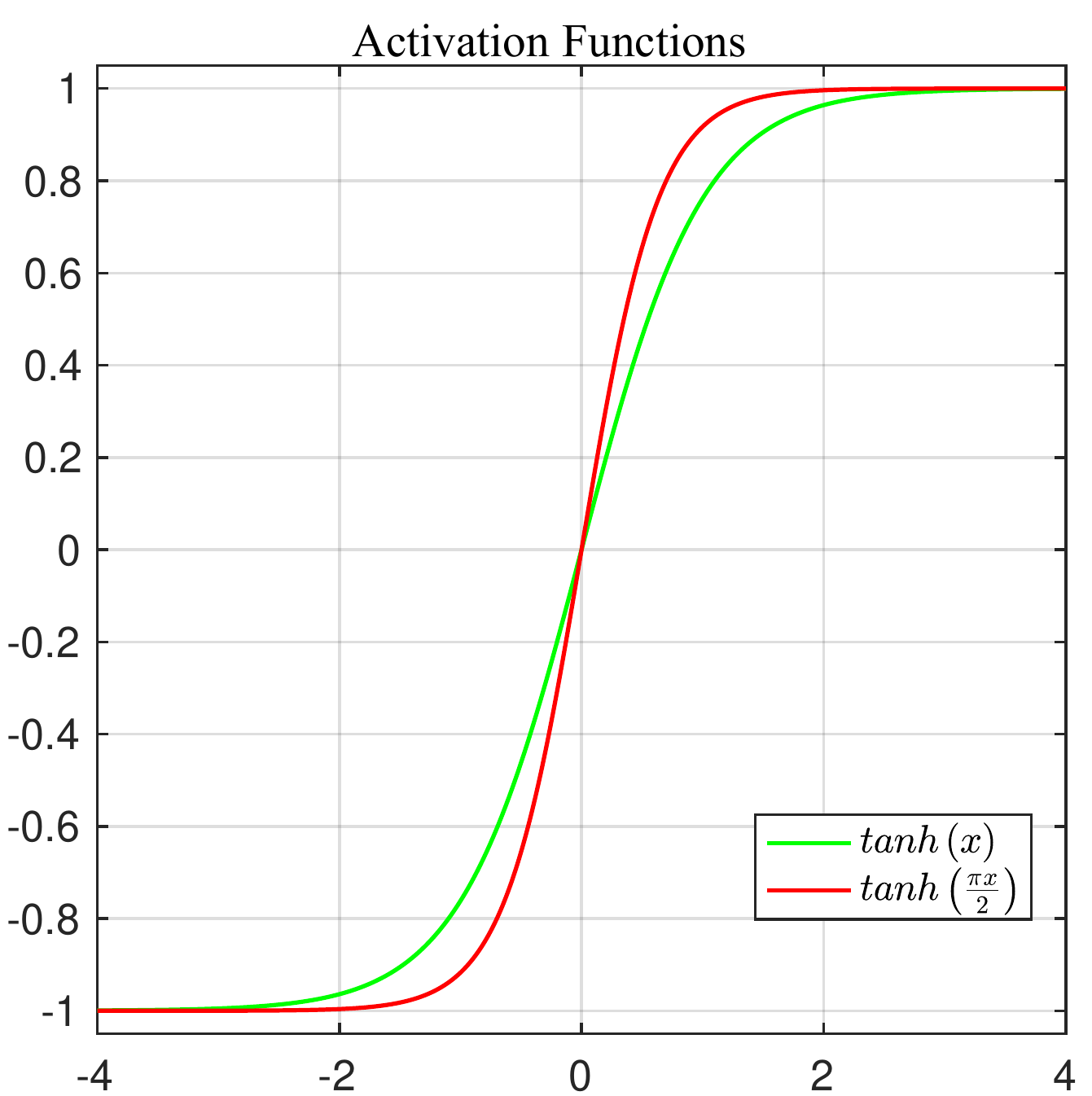}
\end{tabular}
\caption{New activation and original activation in each forms.}
\label{Figure3:activationfun}
\end{figure}

As shown later, this activation function yields better results due to the following reasons: 1.) A larger range of upper values are forced towards $+1$ and a larger range of lower values are close to $-1$, with steeper gradients for the mid-range, which can be seen in Figure \ref{Figure3:activationfun}. Thus, the training is spread uniformly through the feedforward neural network. 2.) For Kirchhoff plate problems, Navier successfully solved those problems with trigonometric Fourier series. Considering that $sin\left(\frac{\pi}{2}x\right),cos\left(\frac{\pi}{2}x\right)$ is periodically changing from $-1$ to $+1$, it can be also adopted as activation function suitable for dynamic analysis. However, periodic activation functions might lead to a "rippling" cost function with bad local minima since a low and high input may produce the same output making the neural network very difficult to train.  The hyperbolic functions, however, satisfy many identities analogous to the trigonometric identities and also map input between $-1$ and $+1$. In this case, the scaled hyperbolic function $tanh\left(\frac{\pi}{2}x\right)$ is preferable, as it yields stable and accurate results and will not slow down the training process. This small change in the activation function has largely improved the Vanishing/Exploding gradient problems in the deep autoencoder neural networks for this model, as will be shown later. The modified new activation has been studied under different layers and different neurons for cases in which Vanishing/Exploding gradient problems are observed  for the $tanh$ activation function. 

Moreover, the number of neurons on each hidden layer and the number of hidden layers can be arbitrarily chosen and are invariably determined through a trial and error procedure or a pruning technique \cite{anwar2017structured}. In the numerical example section, the detailed configuration of this deep autoencoder has been studied and offers an optimum selection of number of hidden layers while keeping the number of neurons on each layer small. An autoencoder thus defines a function $\left(\textit{g}\circ\textit{f}\right)\left(\textit{\textbf{x}};\theta\right)$ depending on the input data $\textit{\textbf{x}}$ and parametrised by $\theta=\left \{ \boldsymbol{\omega}_1,\textbf{\textit{b}}_1,\boldsymbol{\omega}_2,\textbf{\textit{b}}_2\right \}$ consisting of weights and biases in each layer.  It provides an efficient way to approximate unknown field variables and identifies those physical codes behind the model.
\subsection{Basic algorithm for backpropagation}
Like other feedforeward neural networks, deep autoencoders can be trained with all techniques in deep learning such as  minibatch gradient descent method with gradients computed by backpropagation. Backpropagation is an important and computationally efficient mathematical tool to compute gradients in deep learning \cite{nielsen2015neural}. Backpropagation has two main phases, propagation and weight update. The chain rule is recursively applied during the whole process. 

For the deep autoencoder based energy method,  first, the field variable is approximated by a deep autoencoder $\left(\textit{g}\circ\textit{f}\right)\left(\textit{\textbf{x}};\theta\right)$. The components of the linear strain tensor are derivatives of the field variable and can be approximate by a set of deep autoencoder sharing the same hyperparameters. In order to find the hyperparameters of the deep autoencoder including weights and biases, a loss function $\textit{\textbf{L}}\left(\textit{f},w\right)$ is constructed  \cite{Janocha_2017}.  The backpropagation algorithm for the deep autoencoder can be summarised as:
\begin{itemize}
\item \textbf{Input}: Input dataset $x^{1},...,x^{n}$, prepare activation $y^1$ for input layer;
\item \textbf{Feedforward}: For each layer $x^l, l=2,3,...,L$, compute $a^l = \sum_k W^ly^{l-1} + b^l$, and $\sigma \left( a^l \right)$;
\item \textbf{Output error}: Compute the error $\delta^L= \nabla_{y^L} \textit{\textbf{L}} \odot \sigma'_L(a^L)$
\item \textbf{Backpropagation error}: For each $l=L-1,L-2,...,2$, compute $\delta^l = \left((W^{l+1})^T \delta^{l+1}\right) \odot \sigma'_l(a^l)$;
\item \textbf{Output}: The gradient of the loss function is given by $\frac{\partial \textit{\textbf{L}}}{\partial w^l_{jk}} = y^{l-1}_k \delta^l_j$ and $\frac{\partial \textit{\textbf{L}}}{\partial b^l_j} = \delta^l_j$.
\end{itemize}
 Here, $\odot$ denotes the Hadamard product.
 
\section{Deep autoencoder based energy method}
\label{sec4:methodology}
Different from our previous work on deep collocation  \cite{guo2019deep}, the energy method starts from  the principle of minimum energy and  finds the solution by minimizing the total potential energy of this system. The energy method is a weak formulation, which has several advantages over the deep collocation method which is based on the strong form. Firstly, the continuity requirements for the approximating function is reduced in the energy based method requiring less gradient computations compared to the deep collocation method. And secondly, natural boundary condition are automatically satisfied in the energy method, which is especially helpful for fourth order problems.
\subsection{Energy method for bending analysis}
Let us consider Kirchhoff plate bending problems in the context of the DAEM. Recalling Equation \ref{totalenergy} is the total potential energy, the entire problem can be boiled down to minimizing the total potential energy enforcing essential boundary conditions. The transversal deflection $w$ is approximated with the aforementioned deep autoencoder $\left(\textit{g}\circ\textit{f}\right)\left(\textit{\textbf{x}};\theta\right)$ denoted by $\textit{w}^h \left(\textit{\textbf{x}};\theta\right)$. A loss function can thus be constructed to find the approximate solution by minimizing of total potential energy with essential boundary conditions approximated by $\textit{w}^h \left(\textit{\textbf{x}};\theta\right)$. Substituting $\textit{w}^h \left(\textit{\textbf{x}}\,_\Omega;\theta\right)$ into Equation \ref{totalenergy}, Equation \ref{strainenergy}, and Equation \ref{externalwork}, results in
\begin{equation}
\Pi\left(\textit{\textbf{x}}\,_\Omega;\theta\right) = U\left(\textit{\textbf{x}}\,_\Omega;\theta\right)+W_{ext}\left(\textit{\textbf{x}}\,_\Omega;\theta\right).
\label{pi_totalenergy}
\end{equation}	
where 
\begin{equation}
\begin{aligned}
U\left(\textit{\textbf{x}}\,_\Omega;\theta\right)=\frac{1}{2} \mathit{ \mathit{D_{0}}} \int_{\Omega } [ ( k_x\left(\textit{\textbf{x}}\,_\Omega;\theta\right)+k_y\left(\textit{\textbf{x}}\,_\Omega;\theta\right) )^2 \\ +2\left ( 1-\nu  \right )\left ( k_{xy}\left(\textit{\textbf{x}}\,_\Omega;\theta\right)^2-k_x\left(\textit{\textbf{x}}\,_\Omega;\theta\right)k_y \left(\textit{\textbf{x}}\,_\Omega;\theta\right) \right )]d\Omega.
\end{aligned}
\label{pi_strainenergy}
\end{equation}	
and 
\begin{equation}
\begin{aligned}
W_{ext}\left(\textit{\textbf{x}}\,_\Omega;\theta\right)=-[ \int_{\Omega } p\left ( \textit{\textbf{x}}\,_\Omega\right )\textit{w}^h\left(\textit{\textbf{x}}\,_\Omega;\theta\right) d\Omega \\+\int _{\Gamma_{3}}\tilde{q}\textit{w}^h\left(\textit{\textbf{x}}\,_{\Gamma_3};\theta\right)d\Gamma -\int _{\Gamma_{M}}\tilde{M_n}\frac{\partial \textit{w}^h\left(\textit{\textbf{x}}\,_{\Gamma_M};\theta\right)}{\partial n}d\Gamma ],
\end{aligned}
\label{pi_externalwork}
\end{equation} with $\Gamma_{M}=\Gamma_{2}+\Gamma_{3}$,
which yields an energy driven deep neural network $\Pi \left(\textit{\textbf{x}}\,_\Omega;\theta\right)$. 
Moreover, $k_x\left(\textit{\textbf{x}}\,_\Omega;\theta\right)$, $k_y\left(\textit{\textbf{x}}\,_\Omega;\theta\right)$, and $k_{xy}\left(\textit{\textbf{x}}\,_\Omega;\theta\right)$ can be obtained by substituting $\textit{w}^h \left(\textit{\textbf{x}}\,_\Omega;\theta\right)$ into Equation \ref{generalizedstrain}. The boundary conditions from Section 2 can also be learnt by the neural network approximation $\textit{w}^h \left(\textit{\textbf{x}}\,_\Gamma;\theta\right)$:
\noindent On $\Gamma_{1}$, we have
\begin{equation}
  \textit{w}^h \left(\textit{\textbf{x}}\,_{\Gamma_1};\theta\right)=\tilde{w}, \  \frac{\partial \textit{w}^h \left(\textit{\textbf{x}}\,_{\Gamma_1};\theta\right)}{\partial n} = \tilde{ \theta}_{n}.
\end{equation} 

\noindent On $\Gamma_{2}$,
\begin{equation}
  \textit{w}^h \left(\textit{\textbf{x}}\,_{\Gamma_2};\theta\right)=\tilde{w}, \ \tilde{ M}_{n}\left(\textit{\textbf{x}}\,_{\Gamma_2};\theta\right)=\tilde{ M}_{n}, 
\label{bd2}
\end{equation} 
where $\tilde{ M}_{n}\left(\textit{\textbf{x}}\,_{\Gamma_2};\theta\right)$ can be obtained from Equation \ref{moment} by combing $\textit{w}^h \left(\textit{\textbf{x}}\,_{\Gamma_2};\theta\right)$.

\noindent On $\Gamma_{3}$,
\begin{equation}
  M_{n}\left(\textit{\textbf{x}}\,_{\Gamma_3};\theta\right) =\tilde{ M}_{n} , \  \frac{\partial M_{ns}\left(\textit{\textbf{x}}\,_{\Gamma_3};\theta\right) }{\partial s}+Q_{n}\left(\textit{\textbf{x}}\,_{\Gamma_3};\theta\right)=\tilde{q}, 
\end{equation}  
where $M_{ns}\left(\textit{\textbf{x}}\,_{\Gamma_3};\theta\right)$ can be obtained from Equation \ref{moment} and $Q_{n}\left(\textit{\textbf{x}}\,_{\Gamma_3};\theta\right)$ can be obtained from Equation \ref{shearforce} by combing $\textit{w}^h \left(\textit{\textbf{x}}\,_{\Gamma_3};\theta\right)$. Note that $\textit{\textbf{n}},\textit{\textbf{s}}$ refer to the normal and tangent directions along the boundaries.  The induced energy driven neural network $\mathit{\Pi}\left(\textit{\textbf{x}};\theta\right)$ shares the same parameters as $\textit{w}^h \left(\textit{\textbf{x}};\theta\right)$. 

Finally, we construction the loss function for the proposed DAEM, which minimizes the total potential energy subjected to essential boundary conditions:
\begin{equation}
L\left(\theta\right)=\Pi+MSE_{\Gamma_{w}}+MSE_{\Gamma_{\theta_n}},
\label{lossfunBnd}
\end{equation}
with
\begin{equation}
\begin{aligned}
&\Pi=\Pi\left(\textit{\textbf{x}}\,_\Omega;\theta\right),\\
&MSE_{\Gamma_{w}}=\frac{1}{N_{\Gamma_{w}}}\sum_{i=1}^{N_{\Gamma_{w}}}\begin{Vmatrix}
 \textit{w}^h \left(\textit{\textbf{x}}\,_{\Gamma_{w}};\theta\right)-\tilde{w}
\end{Vmatrix}^2,\\
&MSE_{\Gamma_{\theta_{n}}}=\frac{1}{N_{\Gamma_{\theta_{n}}}}\sum_{i=1}^{N_{\Gamma_{\theta_{n}}}}\begin{Vmatrix}
\frac{\partial \textit{w}^h \left(\textit{\textbf{x}}\,_{\Gamma_{\theta_{n}}};\theta\right)}{\partial n} - \tilde{ \theta}_{n}
\end{Vmatrix}^2,\\
\end{aligned}
\end{equation}
where $\Gamma_{w}=\Gamma_{1}+\Gamma_{2}$, $\Gamma_{\theta_{n}}=\Gamma_{1}$; $x\,_\Omega \in {R^N} $, $\theta \in {R^K}$ are the neural network parameters and $L\left(\theta\right)=0$, $\textit{w}^h \left(\textit{\textbf{x}};\theta\right)$ is a solution to transversal deflection.

Note that the proposed DAEM requires a method to evaluate the integrals and also the corresponding quadrature points are deployed as input datasets. We could adopt 'traditional' multivariate numerical quadrature methods such as Gaussian quadrature. A background mesh could therefore be constructed. However, minimizing the total potential energy at those fixed points might results in underfitting. This issue can be avoided by random sampling and therefore, the Monte Carlo integration method \cite{caflisch_1998} is employed in DAEM for the integral calculation. For the two dimensional Monte-Carlo integration method, let us consider the integral $\int_{\Omega } p\left ( x,y \right )\\ \textit{w}^h\left(\textit{\textbf{x}}\,_\Omega;\theta\right) d\Omega$ in Equation \ref{pi_externalwork}, which can be evaluated by
\begin{equation}
\int_{\Omega } p\left ( \textit{\textbf{x}}\,_\Omega\right )\textit{w}^h\left(\textit{\textbf{x}}\,_\Omega;\theta\right) d\Omega=\frac{A}{N_{\Omega}}\sum_{i=1}^{N_{\Omega}}p\left ( \textit{\textbf{x}}\,_{i,\Omega}\right )\textit{w}^h\left(\textit{\textbf{x}}\,_{i,\Omega};\theta\right)
\label{montecarlo}
\end{equation}
$\textit{\textbf{x}}_{i,\Omega}$ denoting the dataset generated by the ramdom sampling in the physical domain and $A$ is the area of the middle surface and $N_{\Omega}$ is the number of random distributed points inside the physical domain.
\subsection{Energy method for vibration and buckling analysis}
The loss function for the vibration and buckling analysis has to be modified. The key objective is to obtain the fundamental natural frequency and critical buckling, respectively. Recalling Equation \ref{Rayleighquotient}, the Rayleigh quotient is defined and derived from Rayleigh's principle and the lowest natural frequency can be retrieved from the minimization of Equation \ref{Rayleighquotient} accounting for essential boundary conditions. Also the mode shape function is approximated by the deep autoencoder $\textit{W}^h \left(\textit{\textbf{x}};\theta\right)$. Accordingly, the loss function can be defined as:
\begin{equation}
L\left(\theta\right)=\frac{2U_{max}\left(\textit{\textbf{x}}\,_\Omega;\theta\right)}{\int_\Omega\rho hW^2\left(\textit{\textbf{x}}\,_\Omega;\theta\right)d\Omega}+MSE_{\Gamma_{W}}+MSE_{\Gamma_{\theta_n}},
\end{equation}
with
\begin{equation}
\begin{aligned}
&MSE_{\Gamma_{W}}=\frac{1}{N_{\Gamma_{W}}}\sum_{i=1}^{N_{\Gamma_{W}}}\begin{Vmatrix}
 \textit{W}^h \left(\textit{\textbf{x}}\,_{\Gamma_{W}};\theta\right)-\tilde{W}
\end{Vmatrix}^2,\\
&MSE_{\Gamma_{\theta_{n}}}=\frac{1}{N_{\Gamma_{\theta_{n}}}}\sum_{i=1}^{N_{\Gamma_{\theta_{n}}}}\begin{Vmatrix}
\frac{\partial \textit{W}^h \left(\textit{\textbf{x}}\,_{\Gamma_{\theta_{n}}};\theta\right)}{\partial n} - \tilde{ \theta}_{n}
\end{Vmatrix}^2,\\
\end{aligned}
\end{equation}
where $\Gamma_{w}=\Gamma_{1}+\Gamma_{2}$, $\Gamma_{\theta_{n}}=\Gamma_{1}$; $x\,_\Omega \in {R^N} $, $\theta \in {R^K}$ are the neural network parameters. The Monte-Carlo quadrature rule is adopted for calculating the Rayleigh quotient.  However, some modification to the loss function is needed to ensure  $\textit{w}^h \left(\textit{\textbf{x}};\theta\right)$ is a nontrivial solution. Therefore, we normalize the mode shape function and ensure the inner product of the mode shape function is unity. This leads to the modified loss function
\begin{equation}
L\left(\theta\right)=\frac{2U_{max}\left(\textit{\textbf{x}}\,_\Omega;\theta\right)}{\int_\Omega\rho hW^2\left(\textit{\textbf{x}}\,_\Omega;\theta\right)d\Omega}+k_p(\int_\Omega W^2\left(\textit{\textbf{x}}\,_\Omega;\theta\right) d\Omega-1)^2+MSE_{\Gamma_{W}}+MSE_{\Gamma_{\theta_n}},
\label{lossfunVib}
\end{equation}
where $k_p$ is a penalty factor. A factor between 1 to 100 already yields good numerical results. The loss function for the buckling analysis can be written as
\begin{equation}
L\left(\theta\right)=\lambda\left(\textit{\textbf{x}}\,_\Omega;\theta\right) +k_p(\int_\Omega (\textit{w}^h\left(\textit{\textbf{x}}\,_\Omega;\theta\right))^2d\Omega-1)^2+MSE_{\Gamma_{W}}+MSE_{\Gamma_{\theta_n}}.
\label{lossfunBcl}
\end{equation}
where $\lambda$ is the load factor. From Equation \ref{criticalbucklingload}, we obtain
\begin{equation}
\begin{aligned}
\lambda\left(\textit{\textbf{x}}\,_\Omega;\theta\right)  = \frac{-\frac{\mathit{ \mathit{D_{0}}} }{2}  \int_{\Omega } \left [ ( \nabla^2\textit{w}^h)^2+2\left ( 1-\nu  \right )\left ( \left ( \frac{\partial^2\textit{w}^h}{\partial xy} \right )^2-\frac{\partial^2 \textit{w}^h}{\partial x^2} \frac{\partial^2 \textit{w}^h}{\partial y^2}\right )\right ]d\Omega }{\frac{1}{2}  \int_{\Omega } \left [ N_x\left ( \frac{\partial \textit{w}^h}{\partial x} \right ) ^2+N_y\left ( \frac{\partial \textit{w}^h}{\partial y} \right ) ^2+2{N}_{xy} \frac{\partial \textit{w}^h}{\partial x} \frac{\partial \textit{w}^h}{\partial y}\right ]d\Omega}
\end{aligned}
\end{equation}
where $\textit{w}^h$ is the approximation of the transversal deflection by the deep autoencoder, i.e. $\textit{w}^h\left(\textit{\textbf{x}}\,_\Omega;\theta\right)$.  Now, we can find the set of parameters $\theta$ such that the approximated deflection $\textit{w}^h \left(\textit{\textbf{x}};\theta\right)$ minimizes the loss $L\left(\theta\right)$, i.e.
\begin{equation}
\textit{w}^h = \argmin_{\theta \in R^K} L\left(\theta\right)
\end{equation}
These hyperparameters are obtained by backpropogation as mentioned before. The L-BFGS \cite{liu1989limited}  optimizer with backpropogation is adopted to tune those hyperparameters of the deep autoencoder with few restrictions. Thus, the solution to thin plate bending, vibration and buckling problems by deep autoencoder based energy method can be reduced to an optimization problem. The general procedure of the proposed DAEM can be summarized as follows:

{\centering
\begin{algorithm}[H]
\DontPrintSemicolon
  
  \KwInput{Create Random Sampling Points $\textbf{\textit{x}}_\Omega$ inside the physical domain and $\textbf{\textit{x}}_\Gamma$ on the boundaries.}
  \KwOutput{The predicted field variables}
  \KwData{Testing data set: $\boldsymbol{\mathit{x^\ast}}$}
  
  Fix the number of neurons on input layer $D_{in}$, hidden layers $H$, encoding layers $iH$, $i \in \left \{ 1,\dots,\mathbb{N_+}  \right \}$  , and output layer $D_{out}$; Choosing activation function $\delta$ and proper optimizer; Fixing the number of hidden layers $N_{hl}$  and number of encoding layers $N_{edl}$; Training iteration $N_{iter}$.

\textbf{Model Training}:
  
  \For{i\enspace from 1 to $N_{iter}$}
    {
\begin{enumerate}[rightmargin=\dimexpr\linewidth-10cm-\leftmargin\relax,label=(\Roman*)]
\item calculate activation function $\textbf{\textit{h}}_i$ on hidden layers in Equation \ref{encodmapwithactv}.
  \vspace{-0.2cm}
  \item calculate the reconstructed output $\textbf{\textit{r}}_i$ from $\textbf{\textit{h}}_i$ in Equation \ref{decodmapwithactv}.
  \vspace{-0.2cm} 
  \item choose and compute the loss function from bending loss function Equation \ref{lossfunBnd}, the vibration loss function Equation \ref{lossfunVib} or the buckling loss function Equation \ref{lossfunBcl}.
  \vspace{-0.2cm} 
  \item  back-propogate error gradient and update weights and bias
\end{enumerate} 
  \vspace{-0.2cm} 
   }

\textbf{Inference}:
  Make predictions and inference based on the trained deep autoencoder.
\caption{Procedure for deep encoder based energy method}
\end{algorithm}
\par}

\section{Numerical Experiments}
\label{sec5:example}
In this section, we demonstrate the performance of DAEM for several numerical examples for plate bending, vibration and buckling analysis. The simulations are done on a 64-bit macOS Mojave server with Intel(R) Core(TM) i7-8850H CPU, 32GB memory. We found that a deep neural structure with less width is preferred over a shallow structures. Hence,  we mainly show results for increasing number of  hidden layers  rather than for increasing number of neurons. The accuracy of the numerical results by using the relative error of maximum deflection and deflection over the whole plate. The relative error is defined as:
\begin{equation}
e=\frac{\| w_{predict}-w_{analytical}\|}{\|w_{analytical}\|}
\end{equation} 
Here, $\|\cdot\|$ refers to the $l^2-norm$.

\subsection{Bending analysis}
We study three benchmark problems including a plate with hole and a plate on the elastic foundation, which can be compared to an  analytical solution. 
\subsection{Square plate under a sinusoidally distributed load}
\label{SSPLBD}
Let us consider a simply-supported square plate under a sinusoidal distribution transverse loading. The sinusoidal distributed load is expressed by
\begin{equation}
\begin{array}{l}
p=\frac{p_{0}}{D}\textrm{sin}\left (\frac{\pi x}{a}  \right )\textrm{sin}\left (\frac{\pi y}{b}  \right ).
\end{array}
\end{equation} 
where $a$,$b$ indicate the length of the plate. The analytical solution for this problem is given by \cite{timoshenko1959theory}:
\begin{equation}
\begin{array}{l}
w=\frac{p_{0}}{\pi^{4} D\left (\frac{1}{a^{2}}+\frac{1}{b^{2}}  \right )^{2}}\textrm{sin}\left (\frac{\pi x}{a}  \right )\textrm{sin}\left (\frac{\pi y}{b}  \right ).
\end{array}
\end{equation}  

\begin{figure}[H]
\captionsetup{width=0.85\columnwidth}
\centering
\begin{subfigure}[b]{6.5cm}
  \centering\includegraphics[height=6cm,width=6.5cm]{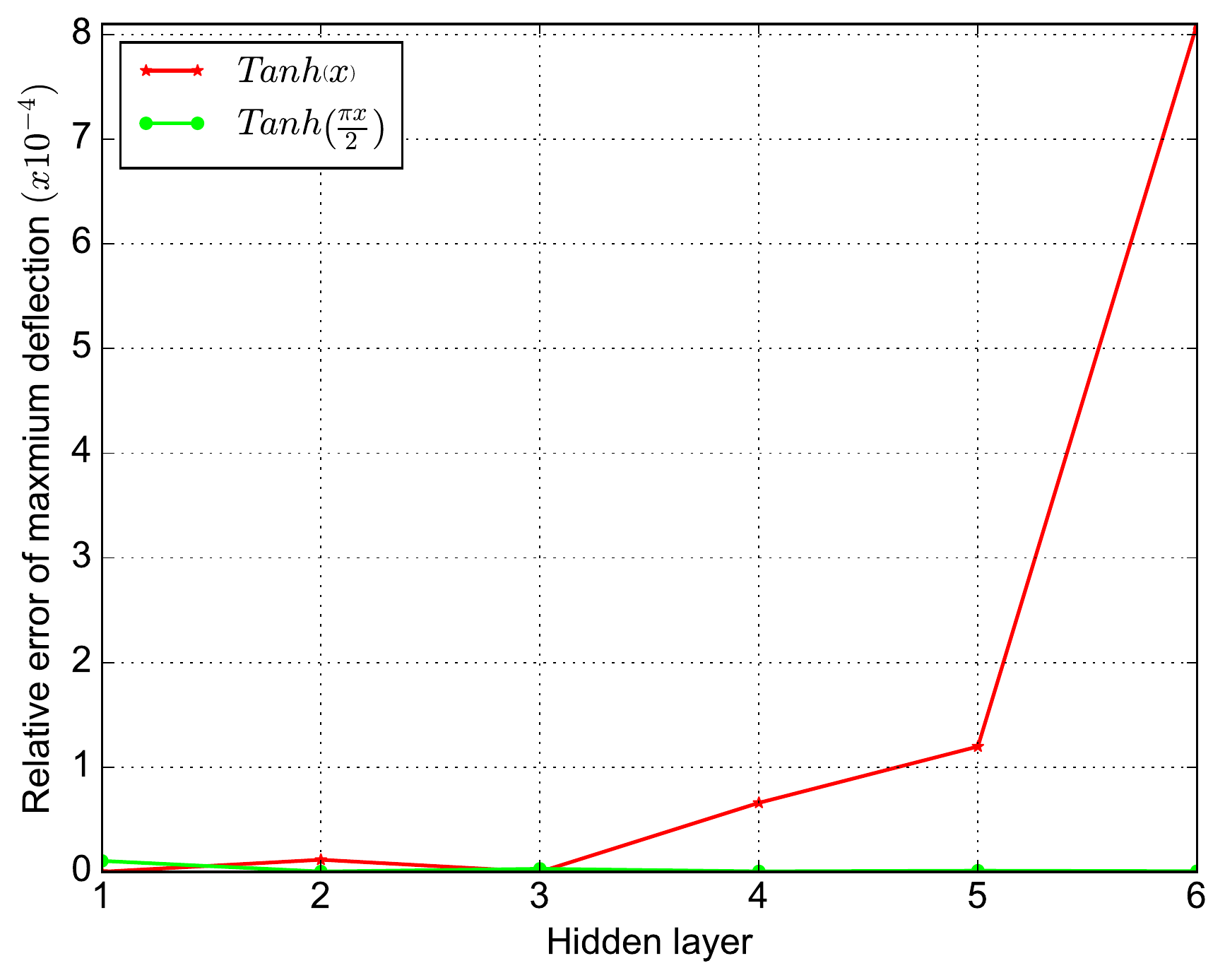}   
   \caption{}\label{}
 \end{subfigure}%
 \begin{subfigure}[b]{6.5cm}
 \centering\includegraphics[height=6cm,width=6.5cm]{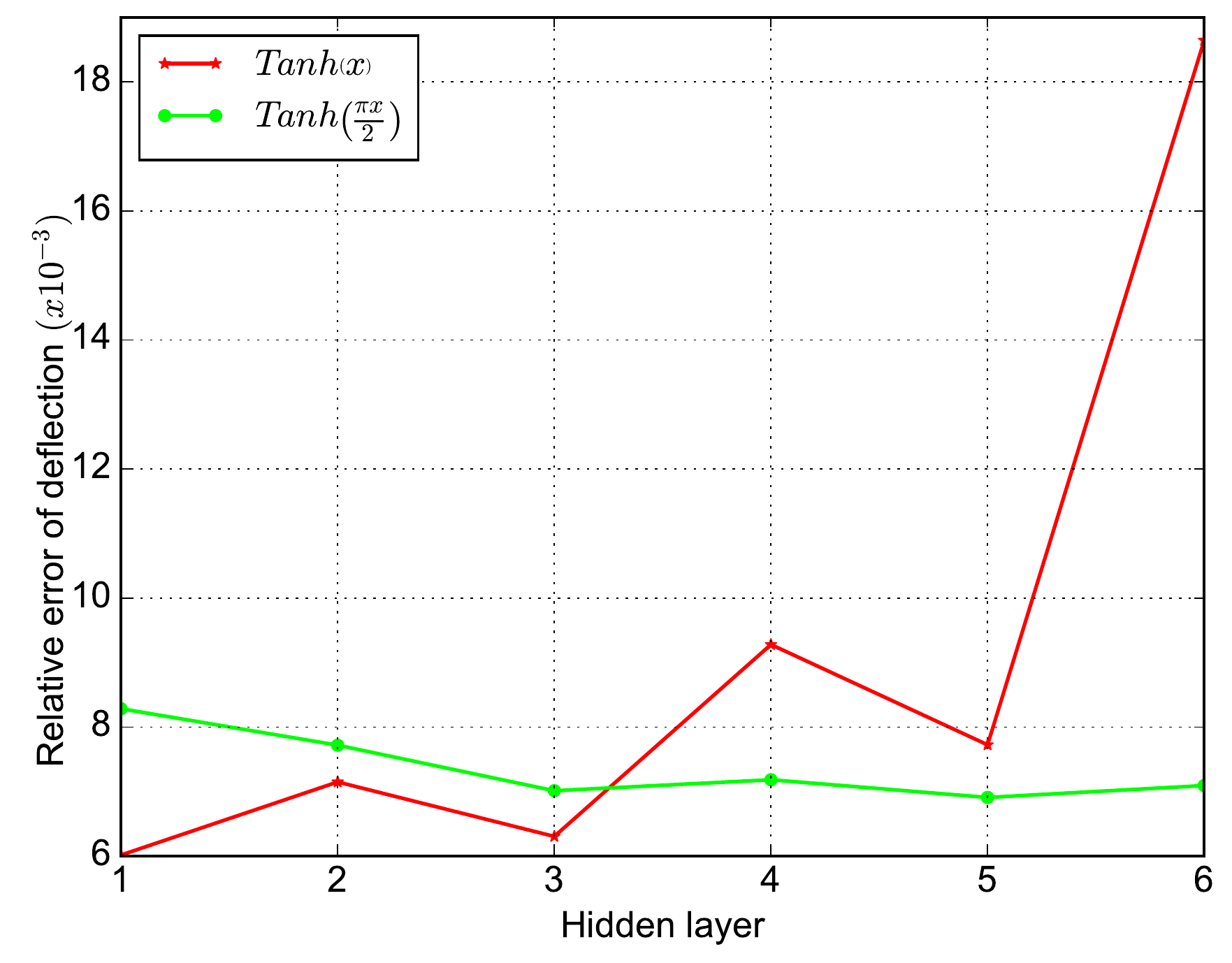}   
   \caption{}\label{}
 \end{subfigure}%
 \caption{Relative error of $\left(a\right)$ maximum central deflection and $\left(b\right)$ all deflection predicted by Tanh and proposed activation function with DNN}
\label{RelativerrDNNAcFun}
\end{figure}

We first study the accuracy and efficiency of the proposed activation function. In general, the test is performed with a deep feedforward neural network with 10 neurons per hidden layer. The relative error of the maximum deflection at the central plate and the deflection over the whole plate vs the increasing of hidden layers are shown in Figure~\ref{RelativerrDNNAcFun}. The modified hyperbolic tangent activation function $tanh\left(\frac{\pi}{2}x\right)$ is less dependent on the number of hidden layers than the original hyperbolic tangent activation function.

\begin{figure}[H]
\captionsetup{width=0.85\columnwidth}
\centering
\includegraphics[height=7cm,width=10cm]{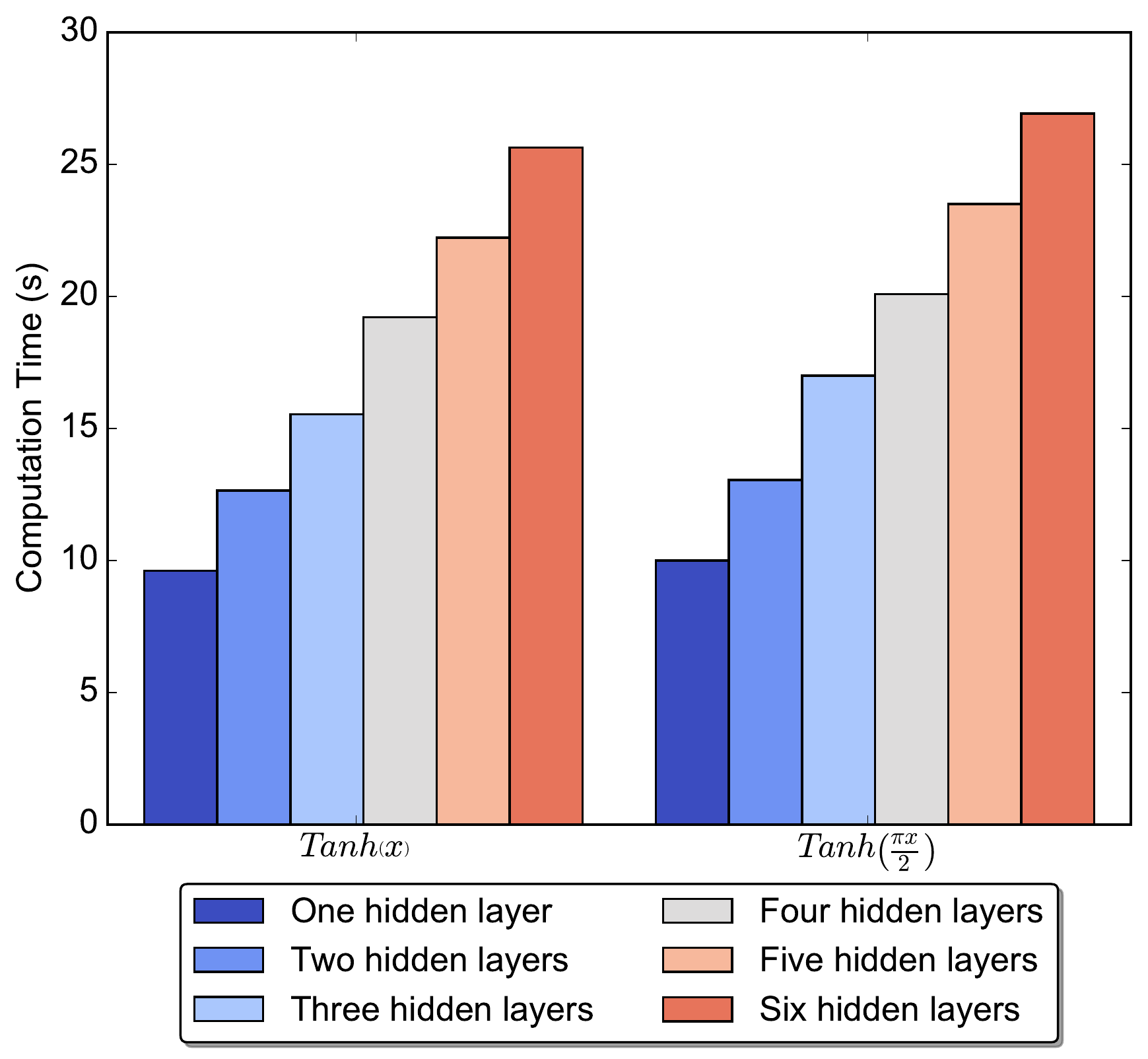}
\caption{The computational cost of two activation functions with increasing hidden layers.}
\label{ComputationDNNCODConf}
\end{figure}

The computational cost of those two schemes is shown in Figure~\ref{ComputationDNNCODConf} and is comparable. The relative error of the deflection is shown in Table~\ref{tab:Table1}. While the gradient explodes for some problems -- indicated by NaN (not a real number) -- for the original $Tanh$ activation function, the modified $tanh$ activation function always yields stable results. The encoding configuration refers here to the hidden layers and neuron numbers specified for the encoder. For the decoder a symmetric configuration is adopted. 

\begin{table}[H] 
\captionsetup{width=0.85\columnwidth}
\caption{The Relative Error of Deflection for DAEM with different activation functions} 
\vspace{-0.3cm}
\centering 
\begin{tabular}{l c c} 
\toprule 
\toprule 
& \multicolumn{2}{c}{\textbf{Relative Error of Deflection}} \\ 
\cmidrule(l){2-3} 
\textbf{Encoder Configuration} & $Tanh\left(\textbf{\textit{x}}\right)$ & $Tanh\left(\frac{\pi}{2}\textbf{\textit{x}}\right)$\\ 
\midrule 
Encoding layer, [30]       & 0,0063740 & 0,0060618 \\ 
Encoding layer, [40]       & NaN & 0,0058869 \\ 
Encoding layer, [50]       & 0,0130702 & 0,0090355 \\ 
Encoding layer, [60]       & 0,0077438 & 0,0104713 \\ 
Encoding layers, [30,10]    & 0,0083107 & 0,0070922 \\ 
Encoding layers, [40,10]    & 0,0082238 & 0,0066074 \\ 
Encoding layers, [50,10]    & 0,0061002 & 0,0082771 \\ 
Encoding layers, [60,10]    & 0,0075465 & 0,0058044 \\ 
Encoding layers, [30,20]    & 0,0075115 & 0,0078926 \\ 
Encoding layers, [40,20]    & 0,0087727 & 0,0086784 \\ 
Encoding layers, [50,20]    & NaN & 0,0055310 \\ 
Encoding layers, [60,20]    & NaN & 0,0084229 \\ 
Encoding layers, [30,20,10] & 0,0083553 & 0,0069302 \\ 
Encoding layers, [50,30,10] & 0,0121318 & 0,0077830 \\ 
Encoding layers, [60,30,10] & 0,0102968 & 0,0053692 \\ 
Encoding layers, [40,30,20] & 0,0075632 & 0,0075470 \\ 
Encoding layers, [50,30,20] & 0,0067328 & 0,0065867 \\ 
Encoding layers, [60,30,20] & NaN & 0,0060054 \\ 

\bottomrule 
\end{tabular}
\label{tab:Table1} 
\end{table}

\begin{figure}[H]
\captionsetup{width=0.85\columnwidth}
\centering
\includegraphics[height=7cm,width=10cm]{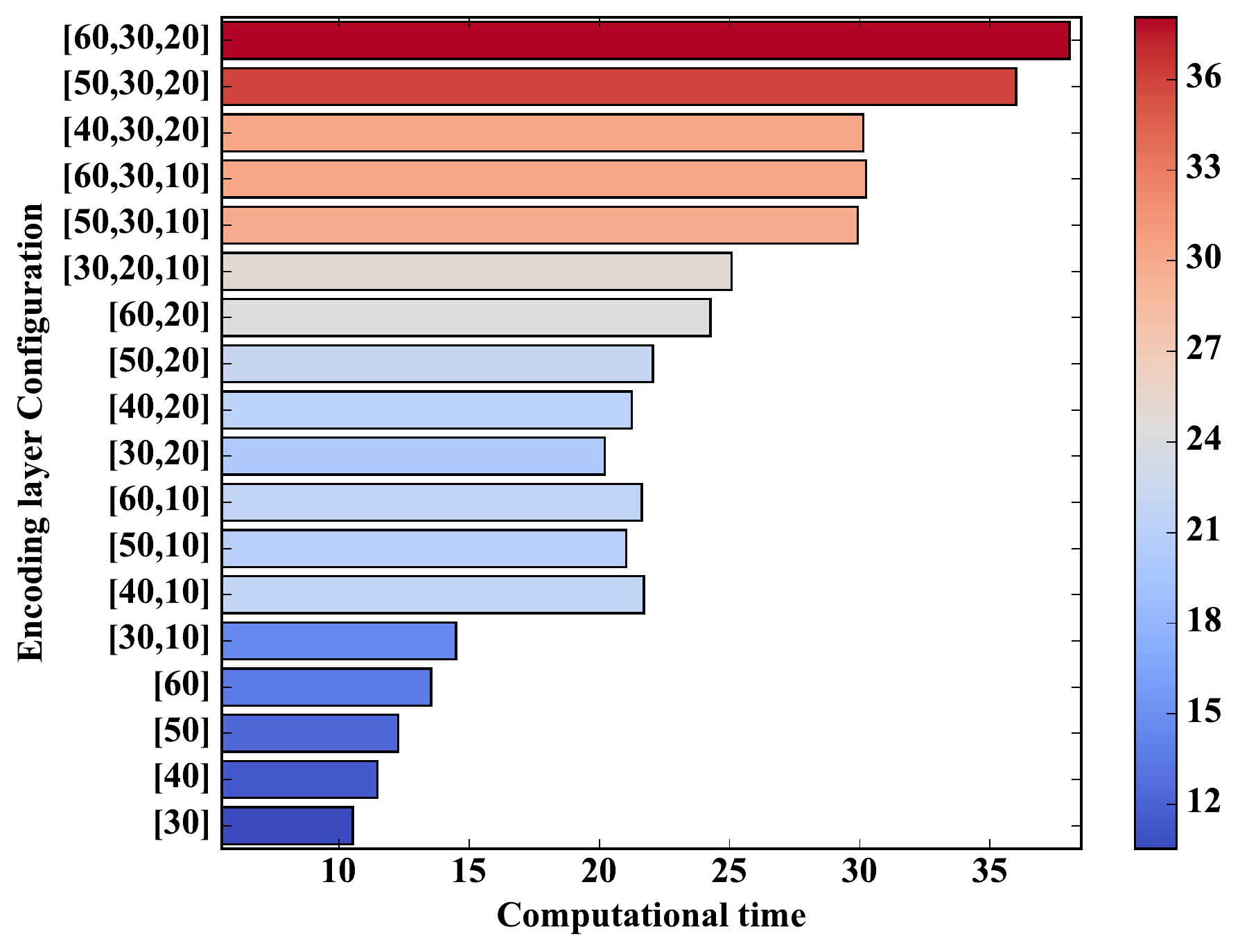}
\caption{The computational time for DAEM with different encoding cconfigurations.}
\label{ComputationDAEMCmptmCODConf}
\end{figure}

Next, we study the recommended deep autoencoder configuration by comparing various encoders with varying layers and neurons per layer. As shown in Figure~\ref{ComputationDAEMCmptmCODConf}, an increasing number of encoding layers results -- as expected -- in increased computational cost. 

\begin{figure}[H]
\captionsetup{width=0.85\columnwidth}
\centering
\includegraphics[height=7cm,width=10cm]{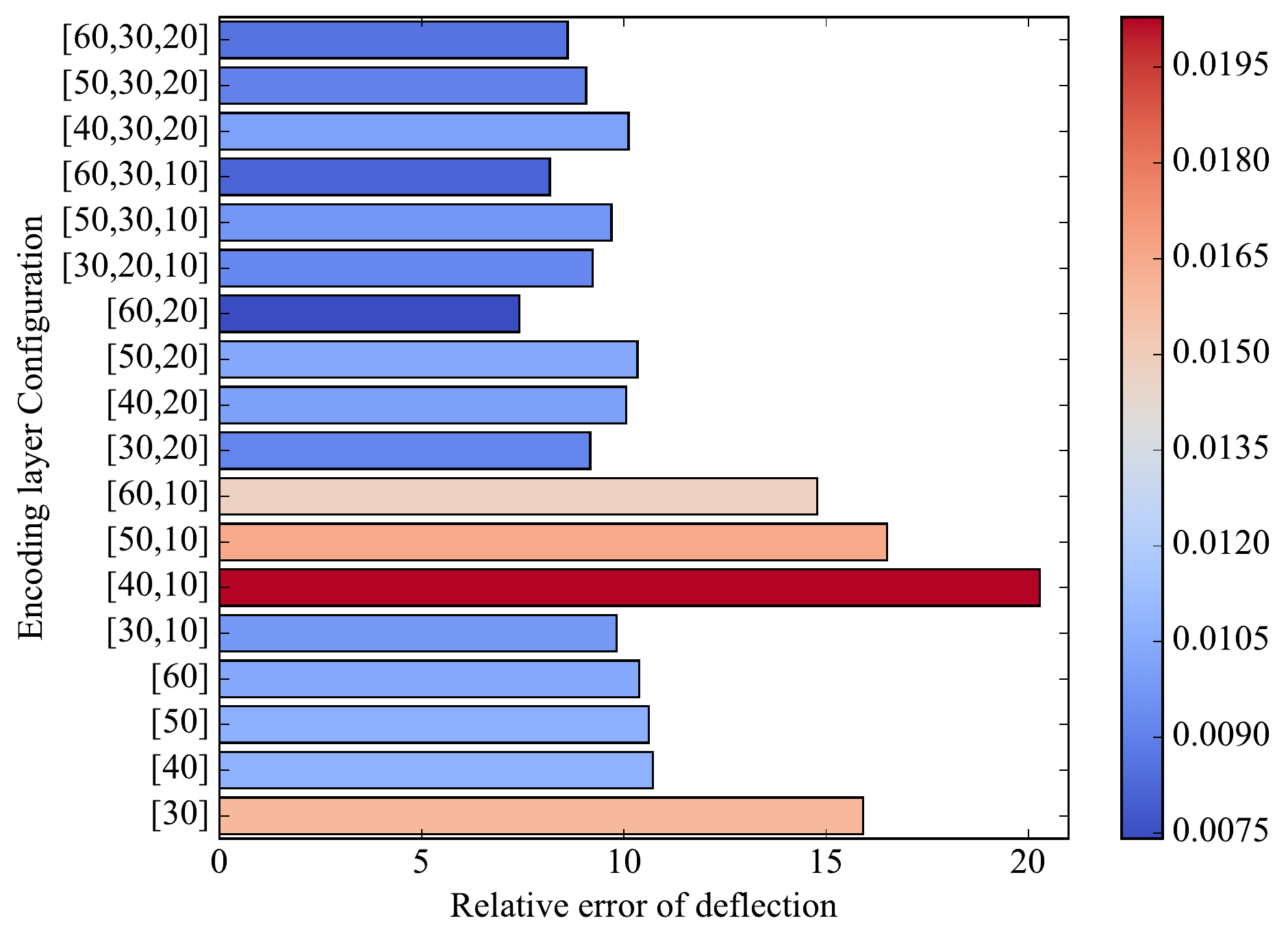}
\caption{The relative error of deflection for DAEM with different encoding cconfigurations.}
\label{ComputationDAEMRerrCODConf}
\end{figure}

 The relative errors obtained by different encoder schemes are shown in Figure~\ref{ComputationDAEMRerrCODConf}. The results converge to the analytical solution with increasing number of  layers. However, more neurons do not necessarily improve the accuracy, especially for encoder $[60]$. Moreover, the results are already quite accurate with only one encoding layer.  In summary, these numerical experiments suggest the change of the DAEM layer configuration rather than increasing the width of the deep neural network.

\begin{figure}[H]
\captionsetup{width=0.85\columnwidth}
\centering
\includegraphics[height=7cm,width=10cm]{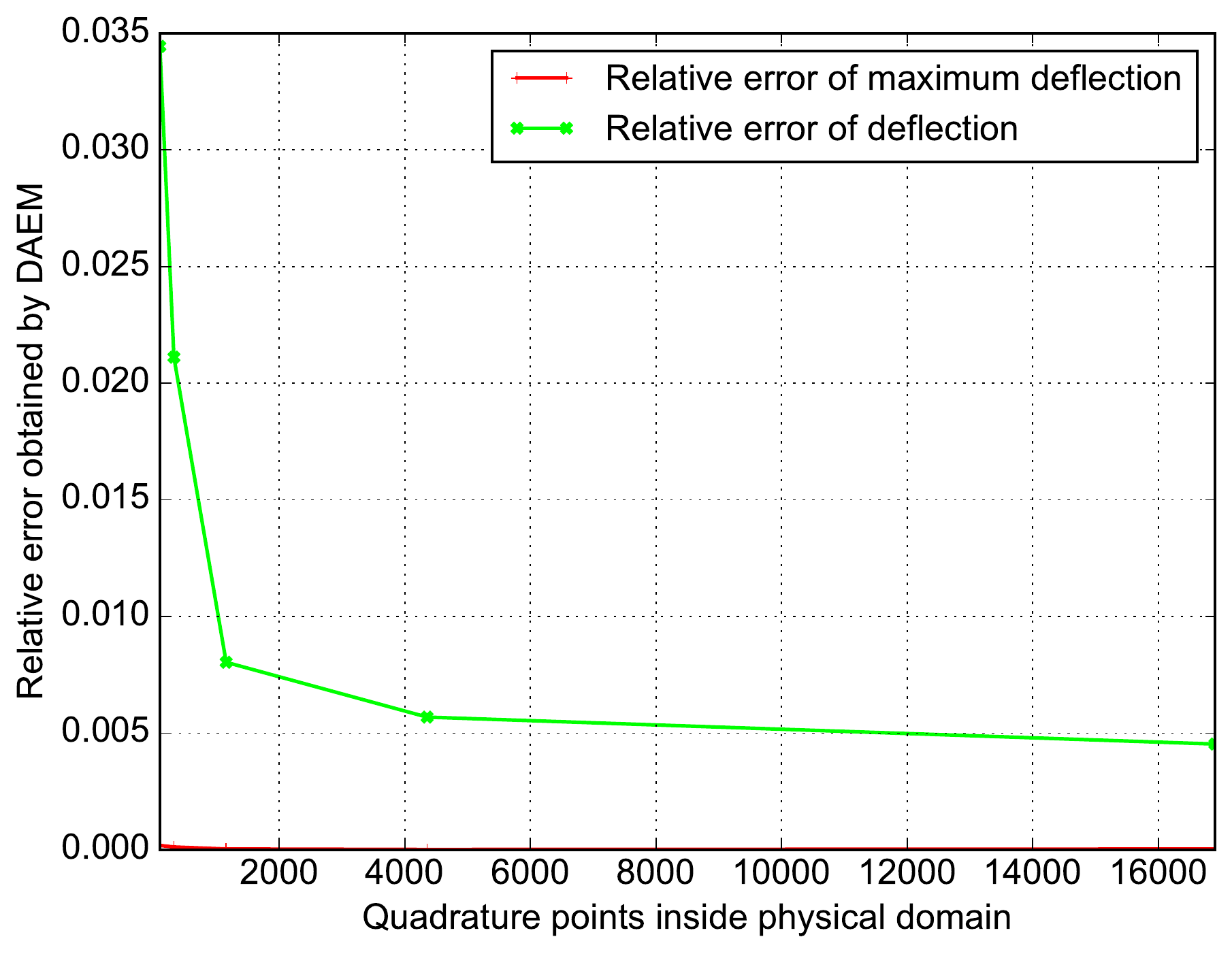}
\caption{The relative error of deflection for DAEM with increasing quadrature points.}
\label{DAEMMaxErrErr}
\end{figure}

\begin{figure}[H]
\captionsetup{width=0.85\columnwidth}
\centering
\begin{subfigure}[b]{6.5cm}
  \centering\includegraphics[height=6cm,width=6.5cm]{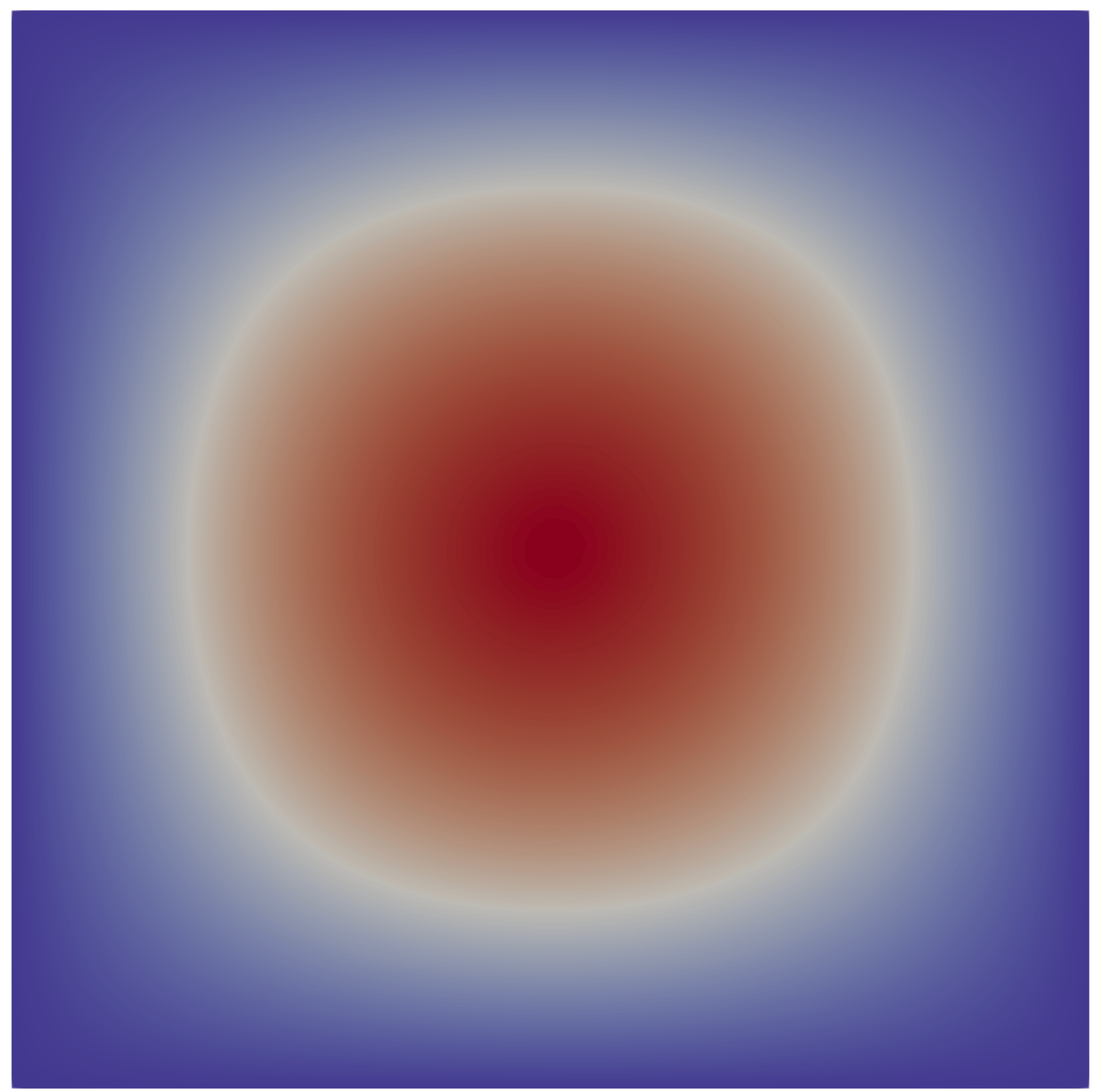}   
   \caption{}\label{}
 \end{subfigure}%
 \begin{subfigure}[b]{6.5cm}
 \centering\includegraphics[height=6cm,width=6.5cm]{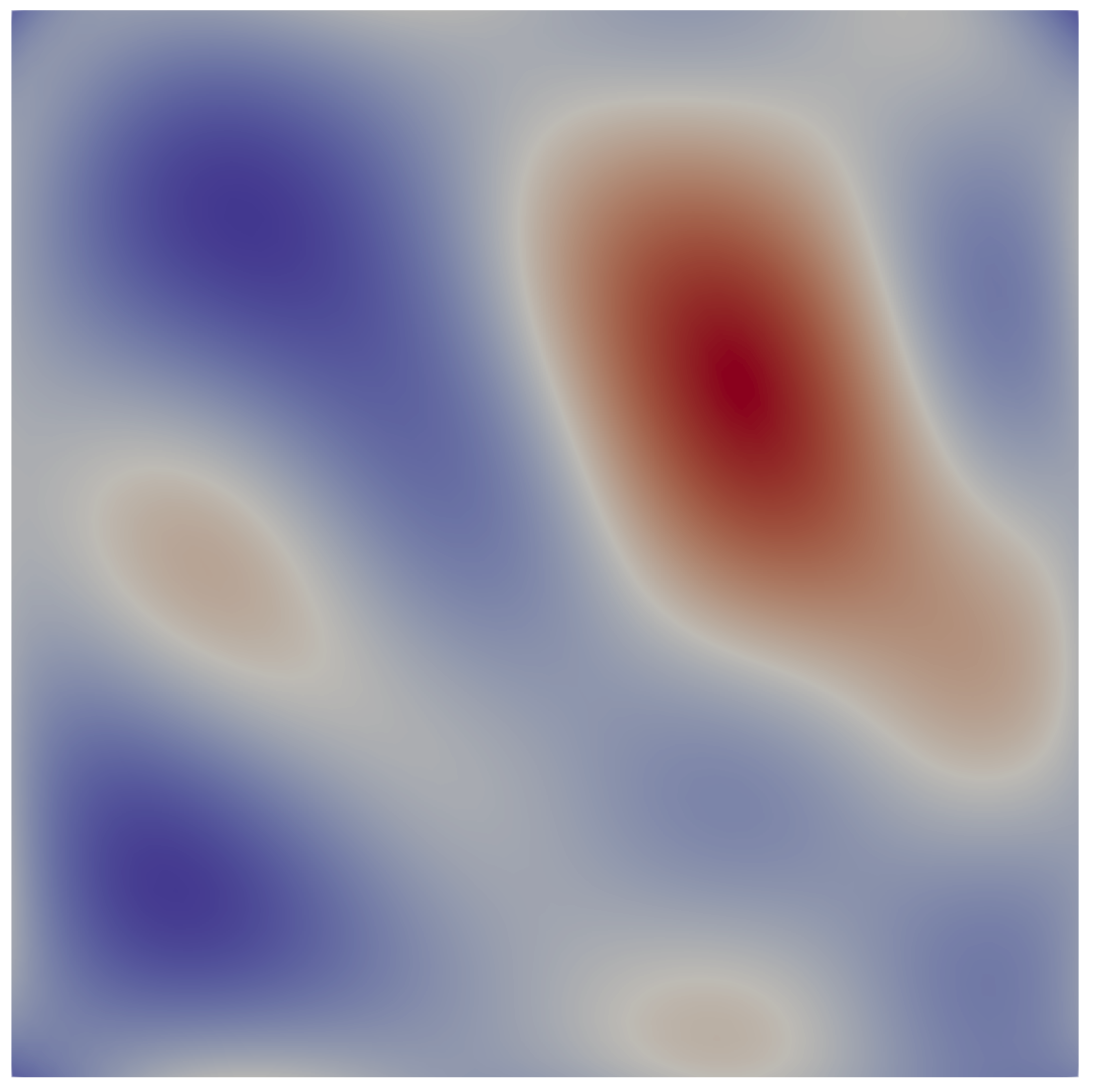}   
   \caption{}\label{}
 \end{subfigure}%
 \caption{$\left(a\right)$ Deflection and $\left(b\right)$ Absolute deflection error contour predicted by DAEM.}
\label{DeflErrContour}
\end{figure}

Next, we test the influence of quadrature points on the accuracy of the solution by calculating the relative error of the maximum deflection and deflection. A series of randomly distributed quadrature points ranging from $[100\quad16900]$ are used to calculate the integrals. The numerical results are shown in Figure~\ref{DAEMMaxErrErr}. Associated contour plots are illustrated in Figure~\ref{DeflErrContour}.

\subsubsection{Annular Plate under uniformly distributed pressure}
\begin{figure}[H]
\captionsetup{width=0.85\columnwidth}
\centering
\begin{subfigure}[b]{6.5cm}
  \centering\includegraphics[height=6cm,width=6.5cm]{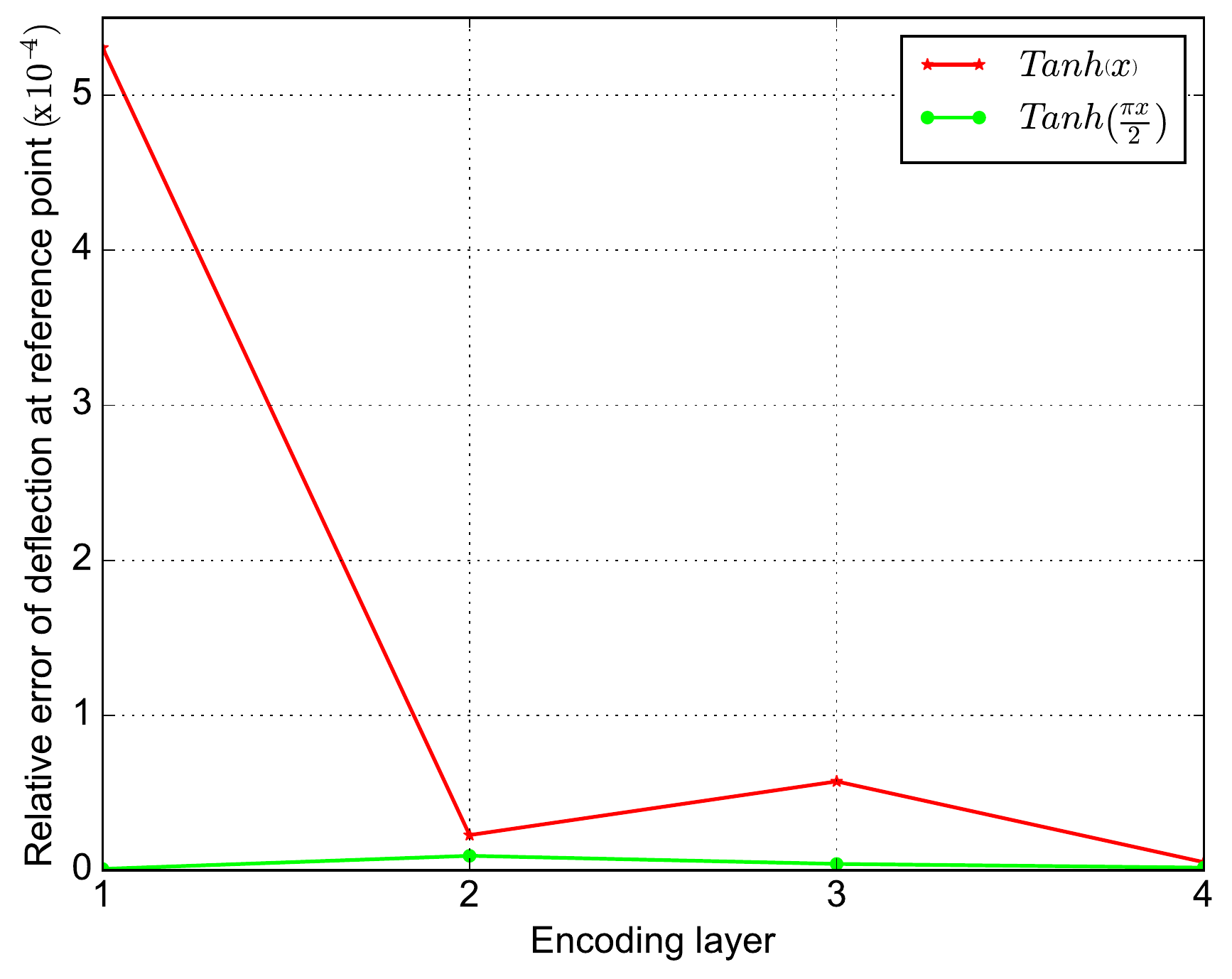}   
   \caption{}\label{}
 \end{subfigure}%
 \begin{subfigure}[b]{6.5cm}
 \centering\includegraphics[height=6.1cm,width=6.5cm]{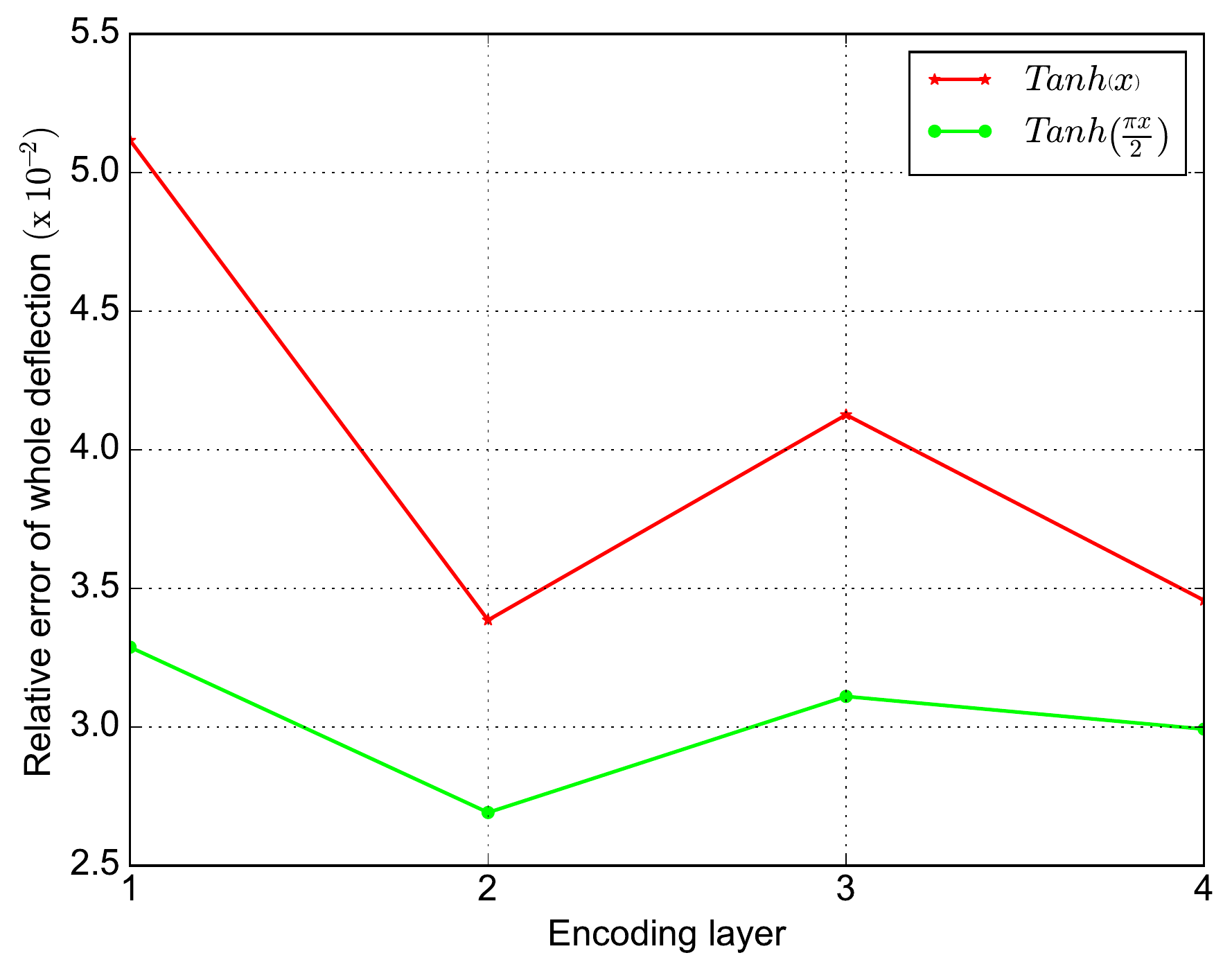}   
   \caption{}\label{}
 \end{subfigure}%
 \caption{Relative error of $\left(a\right)$ deflection at reference point and $\left(b\right)$ all deflection predicted by Tanh and proposed activation function with DAEM.}
\label{RelativerrDNNAcFunanpl}
\end{figure}

\begin{figure}[H]
\captionsetup{width=0.85\columnwidth}
\centering
\begin{subfigure}[b]{6.5cm}
  \centering\includegraphics[height=6cm,width=6.5cm]{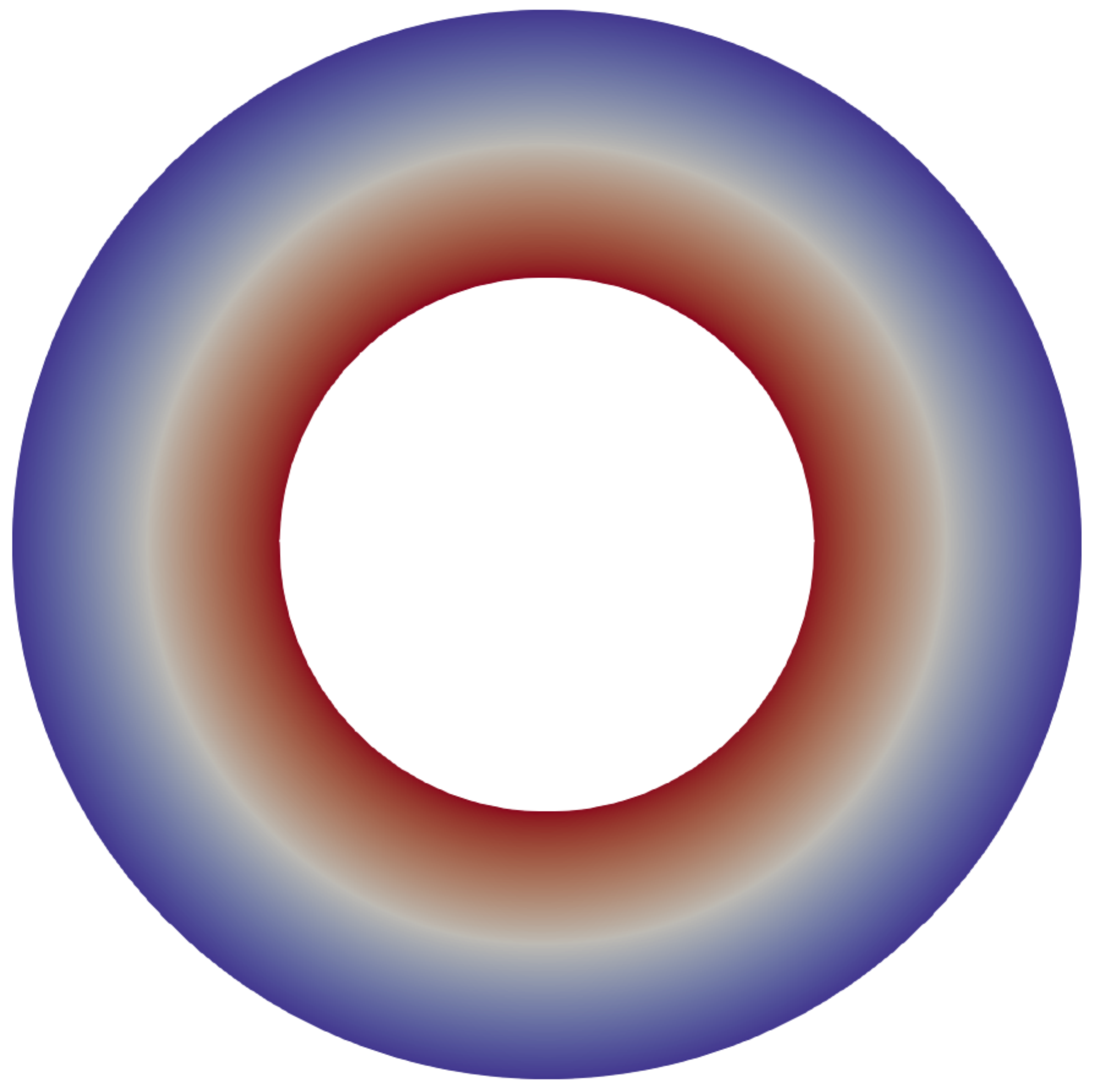}   
   \caption{}\label{}
 \end{subfigure}%
 \begin{subfigure}[b]{6.5cm}
 \centering\includegraphics[height=6cm,width=6.5cm]{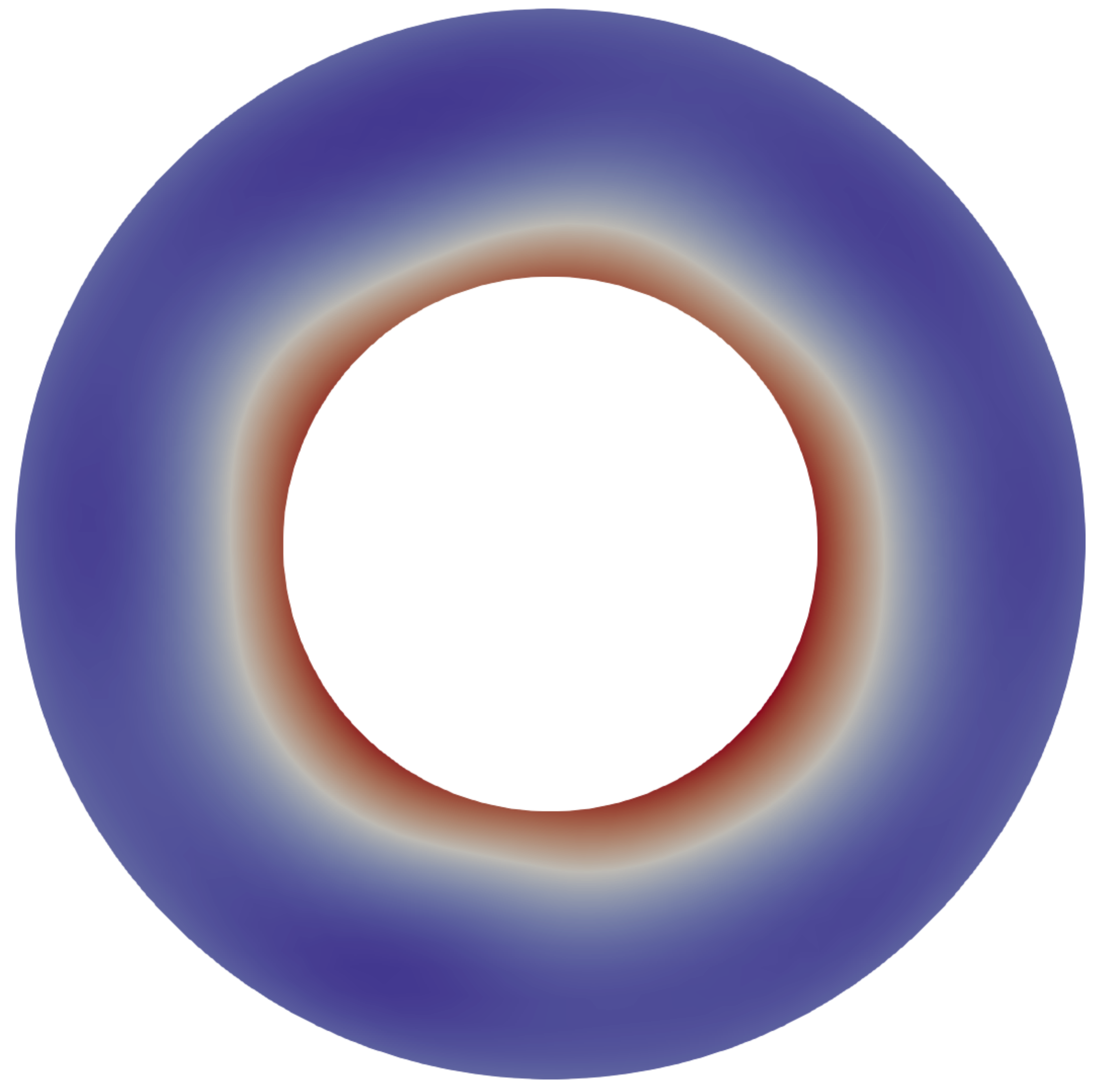}   
   \caption{}\label{}
 \end{subfigure}%
 \caption{$\left(a\right)$ Deflection and $\left(b\right)$ Absolute deflection contour predicted by DAEM.}
\label{DeflErrContourAS}
\end{figure}

Next, we study an annular plate, which is simply-supported on the outer circle and free on the inner circle. The analytical solution of this problem is \cite{timoshenko1959theory}:
\begin{equation}
w=\frac{qa^4}{64D}\left \{ -\left [ 1-\left ( \frac{r}{a} ^{4}\right ) \right ] +\frac{2\alpha_{1}}{1+\nu }\left [ 1-\left (\frac{r}{a}  \right )^{2} \right ]-\frac{4\alpha _{2} \beta^2}{1-\nu }\textup{log}\left ( \frac{r}{a} \right )\right \},
\end{equation}
where $\alpha_1=\left ( 3+\nu  \right )\left ( 1-\beta^2\right )-4\left ( 1+\nu \right )\beta^2\kappa$, $\alpha_2=\left ( 3+\nu  \right )+4\left ( 1+\nu \right )\kappa$, $\beta=\frac{b}{a}$, $ \kappa=\frac{\beta^2}{1-\beta^2}\textup{log}\beta$,  $a$, $b$ being the outer and inner radius of the annular plate, respectively. We also show results for a referennce point $(\frac{a+b}{2},0)$ and study different activation functions. We again observe the exploding gradient problem for DAEM with the $Tanh$ activation function, which can be alleviated by the $Tanh(\frac{\pi}{2})$, see Figure~\ref{RelativerrDNNAcFunanpl}. Contour plots of the deflection and absolute deflection are depicted in Figure~\ref{DeflErrContourAS} showing that the predicted deflection agrees well with the analytical solution.

\subsubsection{Square plate on Winkler foundation}
Finally, we study a simply-supported plate on Winkler foundation assuming the foundation's reaction $p\left(x,y\right)$ is expressed by $p\left(x,y\right)=\textit{k}w$, $\textit{k}$ being the foundation modulus. For a plate on a continuous Winkler foundation, the potential energy needs to be added to the total potential energy, Equation~\ref{pi_totalenergy}:
\begin{equation}
W_{s}\left(\textit{\textbf{x}}\,_\Omega;\theta\right)=\int_{\Omega } k\textit{w}^h\left(\textit{\textbf{x}}\,_\Omega;\theta\right)^2 d\Omega 
\label{Winklerfdenergy} 
\end{equation}   
The analytical deflection is given by  \cite{timoshenko1959theory}:
\begin{equation}
w=\frac{16p}{ab}\sum_{m=1,3,5,\cdots  }^{\infty}\sum_{n=1,3,5,\cdots  }^{\infty}\frac{\textrm{sin}\frac{m\pi x }{a} \textrm{sin}\frac{n\pi y }{b}}{mn\left [ \pi^4D\left ( \frac{m^2}{a^2} +\frac{n^2}{b^2}\right )^2+\textit{k} \right ]}
\end{equation} 

\begin{figure}[H]
\captionsetup{width=0.85\columnwidth}
\centering
\includegraphics[height=7cm,width=10cm]{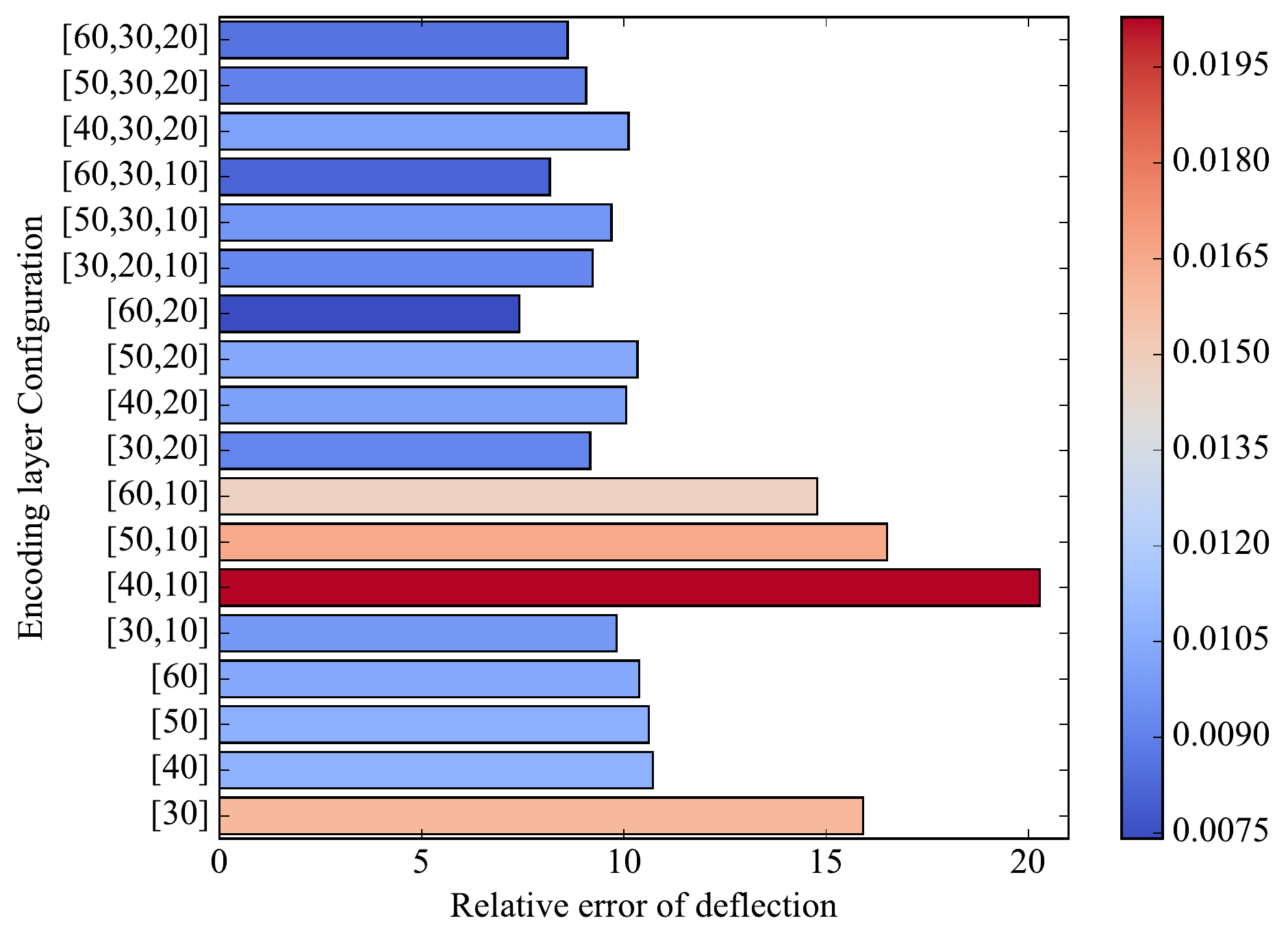}
\caption{The relative error of deflection for DAEM with different encoding cconfigurations.}
\label{ComputationDAEMRerrCODConf1}
\end{figure}

\begin{figure}[H]
\captionsetup{width=0.85\columnwidth}
\centering
\includegraphics[height=7cm,width=10cm]{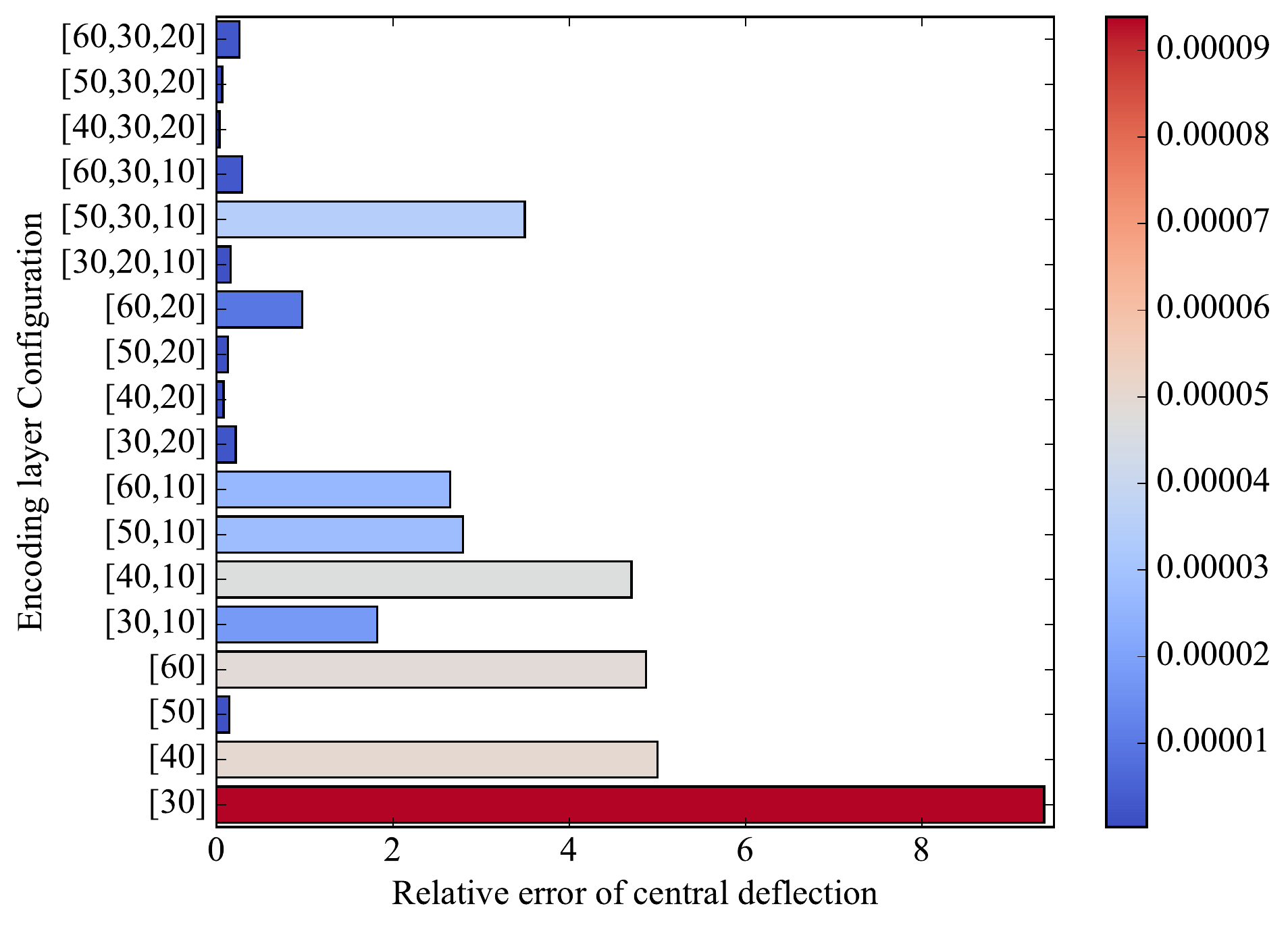}
\caption{The relative error of maximum deflection for DAEM with different encoding cconfigurations.}
\label{ComputationDAEMMaxRerrCODConf1}
\end{figure}

\begin{figure}[H]
\captionsetup{width=0.85\columnwidth}
\centering
\includegraphics[height=7cm,width=10cm]{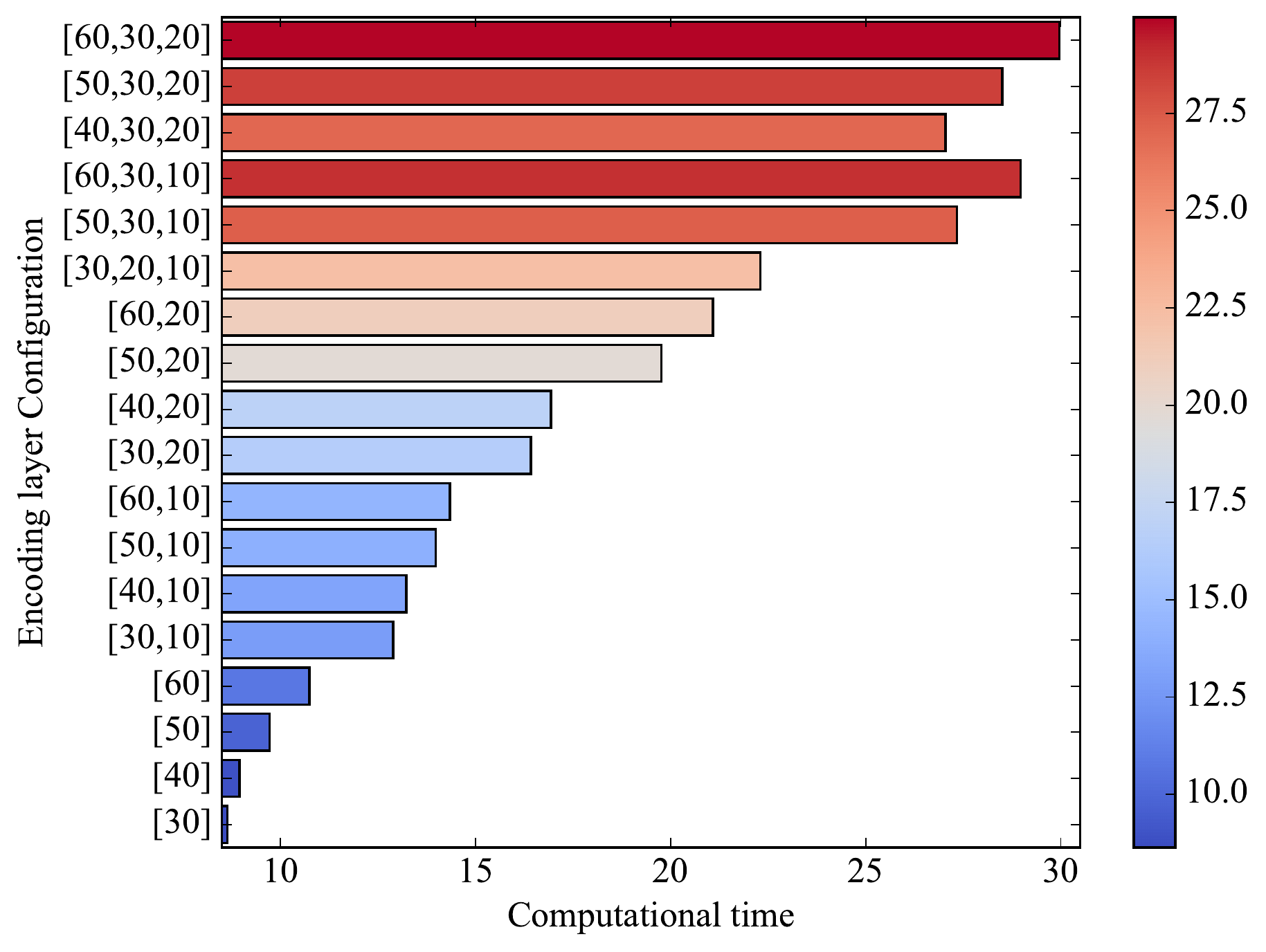}
\caption{The computational time of DAEM with different encoding cconfigurations.}
\label{ComputationDAEMCmptmCODConf1}
\end{figure}

Different configurations of the deep autoencoder are tested. The relative errors in the deflection and maximum deflection are shown in Figure~\ref{ComputationDAEMRerrCODConf1} and \ref{ComputationDAEMMaxRerrCODConf1}, respectively. Increasing the layers leads to more accurate results with increasing encoding layers. For some cases, increasing the width of the neural network does not improve the results, so that a deep neural network is preferable. The computational cost is depicted  in Figure~\ref{ComputationDAEMCmptmCODConf1}. As expected, more encoding layers increases the computational cost. However, note that this includes also the training cost. Once the network has been trained, the solution will be obtained much faster.
  
\subsection{Vibration analysis}
\begin{figure}[H]
\captionsetup{width=0.85\columnwidth}
\centering
\begin{tabular}{c}
\includegraphics[height=6.5cm,width=6.5cm]{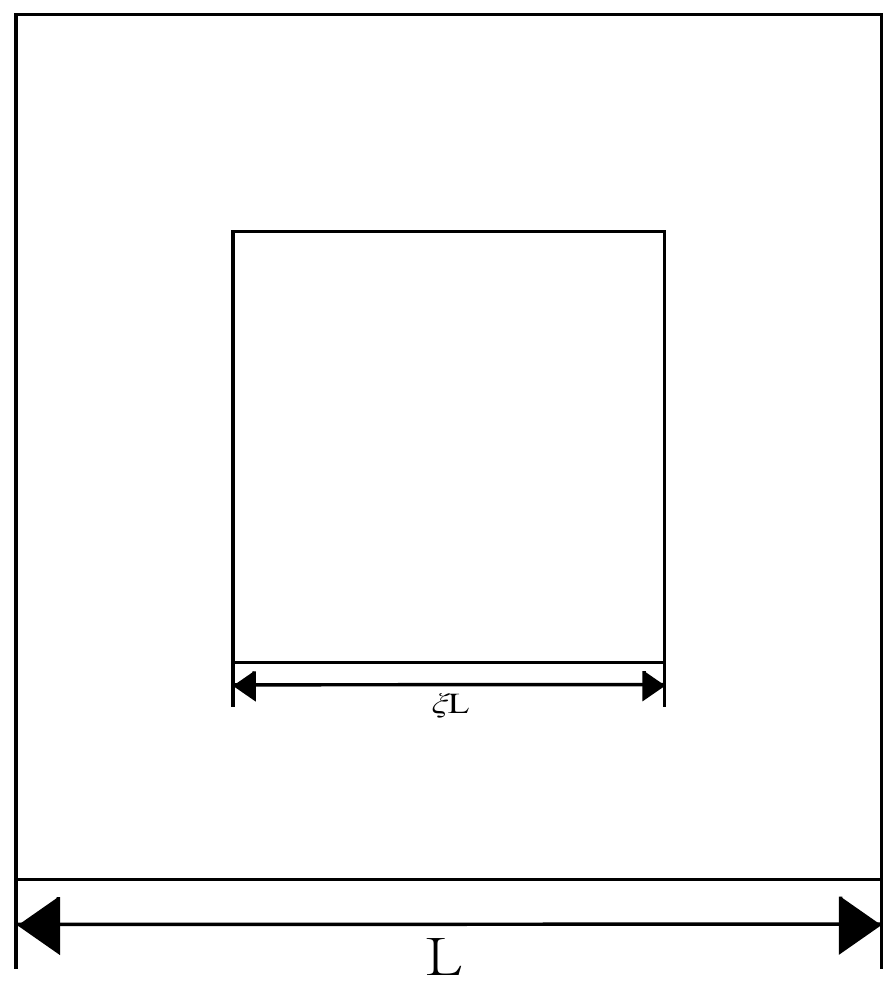}
\end{tabular}
\caption{Kirchhoff thin plate in the Cartesian coordinate system.}
\label{Figure17:platewithsquarecutout}
\end{figure}

We now apply DAEM to extract the fundamental frequency in a transversal vibration analysis. The results are  compared with reference solutions from \cite{ZHANG201865,LAM198949,LIEW2003941,SHUFRIN20166983}.
Let us consider a square plate with a square cutout as shown in Figure~\ref{Figure17:platewithsquarecutout},  $\xi$ is the ratio of the inner to outer square size. Various boundary conditions are studied. The non-dimensional fundamental frequency parameter $\overline{\Omega }=\omega L ^2\sqrt{\rho h/D}$ with different cutout ratio $\xi$ is studied. A deep autoencoder with encoding layers $[40,20]$ is adopted as this architecture provided accurate results for the bending analysis while being computationally efficient. 

The nondimensional fundamental frequency is depcited in Table~\ref{tab:Table2}. The predicted results agree well with reference results of the HBM method \cite{ZHANG201865}, Modified Ritz method \cite{LAM198949}, FEM \cite{LAM198949}, and Discrete Ritz method \cite{SHUFRIN20166983}.

\begin{table}[H] 
\captionsetup{width=0.85\columnwidth}
\caption{Comparison of frequency parameter predicted by DAEM with other reference methods} 
\vspace{-0.3cm}
\centering 
\resizebox{0.9\columnwidth}{!}{%
\begin{tabular}{c c c c c c} 
\toprule 
\toprule 
& \multicolumn{5}{c}{\textbf{Non-dimensional fundamental frequency parameter $\overline{\Omega }=\omega L ^2\sqrt{\rho h/D}$}} \\ 
\cmidrule(l){2-6} 
\textbf{Cutout ratio $\xi$} & DAEM & HBM method & Modified Ritz method&FEM&Discrete Ritz method\\ 
\midrule 
0	&19,7382 &19,7390 & 19,7400 & 19,7520 & 19,7390 \\ 
0,1	&19,3508 &19,4440 & 19,1830 & 19,3570 & 19,4130 \\ 
0,2	&19,0284 &19,1280 & 18,7620 & 19,1200 & 19,0380 \\ 
0,3	&19,3834 &19,4450 & 19,1830 & 19,3570 & 19,3910 \\ 
0,4	&20,8201 &20,7530 & 20,7850 & 20,7320 & 20,7240 \\ 
0,5	&23,4641 &23,4530 & 23,6640 & 23,2350 & 23,4410 \\ 
0,6	&28,2706 &28,3750 & 28,8440 & 28,2410 & 28,5260 \\ 
0,7	&38,1596 &37,5720 & 38,1580 & 35,5790 & 37,8920 \\ 
0,8	&58,0804 &57,4120 & 58,0620 & 57,4520 & 57,8380 \\ 
0,9	&120,9580&120,0200& 121,2300& 120,3900& 120,9900\\ 
\bottomrule 
\end{tabular}
}
\label{tab:Table2} 
\end{table}

The fundamental mode shapes for different cutout ratios are shown in Table~\ref{tab:Table3} and agree well with results in \cite{ZHANG201865}. 

\begin{table}[H] 
\captionsetup{width=0.85\columnwidth}
\caption{Fundamental mode shapes predicted by DAEM with different cutout ratio} 
\vspace{-0.3cm}
\centering 
\resizebox{0.9\columnwidth}{!}{%
\begin{tabular}{c c c c} 
\toprule 
\toprule 
\textbf{Cutout ratio $\xi$} & \textbf{Fundamental mode shape}&\textbf{Cutout ratio $\xi$} & \textbf{Fundamental mode shape}\\ 
\midrule 
0	&\hspace{-3.5cm}\parbox[c]{1em}{\includegraphics[width=4cm]{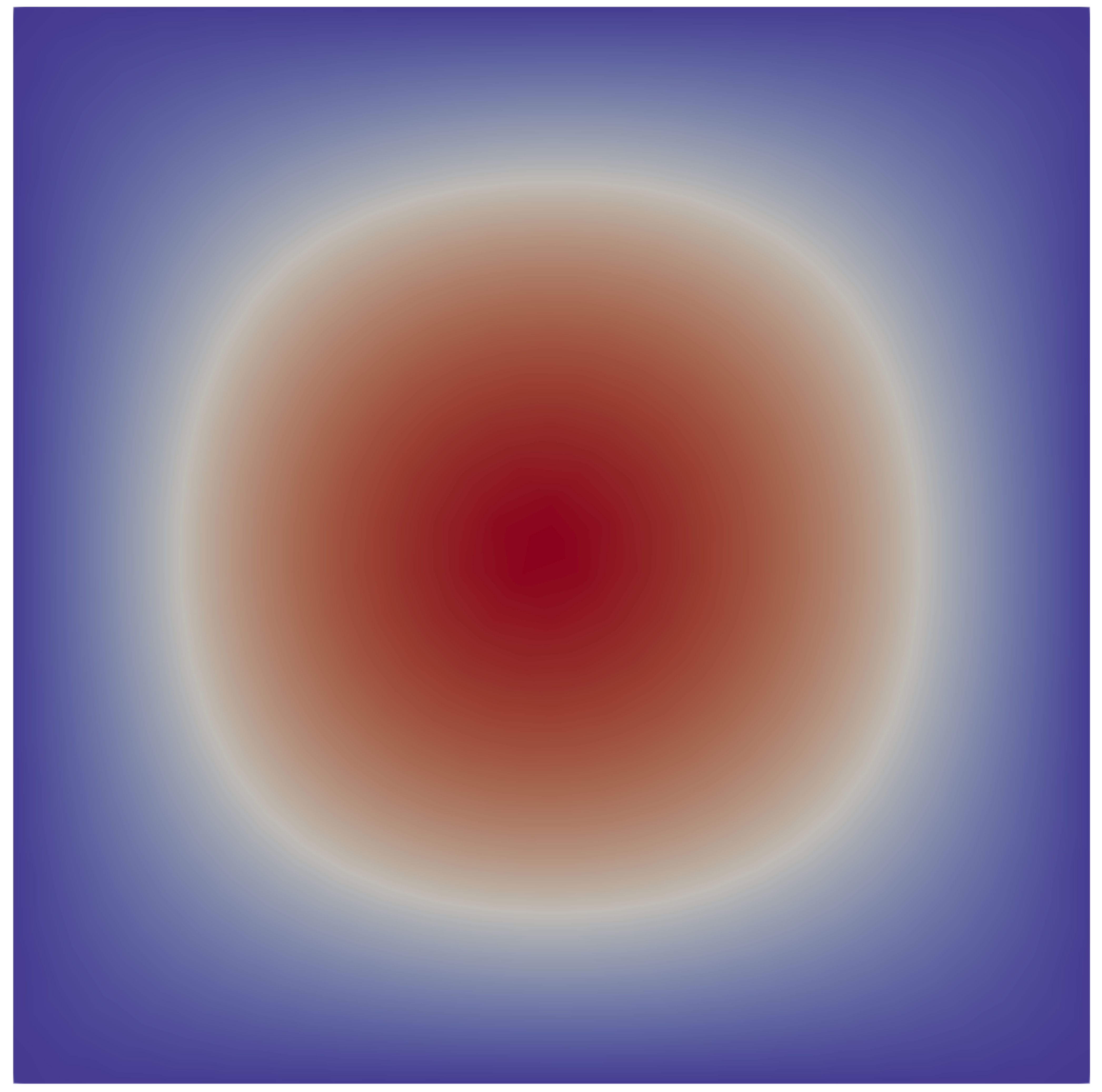}}&0,5	&\hspace{-3.5cm}\parbox[c]{1em}{\includegraphics[width=4cm]{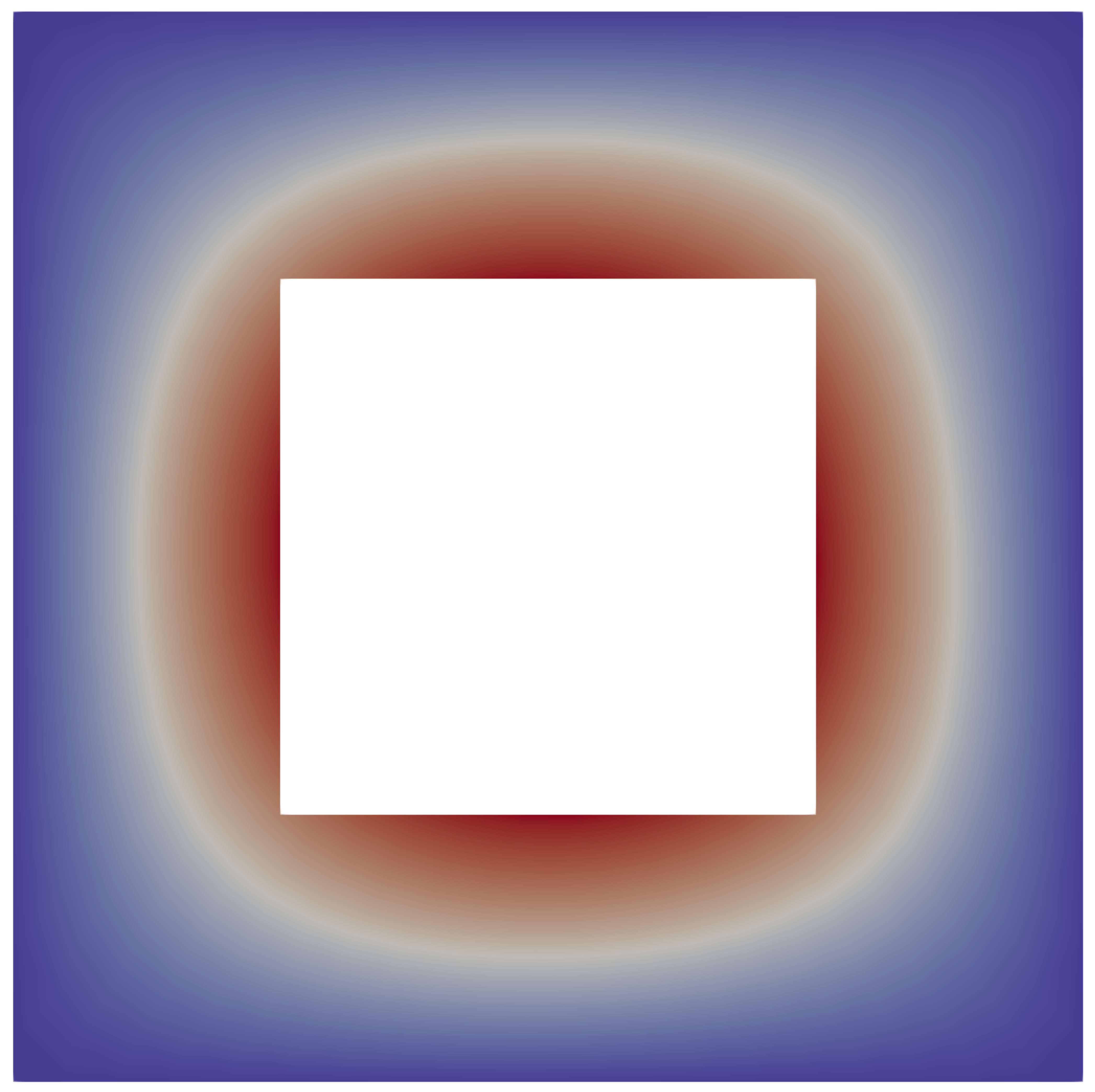}}\\ 
0,1	&\hspace{-3.5cm}\parbox[c]{1em}{\includegraphics[width=4cm]{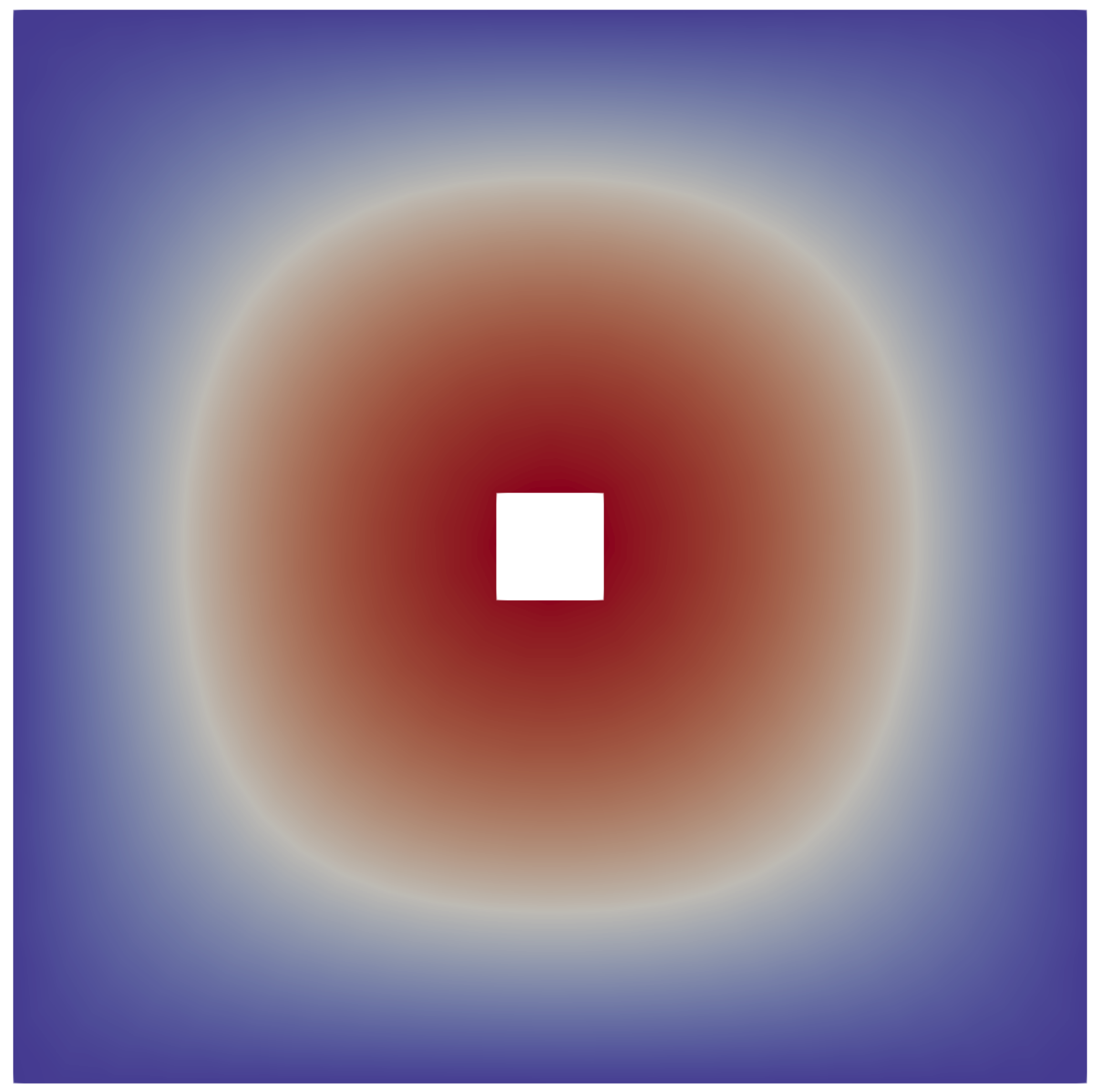}}&0,6	&\hspace{-3.5cm}\parbox[c]{1em}{\includegraphics[width=4cm]{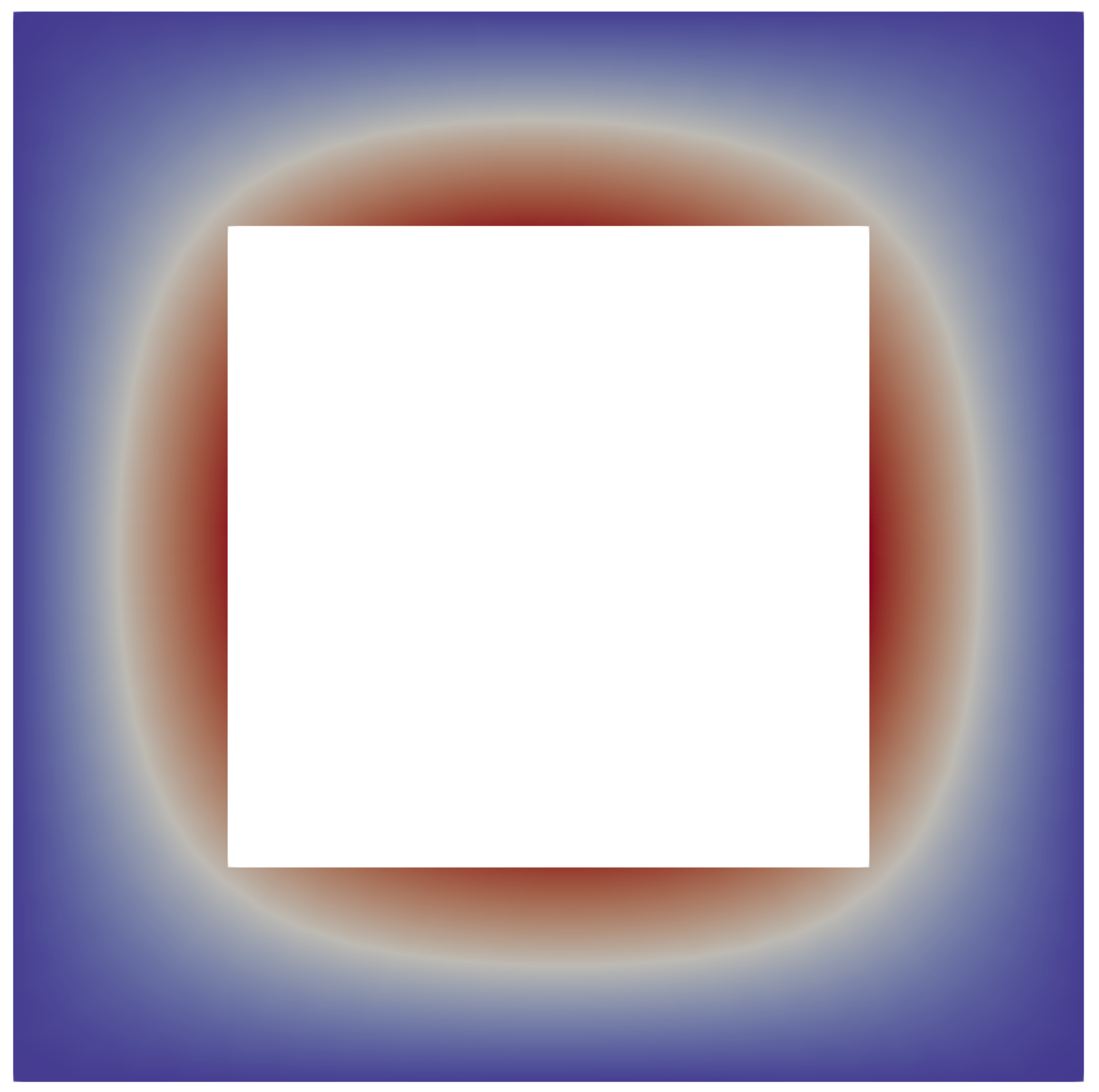}}\\ 
0,2	&\hspace{-3.5cm}\parbox[c]{1em}{\includegraphics[width=4cm]{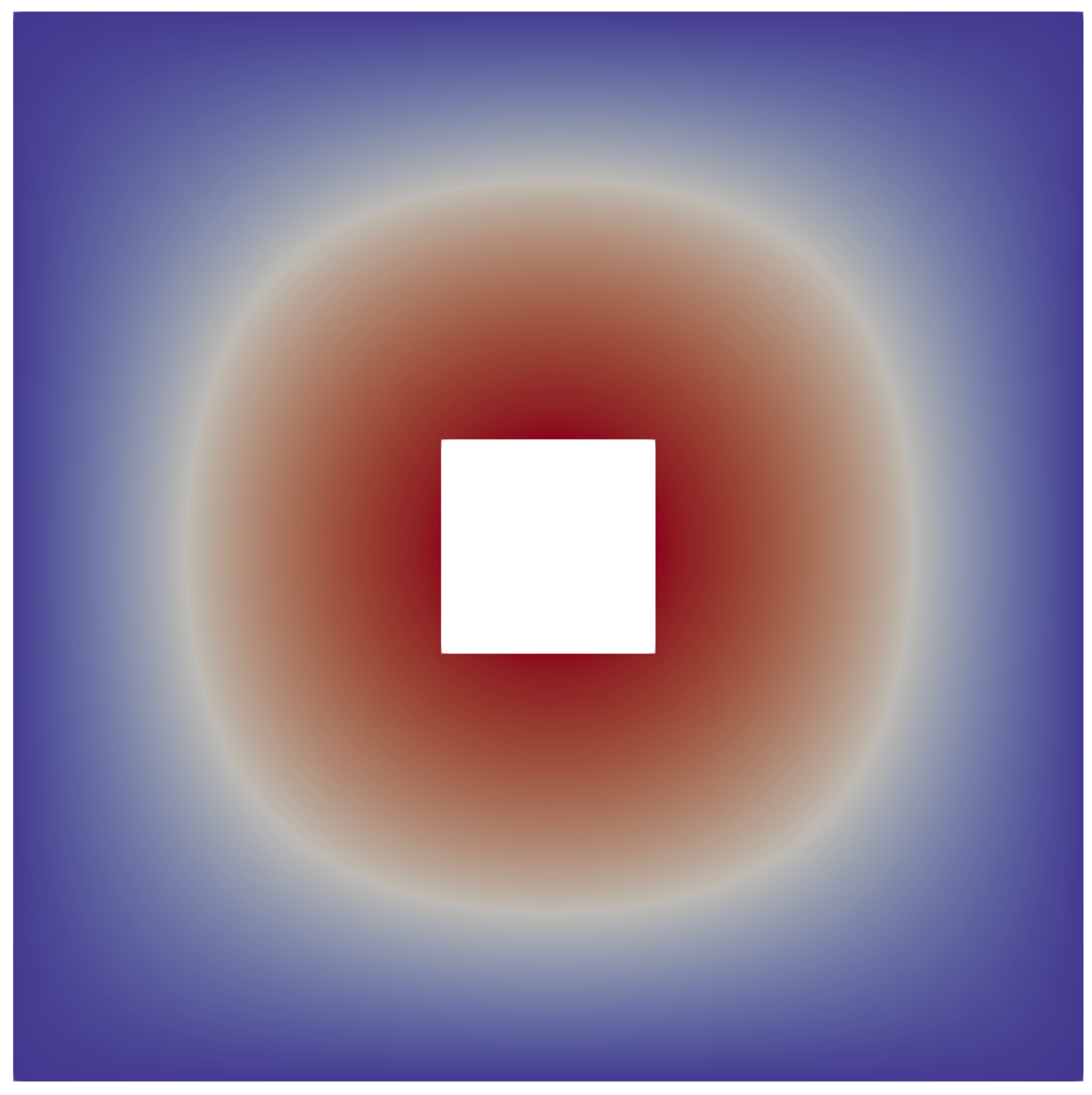}}&0,7	&\hspace{-3.5cm}\parbox[c]{1em}{\includegraphics[width=4cm]{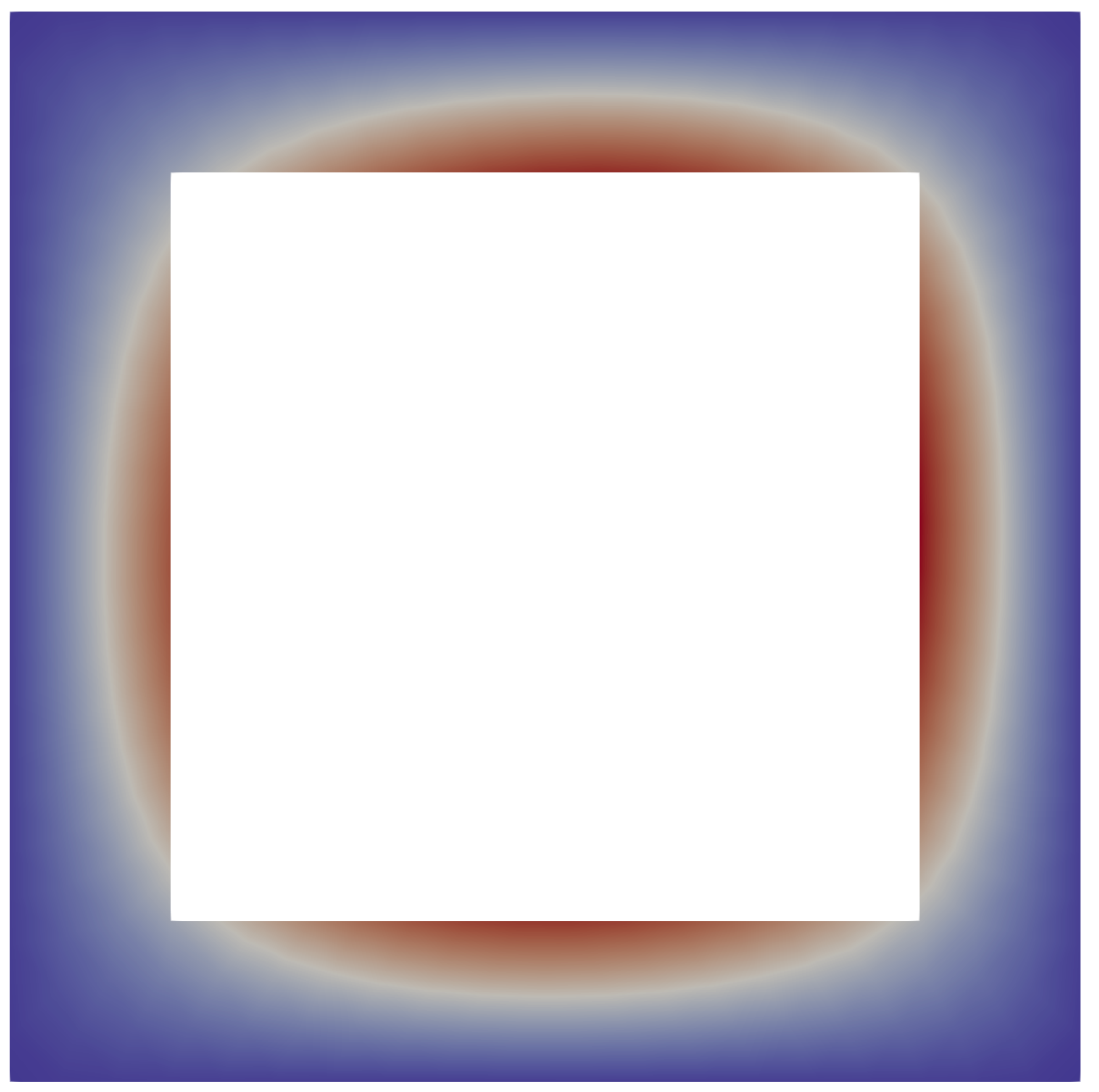}}\\ 
0,3	&\hspace{-3.5cm}\parbox[c]{1em}{\includegraphics[width=4cm]{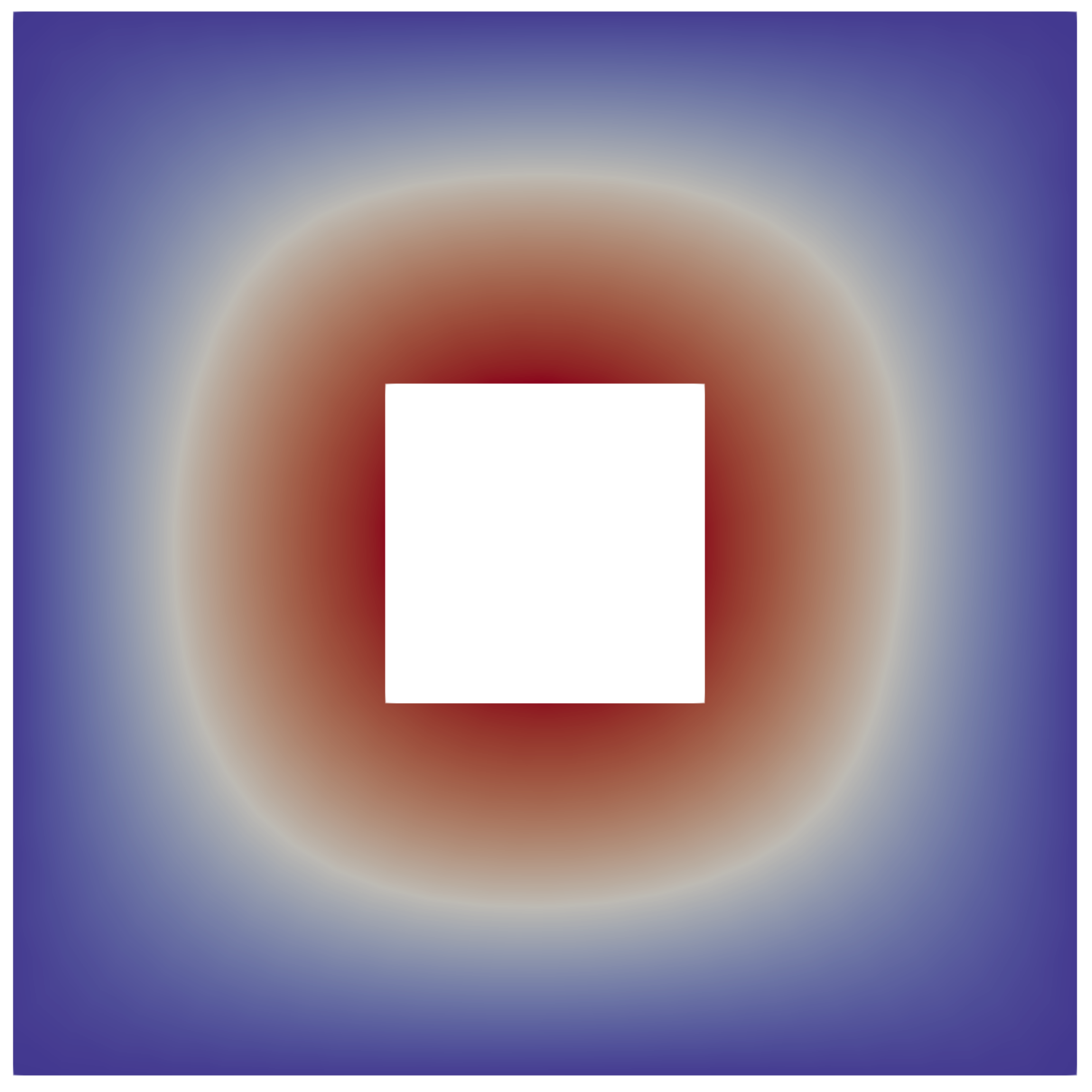}}&0,8	&\hspace{-3.5cm}\parbox[c]{1em}{\includegraphics[width=4cm]{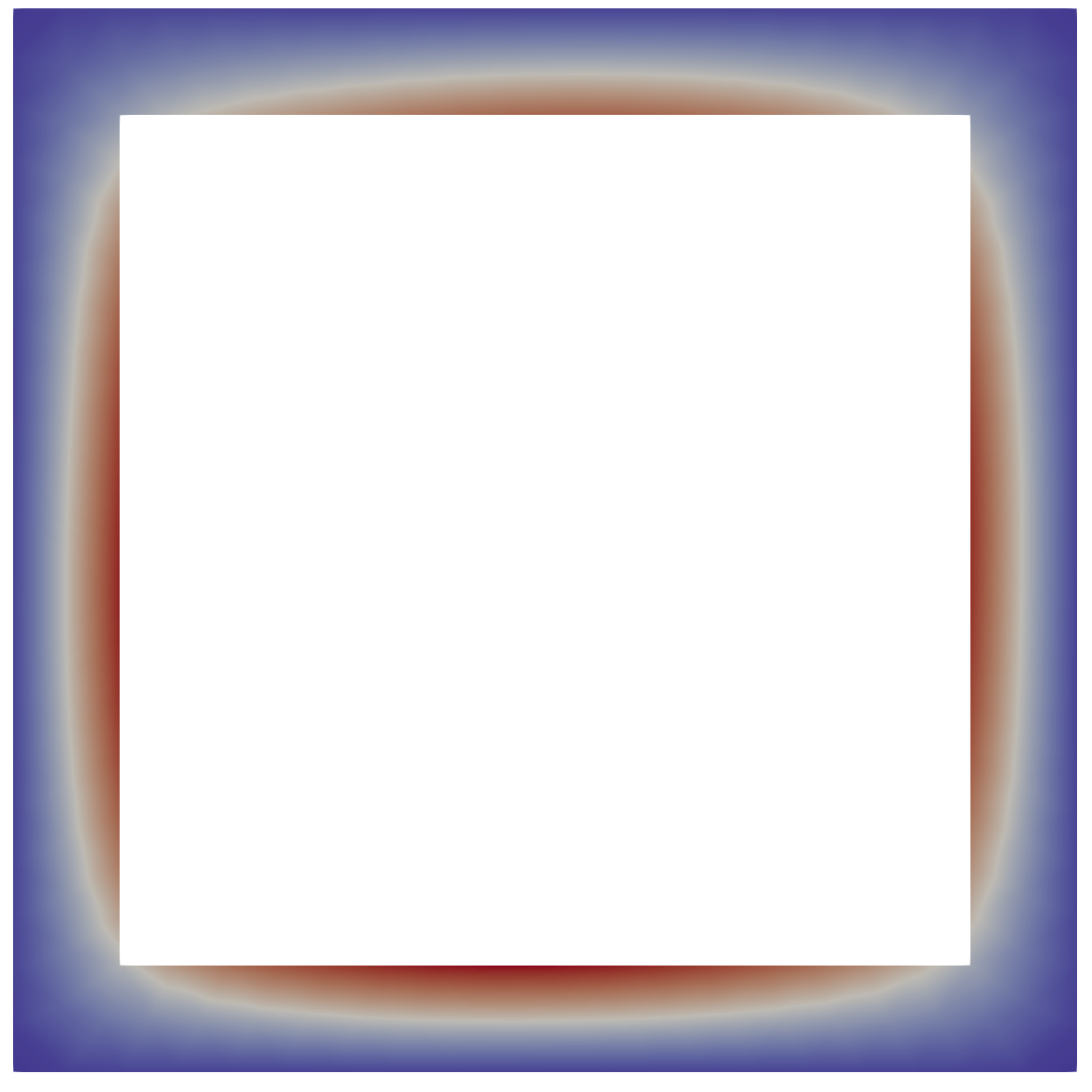}}\\ 
0,4	&\hspace{-3.5cm}\parbox[c]{1em}{\includegraphics[width=4cm]{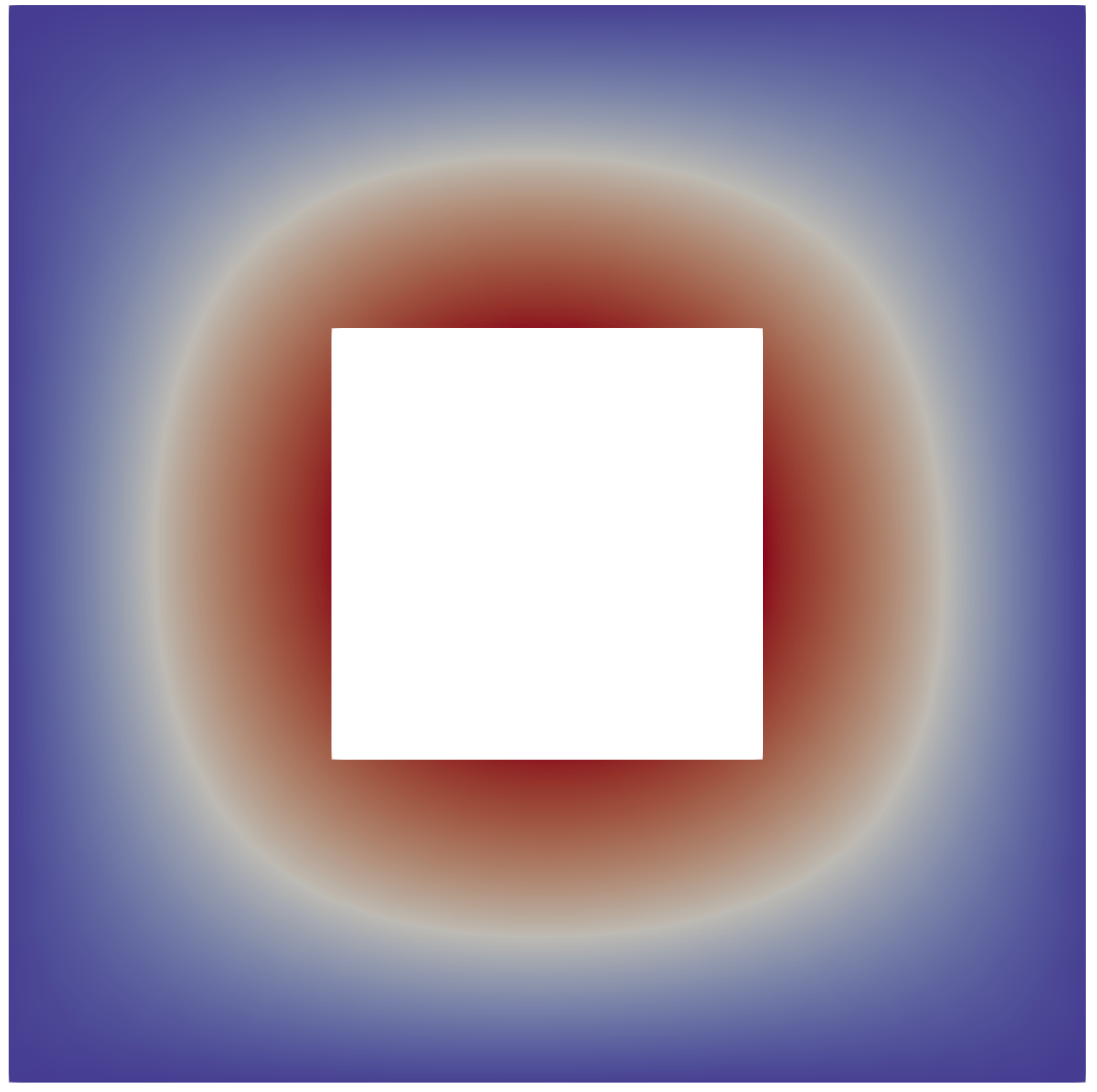}}&0,9	&\hspace{-3.5cm}\parbox[c]{1em}{\includegraphics[width=4cm]{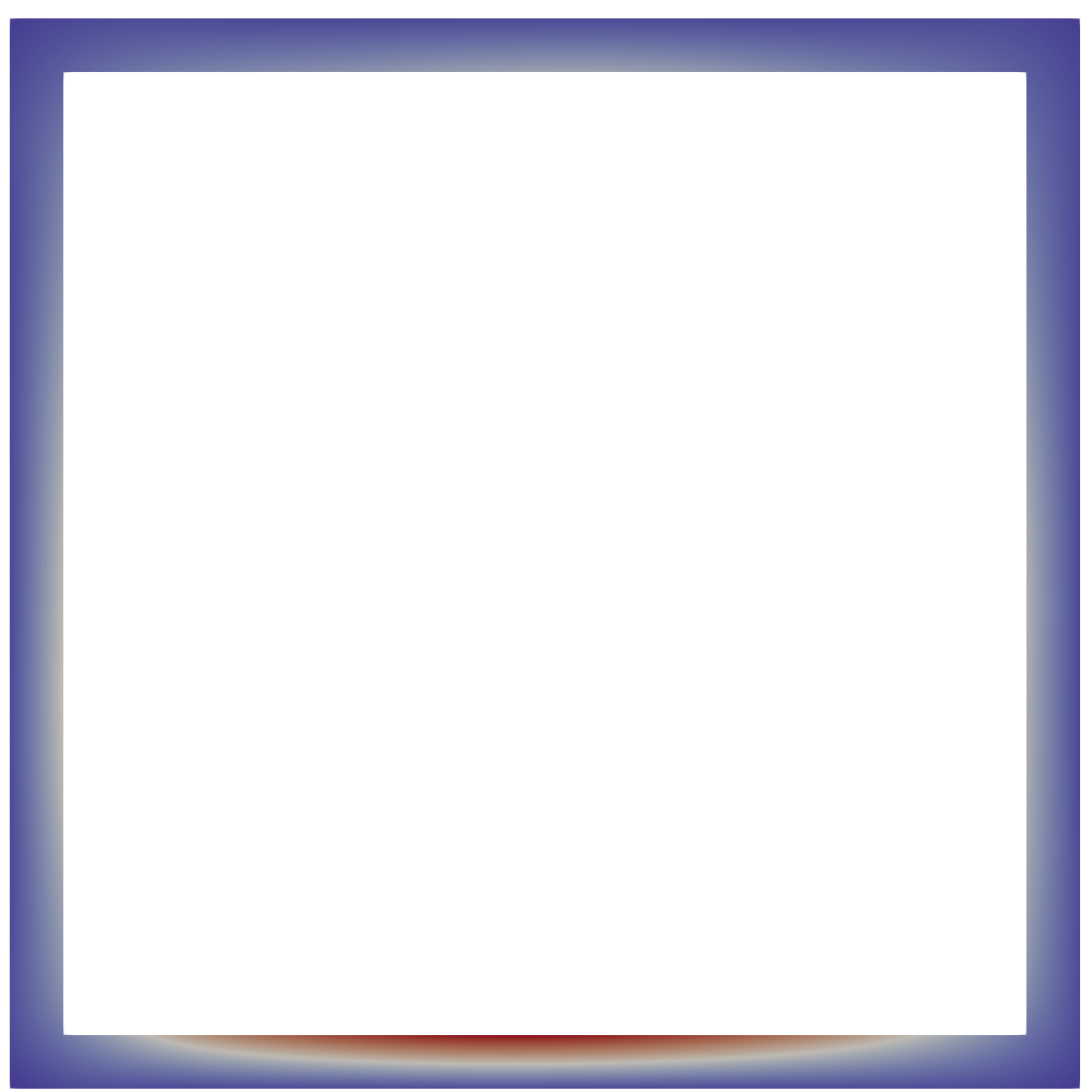}}\\ 
\bottomrule 
\end{tabular}
}
\label{tab:Table3} 
\end{table}

Next, we perform a vibration analysis of the square plate for two other boundary conditions, i.e. 1: all outer edges are clamped and 2. two opposite outer edges are clamped, while the other two outer edges are simply-supported. The cutout ratio of this square plate is selected to be $\xi=0.4$. The results are summarized in Table~\ref{tab:Table4} and agree well with results from \cite{ZHANG201865}. 
\begin{table}[H] 
\captionsetup{width=0.85\columnwidth}
\caption{Fundamental frequency parameter and mode shapes predicted by DAEM considering different boundary conditions $\left(\xi=0.4\right)$} 
\vspace{-0.3cm}
\centering 
\resizebox{0.9\columnwidth}{!}{%
\begin{tabular}{c c c} 
\toprule 
\toprule 
\textbf{Boundary Conditions} & \textbf{$\overline{\Omega }=\omega L ^2\sqrt{\rho h/D}$} & \textbf{Fundamental mode shape}\\ 
\midrule 
CCCC&49,3091&\hspace{-3.5cm}\parbox[c]{1em}{\includegraphics[width=4cm]{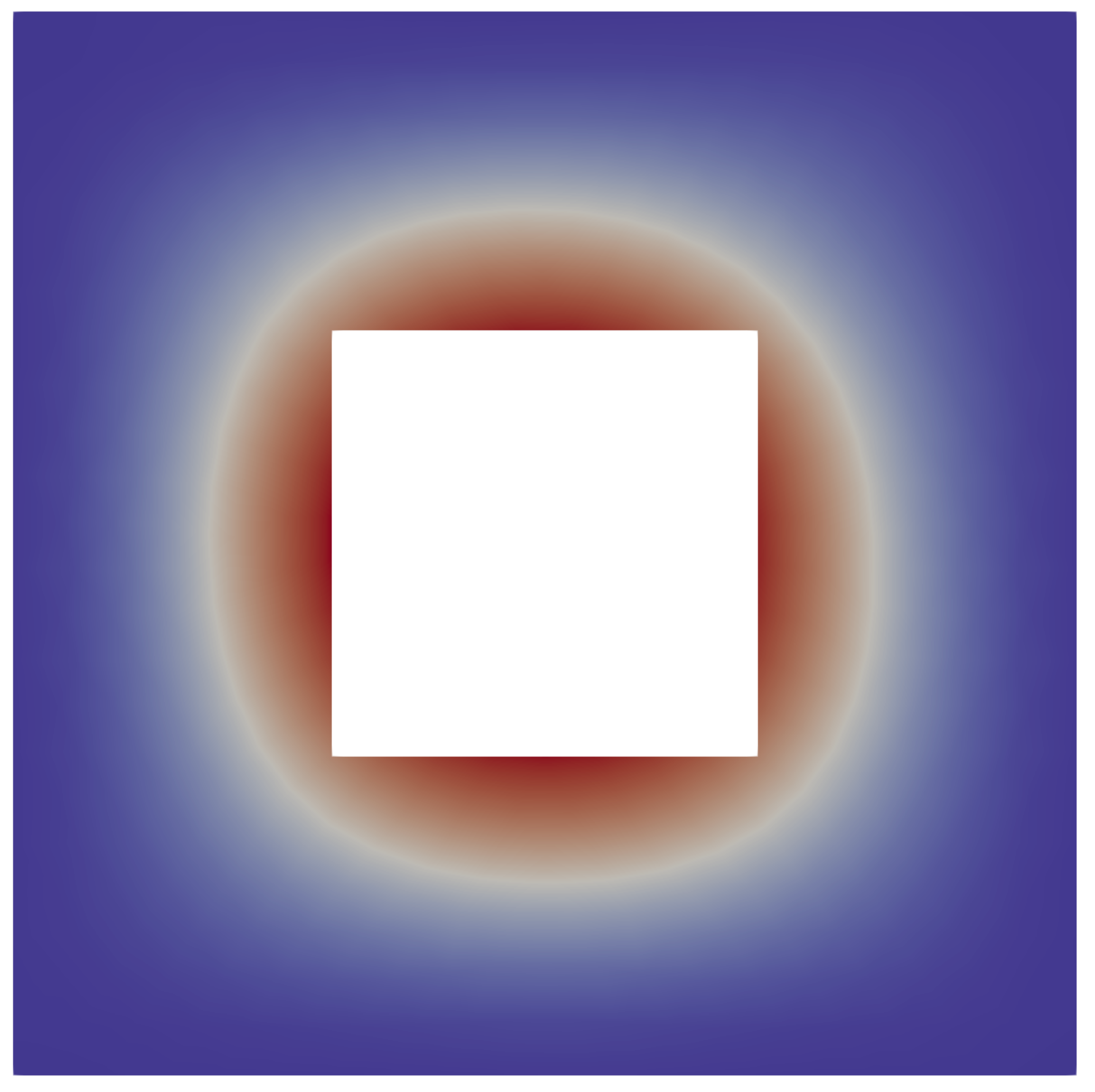}}\\ 
CSCS&35,4996&\hspace{-3.5cm}\parbox[c]{1em}{\includegraphics[width=4cm]{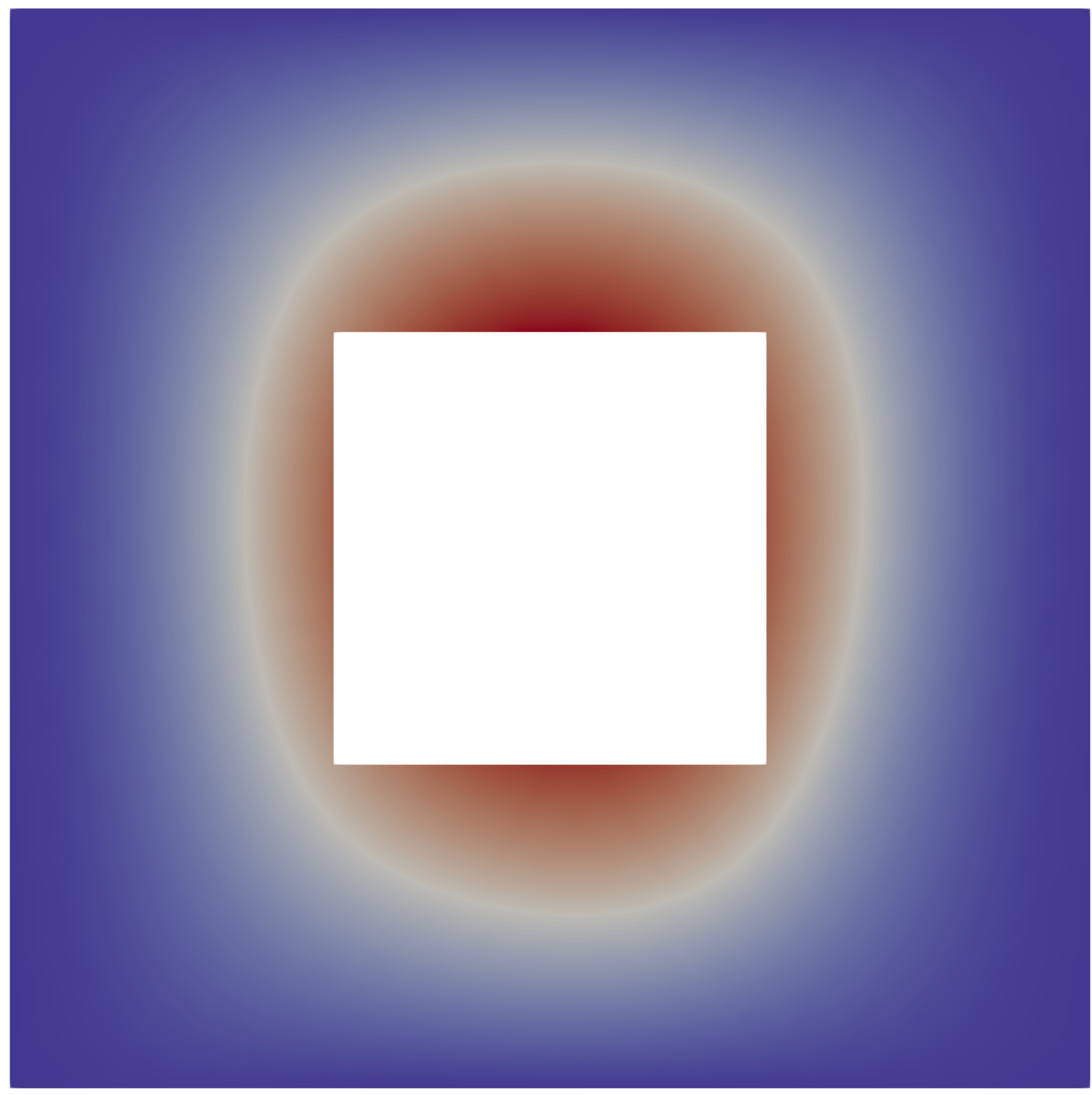}}\\ 
\bottomrule 
\end{tabular}
}
\label{tab:Table4} 
\end{table}

\subsection{Buckling analysis}
\begin{figure}[H]
\captionsetup{width=0.85\columnwidth}
\centering
\begin{tabular}{c}
\includegraphics[height=6.5cm,width=8.5cm]{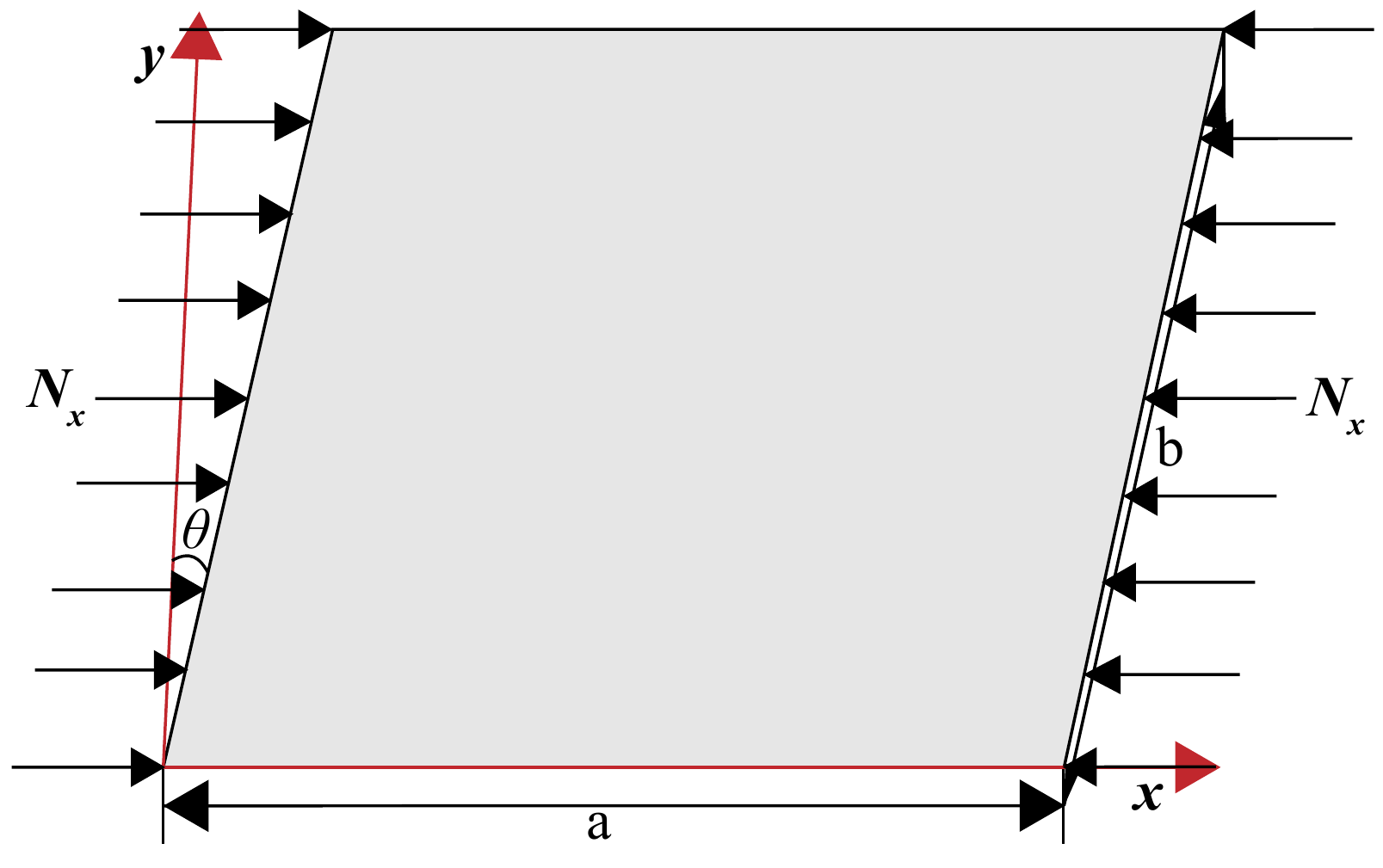}
\end{tabular}
\caption{Kirchhoff thin plate in the Cartesian coordinate system.}
\label{Figure18:skewplate}
\end{figure}

Finally, we  study a skew plate with different skew angles $\theta$ and aspect ratio $\xi=a/b$ subjected to uniaxial inplane compressive loading, see Figure~\ref{Figure18:skewplate}. We consider simply-supported and clamped boundary conditions. The effects of the skew angle, aspect ratio on the critical buckling load factor are computed and compared with the reference solution presented in \cite{srinivasa2012buckling}.

\begin{table}[H] 
\captionsetup{width=0.85\columnwidth}
\caption{Comparison of critical buckling load parameter predicted by DAEM with other reference methods} 
\vspace{-0.3cm}
\centering 
\resizebox{0.9\columnwidth}{!}{%
\begin{tabular}{c c c c c c c} 
\toprule 
\toprule 
& & \multicolumn{5}{c}{\textbf{Non-dimensional critical buckling load parameter $K_{cr}$}} \\ 
\cmidrule(l){3-7} 
\textbf{Aspect ratio $\xi$} &\textbf{Skew angle $\theta$} & DAEM & Rayleigh-Ritz method &FEM&CQUAD4&CQUAD8\\ 
\midrule 
0,5	& 0$^{\circ}$ & 6,2575	  &6,2500	 &6,2510  &6,2010	&6,2180\\
0,5	& 15$^{\circ}$ & 7,0172	  &7,0000	 &6,9800  &6,8550	&6,9080\\
0,5	& 30$^{\circ}$ & 9,9614	  &10,0200   &9,9400  &9,8950	&10,0000\\
0,5	& 45$^{\circ}$ & 19,4074   &19,3000   &9,4200  &18,9510	&19,2520\\
1	& 0$^{\circ}$ & 4,0007	  &4,0000	 &4,0000  &3,9190	&4,0000\\
1	& 15$^{\circ}$ & 4,5073	  &4,4800	 &4,4000  &4,3060	&4,3550\\
1	& 30$^{\circ}$ & 5,8504	  &6,4100	 &5,9300  &5,7610	&5,8750\\
1	& 45$^{\circ}$ & 10,5208   &12,3000   &10,3600 &9,5260	&9,9540\\
1,5	& 0$^{\circ}$ & 4,3710	  &0,0000	 &0,0000  &4,2560	&4,2700\\
1,5	& 15$^{\circ}$ & 4,6558	  &4,7700	 &4,6800  &4,6400	&4,6480\\
1,5	& 30$^{\circ}$ & 5,9504	  &6,3700	 &5,8900  &5,9550	&5,8650\\
1,5	& 45$^{\circ}$ & 9,1843	  &10,9000   &8,9500  &9,0760	&9,1390\\
2	& 0$^{\circ}$ & 3,9354	  &0,0000	 &0,0000  &3,8850	&3,9030\\
2	& 15$^{\circ}$ & 4,3499	  &4,3300	 &4,3400  &4,2710	&4,3130\\
2	& 30$^{\circ}$ & 5,5677	  &6,0300	 &5,5900  &5,5960	&5,6050\\
2	& 45$^{\circ}$ & 8,9418	  &10,3000   &8,8000  &8,8550	&8,8710\\
\bottomrule 
\end{tabular}
}
\label{tab:Table5} 
\end{table}

The nonlinear encoding layer adopted in this study is $[60,20]$, which is better suitable for varying geometries. The simply-supported skew plate is studied first and the associated results are shown in Table~\ref{tab:Table5}. The numerical results are compared with results of the Rayleigh-Ritz method, FEM, CQUAD4 and CQUAD8 \cite{srinivasa2012buckling}. As the skew angle increases, the critical buckling load parameter $K_{cr}=\frac{\lambda_{cr}b^2h}{\pi^2D}$ increases. For the clamped skew plate, we exemplary show results for the skew plate with aspect ratio $\xi=1$. The numerical results can be found in Table~\ref{tab:Table6}. The predicted results agree well with the analytical solution. 

\begin{table}[H] 
\captionsetup{width=0.85\columnwidth}
\caption{Critical buckling load parameter of clamped skew plate (\textbf{$\xi=1$})} 
\vspace{-0.3cm}
\centering 
\resizebox{0.9\columnwidth}{!}{%
\begin{tabular}{c c c c c c} 
\toprule 
\toprule 
& \multicolumn{5}{c}{\textbf{Non-dimensional critical buckling load parameter $K_{cr}$}} \\ 
\cmidrule(l){2-6} 
\textbf{Skew angle $\theta$} & DAEM & Rayleigh-Ritz method &FEM&CQUAD4&CQUAD8\\ 
\midrule 
0$^{\circ}$&10,0909&10,0000&10,0800&9,8540&10,0000\\
15$^{\circ}$&10,7821&10,9000&10,8400&10,6900&10,7750\\
30$^{\circ}$&13,7013&13,5800&13,6000&13,5030&13,5370\\
45$^{\circ}$&20,9890&20,4000&20,7600&20,0920&20,1050\\
\bottomrule 
\end{tabular}
}
\label{tab:Table6} 
\end{table}

\begin{table}[H] 
\captionsetup{width=0.85\columnwidth}
\caption{Buckling mode shapes of simply-supported skew plate predicted by DAEM with different aspect ratio and \textbf{$\theta=30^{\circ}$}} 
\vspace{-0.3cm}
\centering 
\resizebox{0.9\columnwidth}{!}{%
\begin{tabular}{c c c c} 
\toprule 
\toprule 
\textbf{Aspect ratio $\xi$} & \textbf{Buckling mode shape}&\textbf{Aspect ratio $\xi$} & \textbf{Buckling mode shape}\\ 
\midrule 
0,5	&\hspace{-4.5cm}\parbox[c]{1em}{\includegraphics[scale=0.1]{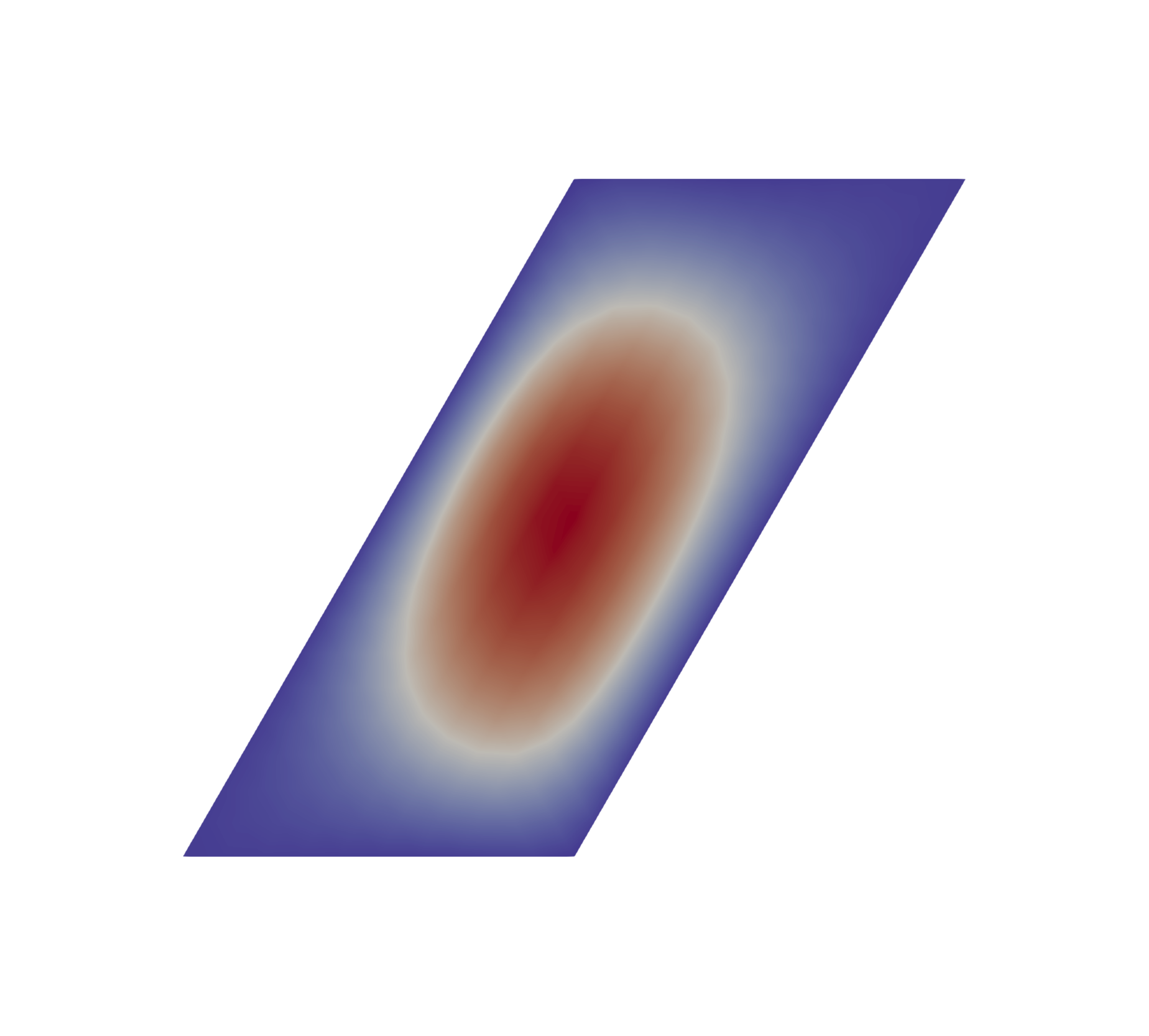}}& 1	&\hspace{-5.5cm}\parbox[c]{1em}{\includegraphics[scale=0.12]{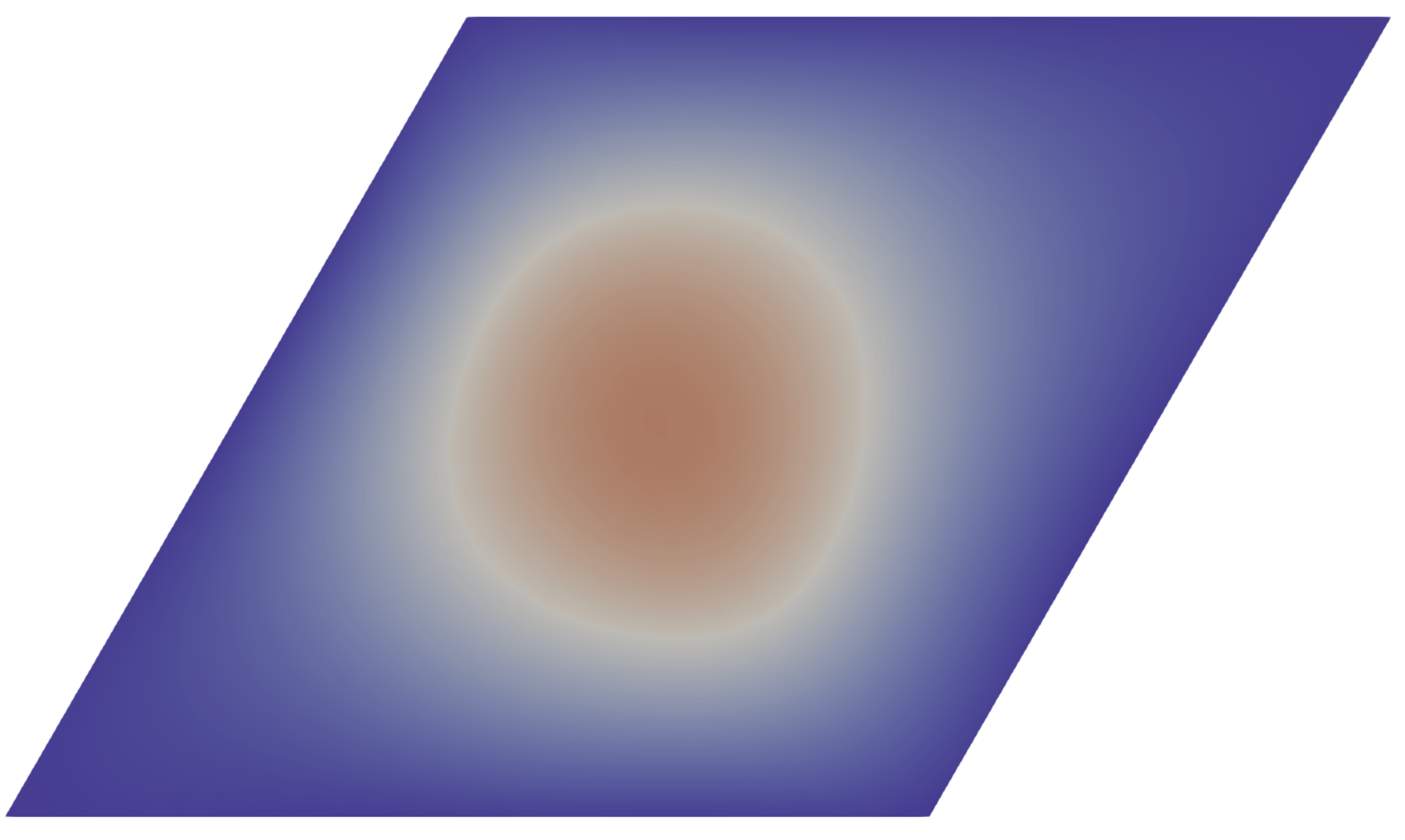}}\\ 
1,5	&\hspace{-5.8cm}\parbox[c]{1em}{\includegraphics[scale=0.12]{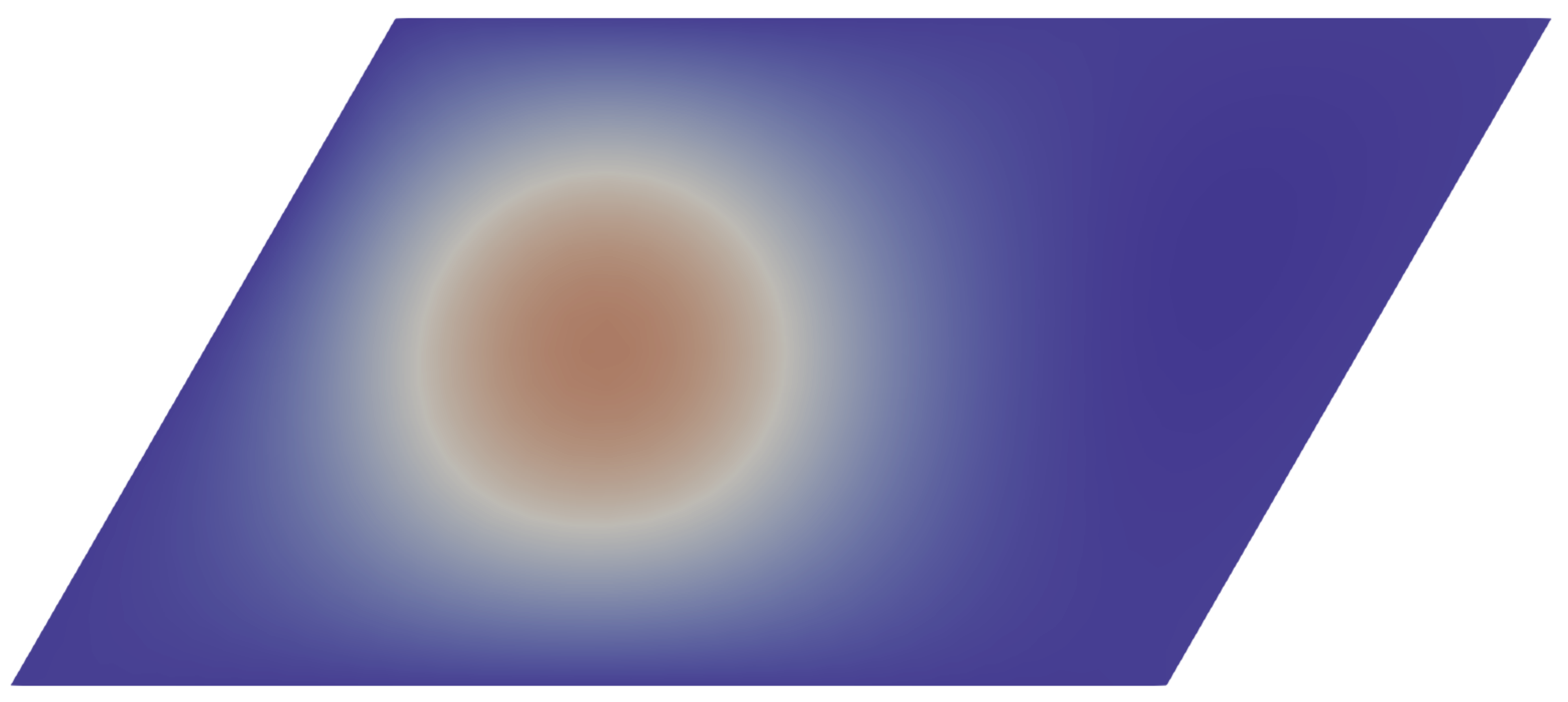}}& 2	&\hspace{-5.8cm}\parbox[c]{1em}{\includegraphics[scale=0.12]{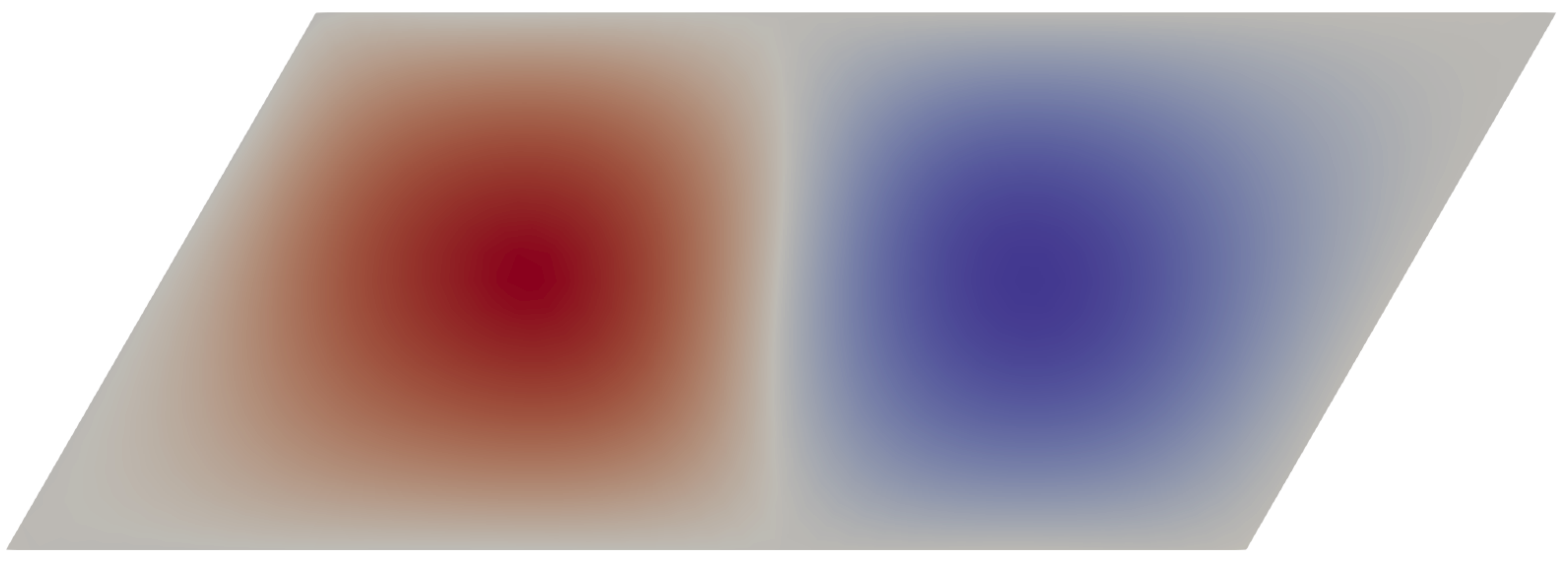}}\\ 
\bottomrule 
\end{tabular}
}
\label{tab:Table7} 
\end{table}

Let us focus exemplary on the mode shape of the simply-supported plate with varying aspect ratios and a fixed skew angle of $\theta=30$. The predicted mode shapes for each case are illustrated in Table~\ref{tab:Table7}. The buckling mode shapes for the clamped skew plate  and a fixed aspect ratio of $\xi=1$ and varying skew angle are listed in Table~\ref{tab:Table8}. The predicted results agree well with the ones in \cite{srinivasa2012buckling}. Note that once the hyperparameters of this neural network are obtained, the network can be used to predict similar problems quickly and since the training is an unsupervised learning, no target value on the solution is needed.
\begin{table}[H] 
\captionsetup{width=0.85\columnwidth}
\caption{Buckling mode shapes of clamped skew plate predicted by DAEM with different skew angle and \textbf{$\xi=1$}} 
\vspace{-0.3cm}
\centering 
\resizebox{0.9\columnwidth}{!}{%
\begin{tabular}{c c c c} 
\toprule 
\toprule 
\textbf{Skew angle $\theta$} & \textbf{Buckling mode shape}&\textbf{Skew angle $\theta$} & \textbf{Buckling mode shape}\\ 
\midrule 
0$^{\circ}$&\hspace{-3.5cm}\parbox[c]{1em}{\includegraphics[width=4cm]{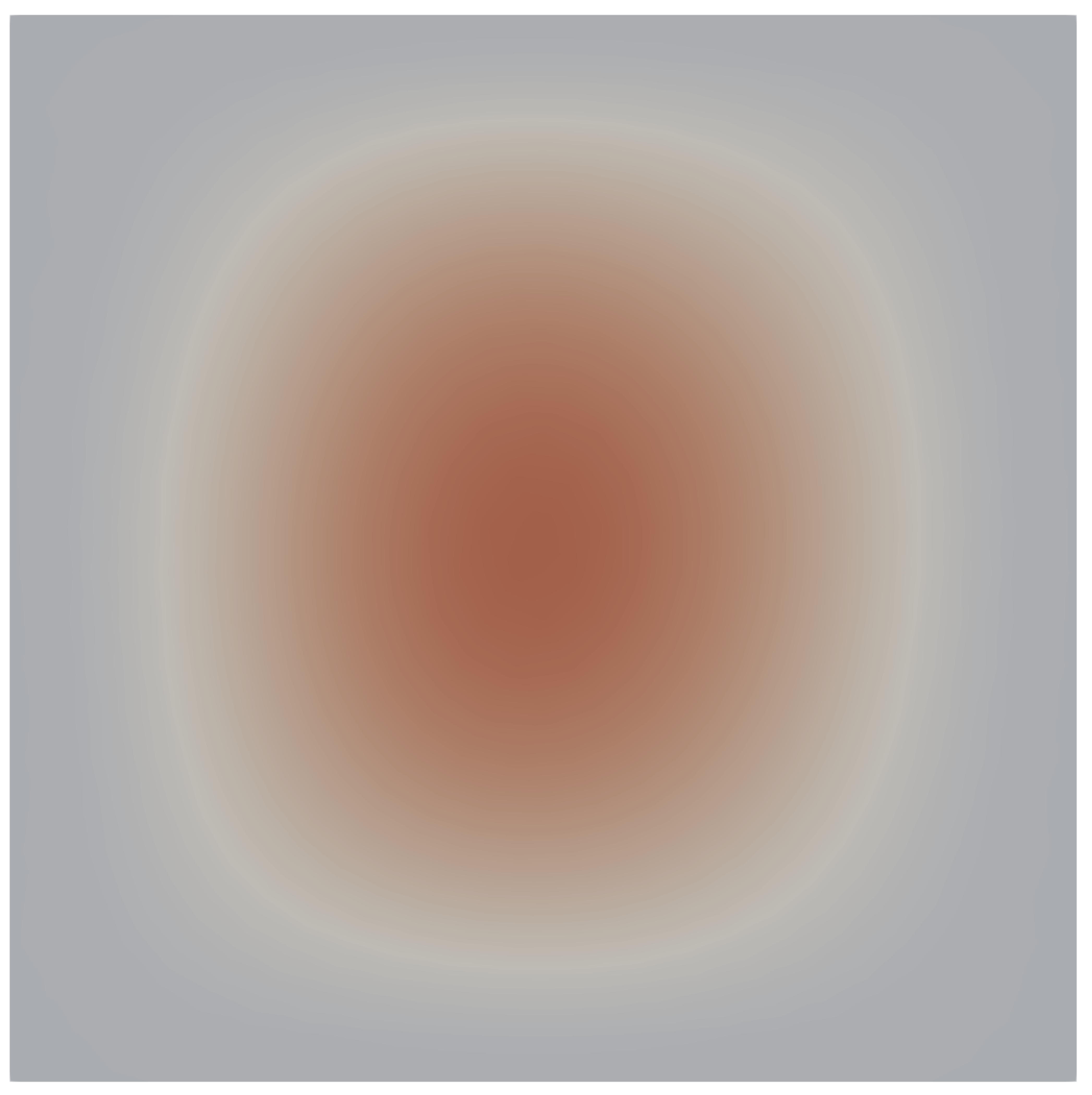}}&15$^{\circ}$&\hspace{-4.5cm}\parbox[c]{1em}{\includegraphics[width=5cm]{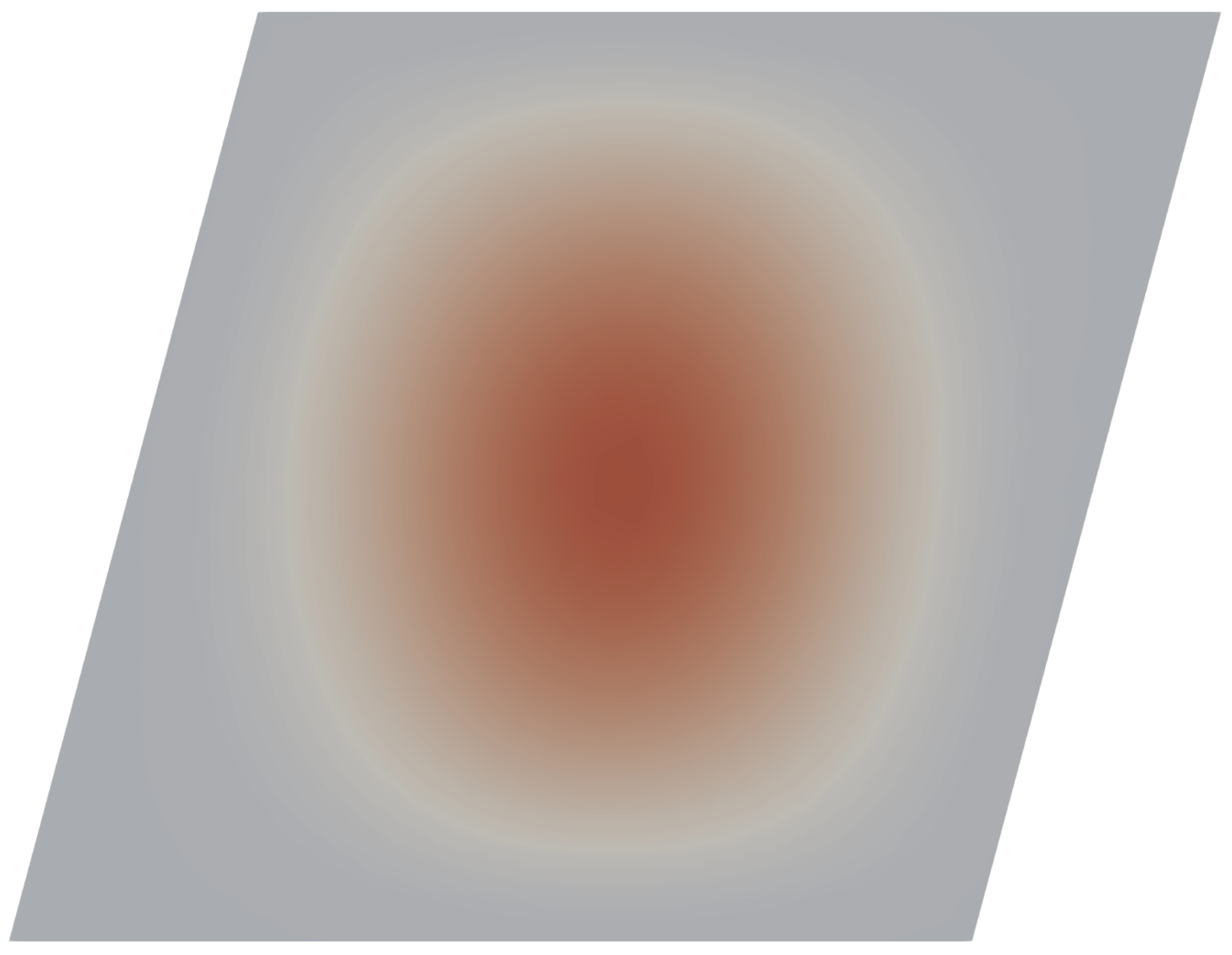}}\\ 
30$^{\circ}$&\hspace{-4.5cm}\parbox[c]{1em}{\includegraphics[width=6cm]{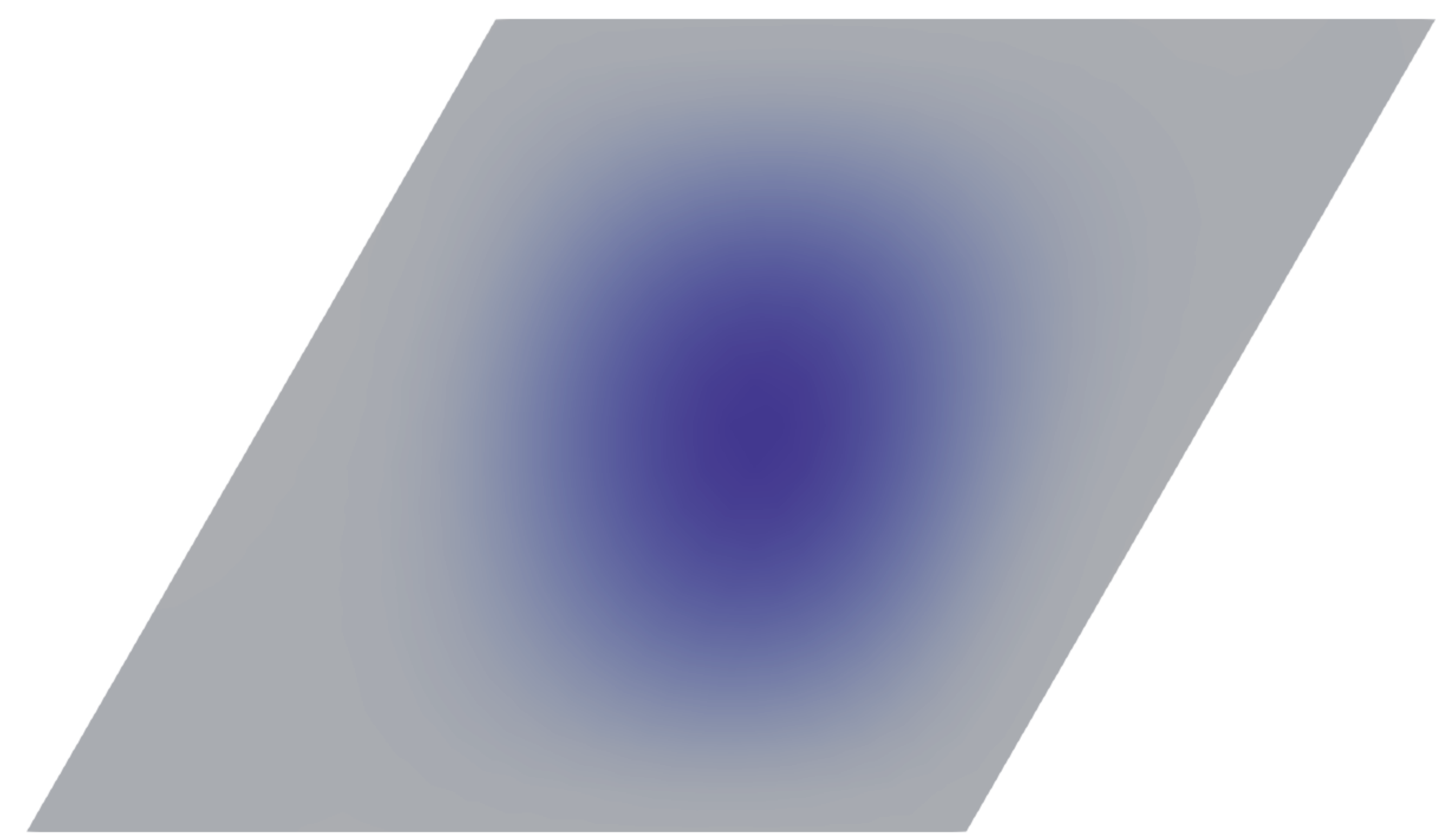}}&45$^{\circ}$&\hspace{-5.6cm}\parbox[c]{1em}{\includegraphics[width=6cm]{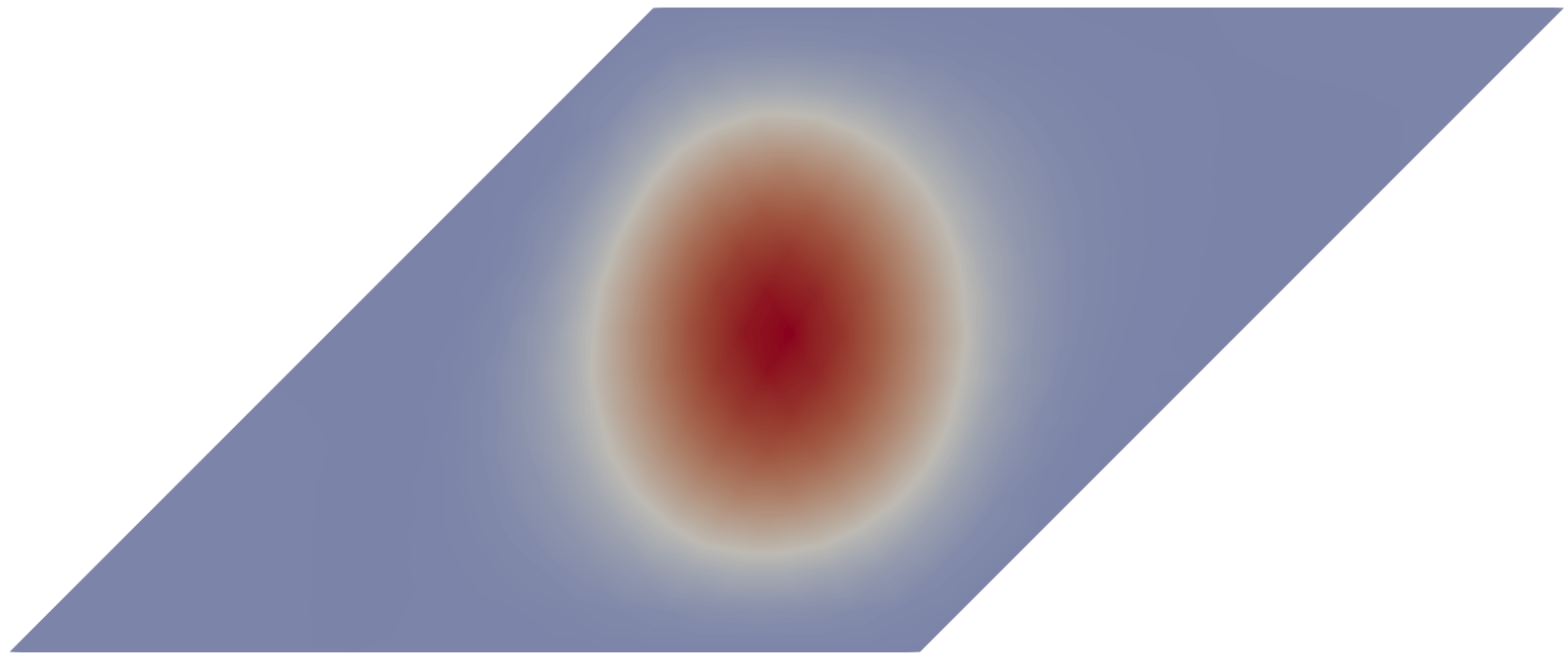}}\\ 
\bottomrule 
\end{tabular}
}
\label{tab:Table8} 
\end{table}

\section{Conclusions}
\label{sec:concl}

In this paper, a deep autoencoder based energy method  for bending, vibration and buckling analysis of Kirchhoff plate is proposed. For the proposed method, the deep autoencoder which is suitable for unsupervised feature extraction is combined with the minimum total potential energy principle to solve michanical analysis of Kirchhoff plate, and it has successfully discovered the underlying physical patterns. Moreover, a tailored activation is proposed for the deep autoencoder based energy method, which has been proven to be more stable and alleviated the gradient explosion problem without compromising computational efficiency. To calculate the total potential energy, the fitted Monte Carlo integration is adopted, and makes the whole method to be truly "meshfree" and very easily accessible. And once the deep autoencoder based energy method is trained, it can predict the physical features readily.

Further, the deep autoencoder based energy method has been applied to extracting funtamental frequency, critical buckling load and corresponding mode shapes based on Rayleigh's principle. Different numerical examples covering various type of Kirchhoff plate models, including bending, vibration, buckling of plate with different geometries, cutout, boundary and load conditions and even on Winkler foundation have been investigated to validate the proposed method. The accuracy and efficiency of the proposed activation is studied and compared in numerical examples. Further, the favourable deep autoencoder configuration are studied to offer practical guidance for application. Though this is still a preliminary research, numerical experiments have demonstrated favourable features for this deep autoencoder method for prediction the physical patterns behind the Kirchhoff plate model. And the proposed autoencoder based energy method is more simple and efficient than the deep collocation method, so it can be further applied to more engineering fields. 

However, those are just preliminary studies of this method. There are still several issues remain to be addressed, like a more suitable global optimization algorithm, and application to more complicated engineering problems, et al. Those are our future research projects, including the geometric and material nonlinear analysis in solid mechanics and computational fluid mechanics with deep autoencoder based energy method. 

\renewcommand\refname{Reference}
\bibliography{DLplate.bib}

\end{document}